\documentclass[graybox]{svmult}

\usepackage{type1cm}        
\usepackage{makeidx}         
\usepackage{graphicx}        
\usepackage{multicol}        
\usepackage[bottom]{footmisc}
\usepackage{breakcites} 
\usepackage{amsmath}
\usepackage{newtxtext}       %
\usepackage{newtxmath}       
\usepackage{pifont}
\usepackage{multirow}

\usepackage{booktabs}

\makeindex             

\usepackage{array}
\newcolumntype{P}[1]{>{\centering\arraybackslash}p{#1}}
\newcolumntype{M}[1]{>{\centering\arraybackslash}m{#1}}

\newcolumntype{C}[1]{>{\centering\arraybackslash }b{#1}}

\usepackage{comment}  
\usepackage{hyperref}
\newcommand{\cmark}{\ding{51}}%
\newcommand{\xmark}{\ding{55}}%

\newcommand{\LL}[1]{\mathrm{{L}}^{#1}}

\usepackage{bm}
\usepackage[normalem]{ulem}

\begin{document}

\title*{An Analysis of Generative Methods for Multiple Image Inpainting}

\author{Coloma Ballester, Aurélie Bugeau, Samuel Hurault,\\ Simone Parisotto, Patricia Vitoria}

\institute{Coloma Ballester \at University Pompeu Fabra, IPCV, DTIC, Spain \email{coloma.ballester@upf.edu}
\and Aurélie Bugeau \at Université de Bordeaux, LaBRI, CNRS UMR 5800, France \email{aurelie.bugeau@labri.fr}
\and Samuel Hurault \at Université de Bordeaux, Bordeaux INP, CNRS, IMB, UMR 5251,F-33400 Talence, France \email{samuel.hurault@math.u-bordeaux.fr}
\and Simone Parisotto \at DAMTP, University of Cambridge, Wilberforce Road, Cambridge CB3 0WA, United Kingdom \email{sp751@cam.ac.uk}
\and Patricia Vitoria \at University Pompeu Fabra,   IPCV, DTIC, Spain \email{patricia.vitoria@upf.edu}
}

%
%
\maketitle

\abstract{
Image inpainting refers to the restoration of an image with missing regions in a way that is not detectable by the observer. The inpainting regions can be of any size and shape.
This is an ill-posed inverse problem that does not have a unique solution. 
In this work, we focus on learning-based image completion methods for multiple and diverse inpainting which goal is to provide a set of distinct solutions for a given damaged image.
These methods capitalize on the probabilistic nature of certain generative models to sample various solutions that coherently restore the missing content. 
Along the chapter, we will analyze the underlying theory and analyze the recent proposals for multiple inpainting.
To investigate the pros and cons of each method, we present quantitative and qualitative comparisons, on common datasets, regarding both the quality and the diversity of the set of inpainted solutions. 
Our analysis allows us to identify the most successful generative strategies in both inpainting quality and inpainting diversity. This task is closely related to the learning of an accurate probability distribution of images. Depending on the dataset in use, the challenges that entail the training of such a model will be discussed through the analysis.
}


\section{Introduction}\label{sec:intro}

Image inpainting, also called \emph{amodal completion} or \emph{disocclusion} in early days, is an active research area in many fields including applied mathematics and computer vision, with foundations in the Gestalt theory of shape perception.
Inpainting relates to the virtual reconstruction of missing content in images in a way that is non-detectable by the observer \cite{BerSapCasBal2000}. 
It is an ill-posed inverse problem that can have multiple plausible solutions. 
Indeed, the fact that the inpainted image is not unique can be understood both mathematically and also because the reconstruction quality is judged by independent humans. 
On top of that, it has a strong impact on many real-life applications, \emph{e.g.}\ in medical imaging (sinograms \cite{Tovey2019}, CT scans \cite{Chen2012}), 3D surface data \cite{Biasutti2019}, \cite{bevilacqua2017joint}, \cite{hervieu2010stereoscopic}, \cite{Parisotto2020}, art conservation (frescoes \cite{BaaFornMarSch2008}, panel paintings \cite{RuiCorPla2011} and manuscripts \cite{Calatroni2018}), image compression \cite{Peter2015}, camera artifact removal \cite{vitoria2019automatic} and the restoration of old movies and videos \cite{Grossauer2006}, \cite{Newson2014}, just to name a few.

State-of-the-art image inpainting methods have achieved amazing results regarding the complex work of filling large missing areas in an image. However, most of the methods generally attempt to generate one single result from a given image, ignoring many other plausible solutions. In this chapter, we focus on analyzing recent advances in the inpainting literature, concentrating on the learning-based approaches for \emph{multiple} and \emph{diverse} inpainting.
The goal of those methods is to estimate multiple plausible inpainted solutions while being as much diverse as possible. Those methods mainly focus on the idea of exploiting image coherency at several levels along with the power of neural networks trained on large datasets of images.
Unlike previous one-to-one methods, multiple image inpainting offers the advantage of exploring a large space of possible solutions.
This procedure gives the capacity to the user to eventually choose the preferred fit under his/her judgment instead of leaving the task of singling out one solution to the algorithm itself.

This chapter is structured as follows. 
Section \ref{sec:stateoftheart}  provides a brief overview of both model-based and learning-based inpainting methods in the literature. 
Section \ref{sec:achieveDiversity} presents the underlying theory of several approaches 
for multiple and diverse inpainting together with a review of the most representative (to the best of our knowledge) state-of-the-art proposals using those particular strategies. 
Section \ref{sec:evalmetrics} presents the evaluation metrics for both inpainting quality and diverse inpainting. The multiple inpainting results of the methods of Section \ref{sec:achieveDiversity} are presented and compared in Section~\ref{sec:results} both quantitatively and qualitatively, on common datasets and masks, concerning three aspects: proximity to ground truth, perceptual quality, and inpainting diversity. 
Finally, Section \ref{sec:conclusions} concludes the presented analysis.

\section{A Walk through the Image Inpainting Literature}
\label{sec:stateoftheart}

In the literature, inpainting methods can fall under different categories, \emph{e.g.} \emph{local vs. non-local} depending on the ability to capture and exploit non-nearby content, or \emph{geometric vs. exemplar-based methods} depending on the action on points or patches.
For our purposes, it is more convenient to distinguish between learning- and model-based approaches, according to the usage or not of machine learning techniques.
For extensive reviews of existing inpainting methods we refer the reader to the works in  \cite{Guillemot2014}, \cite{Schonlieb2015}, \cite{Buyssens2015}, \cite{ParVitBalBugReySch2022}.

\paragraph{\textbf{Model-based Inpainting}}

Model-based inpainting methods are designed to manipulate an image by exploiting its regularity and coherency features with an explicit model governing the inpainting workflow. 
One approach for restoring geometric image content is to locally propagate the intensity values and regularity of the image level lines inward the inpainting domain with curvature-driven ~\cite{Nitzberg1993}, ~\cite{MasMor1998}, ~\cite{BalBerCasSapVer2001}, ~\cite{Chan2001}, ~\cite{ESEDOGLU2002}, ~\cite{Shen2003} and diffusion-based ~\cite{Caselles1998}, ~\cite{Shen2002}, ~\cite{Tschumperle2005} evolutionary partial differential equations (PDEs), possibly of fluid-dynamic nature \cite{BerSapCasBal2000}, \cite{Bertalmio2001}, \cite{Tai2007} or with coherent transport mechanisms \cite{Bornemann2007}, also by invoking variational principles \cite{Grossauer2003}, \cite{Bertozzi2007} and regularization (possibly of higher-order) priors \cite{Papafitsoros2013}. The filling-in of geometry, especially of small scratches and homogeneous content in small inpainting domains, is the most effective scenario of these methods, which perform poorly in the recovery of texture.
Such issue is overcome by considering a patch (a group of neighboring points in the image domain) as the imaging atom containing the essential texture element. The variational formulation of dissimilarity metrics based on the estimation of a correspondence map between patches \cite{Efros1999}, \cite{Bornard2002}, \cite{Demanet03imageinpainting}, \cite{Criminisi2004}, \cite{aujol2010exemplar} has led to the design of optimal copying-pasting strategies for inpainting large domains. However, these methods still fail, \emph{e.g.} in the presence of different scale-space features.
Thus, some researchers have exploited, also using a variational approach, the efficiency of PatchMatch \cite{Barnes2009} in computing a probabilistic approximation of correspondence maps between patches to average the contribution of multiple source patches during the synthesis step. 
For example,   \cite{Arias2011}, \cite{Newson2014} use it in a non-local mean fashion \cite{Wexler}, to inpaint rescaled versions of the original image with results propagated from the coarser to the finer scale, \cite{Cao2011} to guide the inpainting with geometric-sketches, \cite{Sun2005} to guide structures or  \cite{Mansfield2011}, \cite{Eller2016}, \cite{Fedorov2016} to account for geometric transformations of patches. 
However, these mathematical and numerical advances may result computationally expensive while suffering from the single-imaging source, and dependence on the initialization quality and the selection of associated parameters (\emph{e.g.} the size of the patch). 
Thus, it seems natural to explore if image coherency, smoothness and self-similarity patterns can be further exploited by augmenting the dataset of source images and eventually synthesize multiple inpainting solutions: this is where diverse inpainting with deep learning generative approaches is a significant step forward.

One of the earliest model-based inpainting works dealing with multiple source images is \cite{SungHaKang}, where salient landmarks are extracted in a scene under different perspectives and then synthesized by interpolation, guiding the imaging restoration. As said, model-based models are sensitive to initializations and chosen parameters: 
one way to diminish these drawbacks is to perform inpainting of the input image multiple times, by varying parameters like the patch size, the number of pyramid scales, initializations and inpainting methodologies.  
Thus, a final assembling step will produce an inpainted image which encodes locally the most coherent content \cite{Hays2007}, \cite{LeMeur2013}, \cite{Kumar2016}.  
Still, the computational effort of estimating several solutions with different parameters and their fine-tuning is remarkable, leading to the need for a one-encompassing strategy that can locally adapt the synthesis step from multiple source images. 
This task can be solved with learning-based methods.  

\paragraph{\textbf{Learning-based Methods}} 
Learning-based methods address image inpainting by learning a mapping from a corrupted input to the estimated restoration by training on a large-scale dataset. Besides capturing local or non-local regularities and redundancy inside the image or the entire dataset, those methods also exploit high-level information inherent in the image itself, such as global regularities and patterns, or perceptual clues and semantics over the images.

Early learning-based methods tackled the problem as a blind inpainting problem \cite{RenXuYanSun2015}, \cite{cai2017blind} by minimizing the distance between the predicted image and the ground truth. 
This type of methods behaved as an image denoising algorithm and was limited to tiny inpainting domains. 
To deal with bigger and more realistic inpainting regions, later approaches incorporated in the model the information provided by the mask, \emph{e.g.} \cite{kohler2014mask}, \cite{RenXuYanSun2015}, \cite{pathak2016context}, \cite{UlyanovVedaldiLempitsky2018}. 
Also, several modifications to vanilla convolutions have been proposed to explicitly use the information of the mask, like Partial Convolutions \cite{liu2018image}  and Gated Convolutions \cite{YuLinYangShenLuHiang2018free}, where the output of those layers only depends on non-corrupted points.
Additionally, attempts to increase the receptive field without increasing the number of layers have been proposed with dilated convolutions \cite{iizuka2017globally}, \cite{wang2018image} and contextual attention \cite{YuLinYangShenLuHiang2018generative}, \cite{YuLinYangShenLuHiang2018free}.
Learning to inpaint in a single step has shown to be a complex endeavor. Progressive learning approaches have also been introduced to split the learning into several steps: for instance, \cite{zhang2018semantic} progressively fills the holes from outside to inside; similarly, \cite{guo2019progressive}, \cite{zeng2020high}, \cite{li2020recurrent} also learn how to update the inpainting mask for next iteration, and \cite{Li_2019_ICCV} learns jointly structure and feature information.

To train the network, early approaches minimized some distance between the ground truth and the predicted image. But this approach takes into account just one of the several possible plausible solutions to the inpainting problem.  Several approaches have been proposed to overcome this drawback. Some works use perceptual metrics based on generative adversarial networks (GANs) aiming to generate more perceptually realistic results \cite{pathak2016context}, \cite{yeh2017semantic}, \cite{iizuka2017globally}, \cite{ YuLinYangShenLuHiang2018generative}, \cite{vitoria2018semantic}, \cite{vitoria2019semantic}, \cite{dapogny2019missing}, \cite{liu2019coherent}, \cite{lahiri2020prior}. Other works tackle the problem in the feature space by minimizing distances at feature space level \cite{fawzi2016image}, \cite{yang2017high}, \cite{VoDuongNgocPerez2018} by using an additional pre-trained network, or by directly inpainting those features \cite{yan2018shift}, \cite{zeng2019learning}. Also, two-steps approaches have been proposed. They are based on a first coarse inpainting \cite{yang2017high}, \cite{YuLinYangShenLuHiang2018generative}, \cite{liu2019coherent}, edge learning \cite{liao2018edge}, \cite{NazeriNgJosephQureshiEbrahimi2019}, \cite{Li_2019_ICCV} or structure prediction \cite{xiong2019foreground}, \cite{ren2019structureflow}, and followed by a refinement step adding finer texture details. Furthermore, \cite{Liu2019MEDFE} aimed to ensure consistency between structure and texture generation. 
Another big problem of early deep learning methods is that deep models treat input images with limited resolution. While first approaches were able to deal with images of maximum size $64\times64$, the latest methods can deal with $1024\times1024$ resolution images by using, for example, a multi-scale approach \cite{yang2017high}, \cite{zeng2019learning}, or even to $8K$ resolution by generating first a low-resolution solution and second its high-frequency residuals \cite{yi2020contextual}. 

Recent works (\emph{e.g.} \cite{zheng2019pluralistic}, \cite{zhao2020uctgan}, \cite{cai2020piigan}, \cite{peng2021generating}, \cite{wan2021high}, \cite{liu2021pd}) deal with the ill-posed nature of the problem by allowing more than one possible plausible solution to a given image. They aim to generate multiple and diverse solutions by using deep probabilistic models based on variational autoencoders (VAEs), GANs, autoregressive models, transformers, or a combination of them. Note that those type of methods has been also used for real case applications such as diverse fashion image inpainting \cite{han2019finet} and Cosmic microwave background radiation (CMB) image inpainting \cite{yi2020cosmovae}. In this chapter, we will focus on the study of multiple image inpainting methods. More precisely, we will review, analyze and compare, theoretically as well as experimentally, the different approaches proposed on the literature to generate inpainting diversity.

\section{How to Achieve Multiple and Diverse Inpainting Results?} 
\label{sec:achieveDiversity}
In this section, we will describe the different tools and  methods that successfully addressed multiple image inpainting. Later in Section \ref{sec:results}, we will conduct a thorough experimental study comparing these methods visually and quantitatively. 

As previously mentioned, image inpainting is an inverse problem with multiple plausible solutions. 
Generally, ill-posed problems are solved by incorporating some knowledge or priors into the solution. 
Mathematically, this is frequently done using a variational approach where a prior is added to a data-fidelity term to create an overall objective functional that is lastly optimized. 
The selected prior promotes the singling out of a particular solution. 
Traditionally, the incorporated priors were model-based, founded on properties of the expected solution. 

More recently, data-driven proposals have emerged where the prior knowledge on the image distribution is implicitly or explicitly learned via neural networks optimization (we refer to the recent survey \cite{arridge2019solving} and references therein).
Among them, generative methods have been used to learn the underlying geometric and semantic priors of a set of non-corrupted images. Indeed,
generative methods aim to estimate the probability distribution of a large set, ${\mathcal{X}}$, 
of data. In other words, any $x\in{\mathcal{X}}$ is assumed to come from an underlying and unknown probability distribution $\mathbb{P}_{\mathcal{X}}$ and the goal is to learn it from the data in ${\mathcal{X}}$. Due to its capacity to produce several outcomes given a single output, some authors have proposed to address the multiplicity of solutions by leveraging the probabilistic nature of generative models.

Through the chapter, we will assume that ${\mathcal{X}}$ is a set of images.
Images will be assumed to be functions defined on a bounded domain $\Omega \subset \mathbb{R}^2$ with values in $\mathbb{R}^C$, with $C=1$ for gray-level images and $C=3$ for color images. 
With a slight abuse of notation, we will use the same notation to refer to the continuous setting, where $\Omega\subset\mathbb{R}^2$ is an infinite resolution image domain and $x:\Omega\to\mathbb{R}^C$ represents a continuous image, and to the discrete setting where $\Omega$ stands for a discrete domain given by a grid of $H \times W$ pixels, $H,W\in\mathbb{N}$, and $x$ is a function defined on this discrete $\Omega$ and with values in $\mathbb{R}^C$. 
In the latter case, $x$ is usually given in the form of a real-valued matrix of size ${H\times W\times C}$ 
representing the image values. 

In the context of image inpainting, the inpainting domain, denoted here by $O$, represents the region of the image domain $\Omega$ where the image data is missing, and thus to be restored. Its complementary set, $O^c=\Omega\setminus O$, represents the region of $\Omega$ where the values of the image to be inpainted are known. The inpainting mask $M$ will be defined as equal to $1$ on the missing pixels of $O$, and equal to $0$ on $\Omega\setminus O$.

The space ${\mathcal{X}}$ of (complete) natural images is a high dimensional space and its distribution can be very complex. 
However, natural images contain local regularities, non-local self-similarities, global coherency and even semantic structure. 
This is one of the reasons that inspired the use of latent-based models. These models use latent variables $z\in{\mathcal{Z}}$ in a lower dimension space $\text{dim}({\mathcal{Z}})\leq\text{dim}({\mathcal{X}})$, associated with a probability distribution $\mathbb{P}_{\mathcal{Z}}$. Generative latent-based models aim to learn a generative model $G_{\theta}:\mathcal{Z}\to\mathcal{X}$, with parameters $\theta$, mapping a latent variable $z$ to an image $x$. With a slight abuse of notation and if it is understood from the context, we will forget about the $\theta$ subindex and simply write $G$. The main goal of this strategy is two-fold: (1) to be able to generate samples $G(z)$ hoping that $G(z)\in{\mathcal{X}}$ for $z\sim \mathbb{P}_{\mathcal{Z}}$ (or that $\mathbb{P}_G$ is close to $\mathbb{P}_{\mathcal{X}}$ ), and (2) to use it for density estimation 
\begin{equation}\label{eq:1}
    p_{\mathcal{X}}(x) \approx p_{G}(x) = \int p(x|z) p_{\mathcal{Z}}(z) dz,
\end{equation}
where $p_{G}$ stands for the parametric density of $\mathbb{P}_G = G_{\#} \mathbb{P}_{\mathcal{Z}}$, the pushforward measure of $\mathbb{P}_{\mathcal{Z}}$ through $G$ (defined in brief as $G_{\#} \mathbb{P}_{\mathcal{Z}}(B)=\mathbb{P}_{\mathcal{Z}}(\{z\in{\mathcal{Z}} | G(z)\in B \})$, for any $B$ in the Borel $\sigma$-algebra associated to ${\mathcal{X}}$). Let us notice that the likelihood $p(x|z)$ depends on $G$, and can be interpreted as a measure of how close $G(z)$ is to $x$. 

Numerous strategies have been developed to parametrize $G$ (or $G_{\theta}$) as a neural network model and to learn the appropriate parameters $\theta$ by making $\mathbb{P}_G$ as close as possible to $\mathbb{P}_{\mathcal{X}}$ for some probability distance $d(\mathbb{P}_G,\mathbb{P}_{\mathcal{X}})$. Among these strategies we quote variational autoencoders, normalizing flows, generative adversarial networks, or autoregressive models. 

The problem of image inpainting can also be naturally formulated in a probabilistic manner. 
Let $y$ denote an observed incomplete image, which is unknown in $O$.
We are interested in modeling the conditional distribution, $p(x|y)$, over the values of the variable $x$ (corresponding to the complete image) conditioned on the value of the observed variable $y$. 
As possibly many plausible images are consistent with
the same input image $y$, the distribution $p(x|y)$ will likely be multimodal. 
Then, each of the multiple solutions can be generated by sampling from that distribution using a given sampling strategy. 
Thus, the goal is not only to obtain a generative model that minimizes $d(\mathbb{P}_G,\mathbb{P}_{\mathcal{X}_s})$, where ${\mathcal{X}_s}\subset{\mathcal{X}}$ is the set of possible solutions, but also to design a mechanism able to sample the conditional distribution $p(x|y)$, \emph{i.e.} for a given damaged incomplete image $y$, output a set of plausible completions $x$ of $y$.
\begin{table}[!ht]
\caption{Generative methods used in the analyzed state-of-the-art proposals for diverse inpainting.
}
\label{tab:methodslist}
\begin{center}
\begin{tabular}{p{4.7cm}P{1cm}P{2cm}P{1cm}P{2cm}}
\svhline\noalign{\smallskip}
\textbf{Method} &  \textbf{VAE} &  \textbf{Autoregressive} & \textbf{GAN} &  \textbf{Transformers} \\
\noalign{\smallskip}\svhline\noalign{\smallskip}
PIC (\cite{zheng2019pluralistic}) &   \cmark   & & \cmark   &   \\
\noalign{\smallskip}\hline\noalign{\smallskip}
PiiGAN (\cite{cai2020piigan}) &    & & \cmark   &   \\
\noalign{\smallskip}\hline\noalign{\smallskip}
UCTGAN (\cite{zhao2020uctgan}) & \cmark & & \cmark  & \\
\noalign{\smallskip}\hline\noalign{\smallskip}
DSI-VQVAE (\cite{peng2021generating}) & \cmark & \cmark  & \cmark  &  \\
\noalign{\smallskip}\hline\noalign{\smallskip}
ICT (\cite{wan2021high}) &  &   & \cmark   & \cmark   \\
\noalign{\smallskip}\hline\noalign{\smallskip}
PD-GAN (\cite{liu2021pd}) &  & & \cmark    &    \\
\noalign{\smallskip}\hline\noalign{\smallskip}
BAT (\cite{yu2021diverse} )   &  &  & \cmark  & \cmark   \\
\noalign{\smallskip}\svhline\noalign{\smallskip}
\end{tabular}
\end{center}
\end{table}

In this section, we will analyze the different families of generative models proposed in the literature to realize diverse image inpainting. We will in particular describe generative adversarial networks (GAN), variational autoencoders (VAE), autoregressive models and transformers. We will also detail the different objective losses proposed to train these networks. Finally, for each family of models, we will review several state-of-the-art diverse inpainting methods that relate to this model. Table \ref{tab:methodslist} lists all the methods that will be reviewed in this section.

\subsection{Generative Adversarial Networks}\label{ssec:GAN}
Generative Adversarial Networks (GANs) are a type of generative models that have received a lot of attention since the seminal work of \cite{goodfellow2014generative}.
The GAN strategy is based on a game theory scenario between two networks, a generator network and and a discriminator network, that are jointly trained competing against each other in the sense of a Nash equilibrium.
The generator maps a vector from the latent space, $z\sim \mathbb{P}_{\mathcal{Z}}$, to the image space trying to trick the discriminator, while the discriminator receives either a generated or a real image and must distinguish between both. 
The parameters of the generator and the discriminator are learned jointly by optimizing a GAN objective by a min-max procedure. This procedure leads the probability distribution of the generated data to be as close as possible, for some distance, to the one of the real data. Several GAN variants have appeared. They mainly differ on the choice of the distance $d(\mathbb{P}_1,\mathbb{P}_2)$ between two probability distributions $\mathbb{P}_1$ and $\mathbb{P}_2$. The first GAN by \cite{goodfellow2014generative} (also referred to as vanilla GAN) makes use of the Jensen-Shannon divergence, which is defined from the Kullback–Leibler divergence (${\mathrm{KL}}$), by
\begin{equation}
    d_{\mathrm{JS}}(\mathbb{P}_1,\mathbb{P}_2)=  \frac{1}{2}\left[{\mathrm{KL}}\left(\mathbb{P}_1 | | \frac{\mathbb{P}_1+\mathbb{P}_2}{2}\right)+{\mathrm{KL}}\left(\mathbb{P}_2| | \frac{\mathbb{P}_1+\mathbb{P}_2}{2}\right)\right],
\end{equation}
where the ${\mathrm{KL}}$ is defined, for discrete probability densities, as
\begin{equation}
     \mathrm{KL}(\mathbb{P}_1,\mathbb{P}_2)=    \sum_x \mathbb{P}_1(x) \log \left( \frac{\mathbb{P}_1(x)}{\mathbb{P}_2(x)}\right).
     \label{eq:KLdiscrete}
\end{equation}
and, for continuous densities, as
\begin{equation}
\mathrm{KL}(\mathbb{P}_1,\mathbb{P}_2) = \int_{\mathcal{X}} \mathbb{P}_1(x) \log{\frac{\mathbb{P}_1(x)}{\mathbb{P}_2(x)}} dx.
\label{eq:KL}
\end{equation}
The Wasserstein GAN \cite{arjovsky2017wasserstein} uses the Wasserstein-1 distance, given by
\begin{equation}
\mathbb{W}_1(\mathbb{P}_1,\mathbb{P}_2)=\inf_{\pi \in \Pi(\mathbb{P}_1,\mathbb{P}_2)}{\mathbb{E}}_{x,y \sim \pi}(\|x-y\|),
\end{equation}
where $\Pi(\mathbb{P}_1,\mathbb{P}_2)$ is the set of all joint distributions $\pi$ whose marginals are respectively $\mathbb{P}_1$ and $\mathbb{P}_2$. 
By Kantorovitch-Rubenstein duality, the Wasserstein-1 distance can be computed as
\begin{equation}
\mathbb{W}_1(\mathbb{P}_1,\mathbb{P}_2)=\sup_{D\in{\cal{D}}} \left( {\mathbb{E}}_{x\sim\mathbb{P}_1}[D(x)]- {\mathbb{E}}_{y\sim\mathbb{P}_2}[D(y)] \right), 
\end{equation}
where ${\cal{D}}$ denotes the set of 1-Lipschitz functions. In practice, the dual variable $D$ is parametrized by a neural network 
and it represents the so-called discriminator. 

Both the generator and discriminator are jointly trained to solve 
\begin{equation}\label{eq:WassGAN}
\min_{G}\sup_{D\in{\cal{D}}} \left( {\mathbb{E}}_{x\sim\mathbb{P}_{\mathcal{X}}}[D(x)]- {\mathbb{E}}_{y\sim\mathbb{P}_G}[D(y)] \right), 
\end{equation}
in the case of the  Wasserstein GAN, and
\begin{equation}\label{eq:vanillaGAN}
             \min_G \max_D 
             \mathbb{E}_{x \sim \mathbb{P}_{\mathcal{X}}}\left[\log D(x)\right] + \mathbb{E}_{y \sim \mathbb{P}_{G}}\left[\log (1-D(y))\right] 
\end{equation}
for the vanilla GAN. In \eqref{eq:vanillaGAN}, the discriminator $D$ is simply a classifier that tries to distinguish samples in the training set $\mathcal{X}$ (real samples) from the generated samples $G(z)$ (fake samples) by designing a probability $D(x)\in [0,1]$ for its likelihood to be from the same distribution as the samples in $\mathcal{X}$.

GANs are sometimes referred to as implicit probabilistic models due to the fact that they are defined through a sampling procedure where the generator learns to generate new image samples. This is in contrast to variational autoencoders, autoregressive models, and methods that explicitly maximize the likelihood.

For the task of inpainting, several proposals set the problem as a conditioned one. The GAN approach is modified such that the input of the generator $G$ is both an incomplete image $y$ and a latent vector $z\sim \mathbb{P}_{\mathcal{Z}}$, and $G$ performs conditional image synthesis where the conditioning input is $y$.
In the GAN-based works that we present in this section (\cite{cai2020piigan,liu2021pd}), the authors focus on multimodal conditioned generation where the goal is to generate multiple plausible output images for the same given incomplete image.

Finally, let us mention that in these works, and in general in some works described in this chapter, the used generative methods are combined with consistency losses that encourage the inpainted images to be close to the ground truth. Examples of those consistency losses include value and feature reconstruction losses, and perceptual losses. Nonetheless, multiple inpainting researchers acknowledge that it can be counterproductive to rely on consistency losses due to the fact that the ground truth is only one of the multiple solutions.

\bigskip
\noindent{\textbf{PiiGAN: Generative Adversarial Networks for Pluralistic Image Inpainting (\cite{cai2020piigan})} }
\bigskip

\begin{figure}
    \centering
    \includegraphics[width=\textwidth]{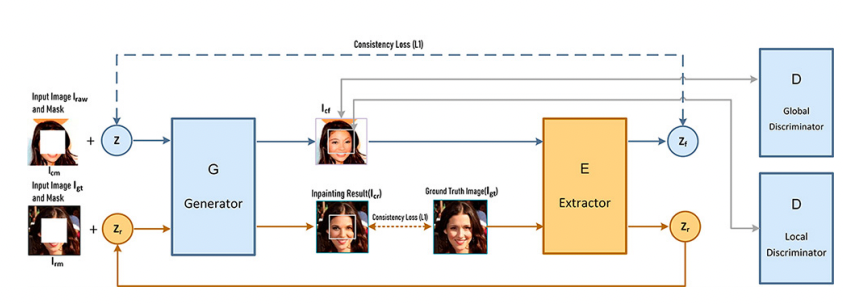}
    \caption{Overview of the architecture of PiiGAN: Generative Adversarial Networks for Pluralistic Image Inpainting \cite{cai2020piigan}. Figure from \cite{cai2020piigan}.}
    \label{fig:PiiGAN}
\end{figure}

One of the first methods that use GANs in order to generate pluralistic results is PiiGAN \cite{cai2020piigan}. PiiGAN is a deep generative model that incorporates a style extractor that can extract the style features, in the form of a latent vector, from the ground truth image.

To be more precise, the network is composed of one \emph{generator} and \emph{extractor} network. 
The training follows two different paths.

First, given a ground truth image $x_{gt}$, the extractor is used to estimate the style feature $z_{gt}$.  Once the style feature $z_{gt}$ is obtained, the ground truth image $x_{gt}$ is masked and concatenated with the computed style feature $z_{gt}$ and used as input of the \emph{generator} network. The generator network will estimate the inpainted version $x$ of the masked ground truth image $y$. The estimated inpainted image $x_{out,1}$ is passed through the extractor network to estimate the corresponding style feature $z_{out,1}$. This path is supervised using the KL-divergence between the style features  $z_{gt}$ and  $z_{out,1}$.

In parallel, another path estimates inpainted images from masked images without ground truth. That is, masked images without ground truth $y_{raw}$ are fed to the generator with a random vector $z_{raw}$. An inpainted image $x_{out,2}$ is predicted followed by style feature prediction $z_{out,2}$. Additionally, they frame the inpainting of $y_{raw}$ in an adversarial approach equipped with a local (that focuses just in the inpainted area) and a global discriminator applied to the inpainted image $x_{raw}$. This path, is supervised using the $\LL{1}$ norm of the difference between the style features  $x_{raw}$ and  $x_{out,2}$, and adversarial loss applied to the inpainted image $x_{out,2}$ based on the Wasserstein loss \eqref{eq:WassGAN} with gradient penalty as defined in \cite{gulrajani2017improved}. 

The authors claim that their results are diverse and natural, especially for images with large missing areas. Figure \ref{fig:PiiGAN} shows an overview of the algorithm pipeline.

\bigskip
\noindent \textbf{PD-GAN: Probabilistic Diverse GAN for Image Inpainting (\cite{liu2021pd})}
\bigskip

\begin{figure}
    \centering
    \includegraphics[width=\textwidth]{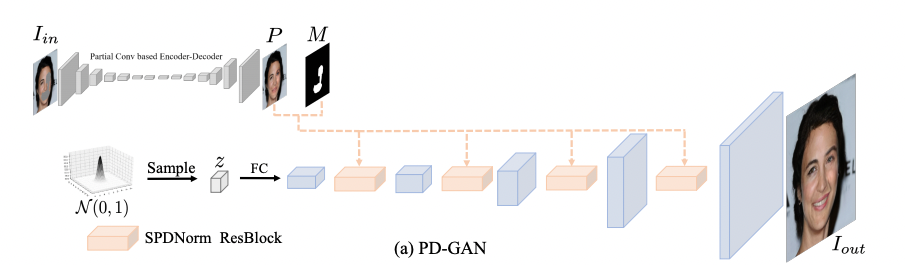}
    \caption{Overview of the architecture of PD-GAN: Probabilistic Diverse GAN for Image Inpainting \cite{liu2021pd}. Figure from \cite{liu2021pd}.}
    \label{fig:PDGAN}
\end{figure}

The authors of \cite{liu2021pd} propose a method to perform diverse image inpainting called PD-GAN. 
PD-GAN takes advantage of the benefits of GANs in generating diverse content from different random noise inputs. 
Figure \ref{fig:PDGAN} displays an overview of the algorithm pipeline.
In contrast to the original vanilla GAN, in PD-GAN all the decoder deep features are modulated from coarse to fine by injecting prior information at each scale. This prior information is extracted from an initially restored image at a coarser resolution together with the inpainting mask. 
For that purpose, they introduce a  probabilistic diversity normalization (SPDNorm) module based on the Spatially-adaptive denormalization (SPADE) module proposed in \cite{park2019semantic}.
SPDNorm works by modeling the probability of generating a pixel conditioned on the context information. It allows more diversity towards the center of the inpainted hole and more deterministic content around the inpainting boundary.

The objective loss is a combination of several losses, including a diversity loss, a reconstruction loss,  an adversarial loss, and a feature matching loss (difference in the output feature layers computed with the learned discriminator). In general, in the context of multiple image synthesis, \emph{diversity losses} aim at ensuring that the different reconstructed images are diverse enough. In particular, the authors of PD-GAN \cite{liu2021pd} use the so-called \emph{perceptual diversity loss}, defined as
\begin{equation}
            \mathcal{L}_{pdiv}(x_{{out}_a},x_{{out}_b}) = \frac{1}{ \sum_l \|M\odot (\Phi_l(x_{{out}_a})-\Phi_l(x_{{out}_b}))\|_1+\epsilon}.
            \label{eq:perceptual_diversity_loss}
\end{equation}
where $x_{{out}_a}$ and $x_{{out}_b}$ are two inpainted results, and $M$ the inpainting mask (with $1$ values on the missing pixels and $0$ elsewhere).
The minimization of \eqref{eq:perceptual_diversity_loss} favors the maximization of the perceptual distance of inpainted regions in $x_{{out}_a}$ and $x_{{out}_b}$. Notice that the non-masked pixels are not affected by this loss. A similar diversity loss was proposed in \cite{mao2019mode}.

\subsection{Variational Autoencoders and Conditional Variational Autoencoders}\label{ssec:VAE}

Variational Autoencoders (VAE) (\cite{kingma2013auto}) are generative models for which the considered distance between probability distributions is the Kullback-Leibler (KL) divergence. 
Minimizing $\mathrm{KL}(\mathbb{P}_{G_{\theta}} | |\mathbb{P}_{\mathcal{X}})$ with respect to the parameters $\theta$, 
is equivalent to the log likelihood maximization
\begin{equation}\label{eq:vae1}
    \max_{\theta}{\mathbb{E}}_{x\sim\mathbb{P}_{\mathcal{X}}} \log p_{G_{\theta}}(x)
\end{equation}
where, in the VAE context, $G_\theta$ is referred to as the \emph{decoder}. 

Let us first derive the vanilla VAE formulation in the general context of non-corrupted images $x\in{\mathcal{X}}$. Using Bayes rule, the likelihood $p_{G_\theta}(x)$, for $x\sim\mathbb{P}_{\mathcal{X}}$ and $z\sim\mathbb{P}_{\mathcal{Z}}$ , is given by 
\begin{equation}
    p_{\theta}(x)=\frac{p_{\theta}(x,z)}{p_{\theta}(z|x)}=\frac{p_{\theta}(x|z) p_{\mathcal{Z}}(z)}{p_{\theta}(z|x)}
\end{equation}
where, to simplify notations, we have denoted $p_{G_{\theta}}$ simply by $p_{\theta}$. In order to bypass the intractability of the posterior $p_{\theta}(z|x)$,
Variational Autoencoders introduce a second neural network, $q_{\psi}(z|x)$, to parametrize an approximation of the true posterior. This neural network is referred to as the \emph{encoder}. Let us now derive the VAE objective function. Following \cite{kingma2019introduction},
\begin{align}\label{eq:elbo3}
    \log p_{\theta}(x) = & \mathbb{E}_{q_{\psi}(z|x)} \left[\log\left[\frac{p_{\theta}(x,z)}{q_{\psi}(z|x)}\right]\right] + \mathbb{E}_{q_{\psi}(z|x)} \left[\log\left[\frac{q_{\psi}(z|x)}{p_{\theta}(z|x)}\right]\right]\\
    = & \mathbb{E}_{q_{\psi}(z|x)} \left[\log p_{\theta}(x,z)-\log q_{\psi}(z|x)\right] + {\mathrm{KL}}\left(q_{\psi}(z|x)| | p_{\theta}(z|x)\right) \\
    = & {\mathcal{L}_{\theta,\psi}(x)} + {\mathrm{KL}}\left(q_{\psi}(z|x)| | p_{\theta}(z|x)\right).
\end{align}
$\mathcal{L}_{\theta,\psi}$ is the so-called Evidence Lower Bound (ELBO).  By positivity of the KL, it verifies
\begin{equation}\label{eq:elbo4}
    {\mathcal{L}}_{\theta,\psi}(x)=\log p_{\theta}(x)-{\mathrm{KL}}\left(q_{\psi}(z|x)| | p_{\theta}(z|x)\right)\leq \log p_{\theta}(x),
\end{equation} 
and $ {\mathcal{L}}_{\theta,\psi}(x)=\log p_{\theta}(x)$ if and only if $q_{\psi}(z|x)$ is equal to $p_{\theta}(z|x)$.

VAE training consists in maximizing ${\mathcal{L}}_{\theta,\psi}$ in \eqref{eq:elbo4} with respect to the parameters $\{\theta,\psi\}$ of the encoder and of the decoder, simultaneously. The goal is to obtain a good approximation $q_{\psi}(z|x)$ of the true posterior $p_{\theta}(z|x)$ while maximizing the marginal likelihood $p_{\theta}(x)$.

The work \cite{sohn2015learning} extends VAEs by proposing Conditional Variational Autoencoders (CVAE). Their targeted distribution is the conditional distribution of $x$ given an input "conditional" variable $c$ and \eqref{eq:vae1} becomes 
\begin{equation}\label{eq:vae2}
    \max_{\theta}{\mathbb{E}}_{x\sim\mathbb{P}_{\mathcal{X}}} \log p_{G_{\theta}}(x|c).
\end{equation}
CVAE loss is obtained with a similar argument as in \eqref{eq:elbo3}-\eqref{eq:elbo4} by maximizing the conditional log-likelihood which gives the variational lower bound of the conditional log-likelihood
\begin{equation}\label{eq:elbo5}
  \mathbb{E}_{q_{\psi}(z|x)}\left[\log p_{\theta}(x|c,z)\right]-{\mathrm{KL}}\left(q_{\psi}(z|x,c)| | p_{\theta}(z|x)\right)\leq \log p_{\theta}(x|c).
\end{equation}
Then, the idea of the deep conditional generative modeling is simple: given an  observation (input) $x$, $z$ is drawn from a prior distribution $p_{\theta}(z|x)$. Then, the output is generated from the distribution $p_{\theta}(x|z,c)$.
\cite{bao2017cvae} combines a CVAE with a GAN (CVAE-GAN) for fine-grained category image generation. Even if inpainting results are shown,
the network is not trained explicitly for inpainting but for image generation conditioned on image labels. 

In the context of multiple image inpainting, or more generally of multiple image restoration, a straightforward idea is to condition the generative model on the input degraded image $y$ and to generate multiple images $x$ sampling from $p_{\theta}(x|z,c=y)$. BicycleGAN \cite{zhu2017toward} uses this idea for diverse image-to-image translation. Their goal is to learn a bijective mapping between two image domains with a multi-modal conditional distribution. They combine CVAE-GAN with latent regressors and  show that their method can produce both diverse and realistic results across various image-to-image translation problems. However, their method is not explicitly applied for image inpainting.
Moreover, as observed by several authors (see, \emph{e.g.} \cite{zheng2019pluralistic}, \cite{wan2021high}), using standard conditional VAEs or CVAE-GAN for the specific task for image inpainting still leads to minimal diversity and quality. Several extension of these models have recently appeared for diverse image inpainting. They are  presented below with more details.

Finally, let us notice that VAE model has been extended in \cite{van2017neural,razavi2019generating} to Vector Quantized-
Variational AutoEncoder (VQ-VAE) that uses vector quantization to model discrete latent variables. Such discretization is done to avoid posterior collapse. The quantization codebook is trained at the same time as the auto-encoder with an objective loss made of a reconstruction term and a regularization term that ensures that the embedding commits to the encoder outputs and respectively. 
The work \cite{razavi2019generating} is an hierarchical extension of \cite{van2017neural}. In particular, the authors of \cite{razavi2019generating} show that, by only considering two levels of a multi-scale hierarchical organization of VQ-VAE \cite{van2017neural}, the information about image texture is disentangled from the information about the structure and geometry of the objects in an image. By combining the obtained hierarchical
multi-scale latent data with an autoregressive model as prior (see Section \ref{ssec:autoregressive} below), they show an improved ability for generating high-resolution images.

\bigskip
\bigskip
\noindent\textbf{Pluralistic Image Completion \cite{zheng2019pluralistic}}
\bigskip

\begin{figure}[h!]
    \centering
    \includegraphics[width=0.95\textwidth]{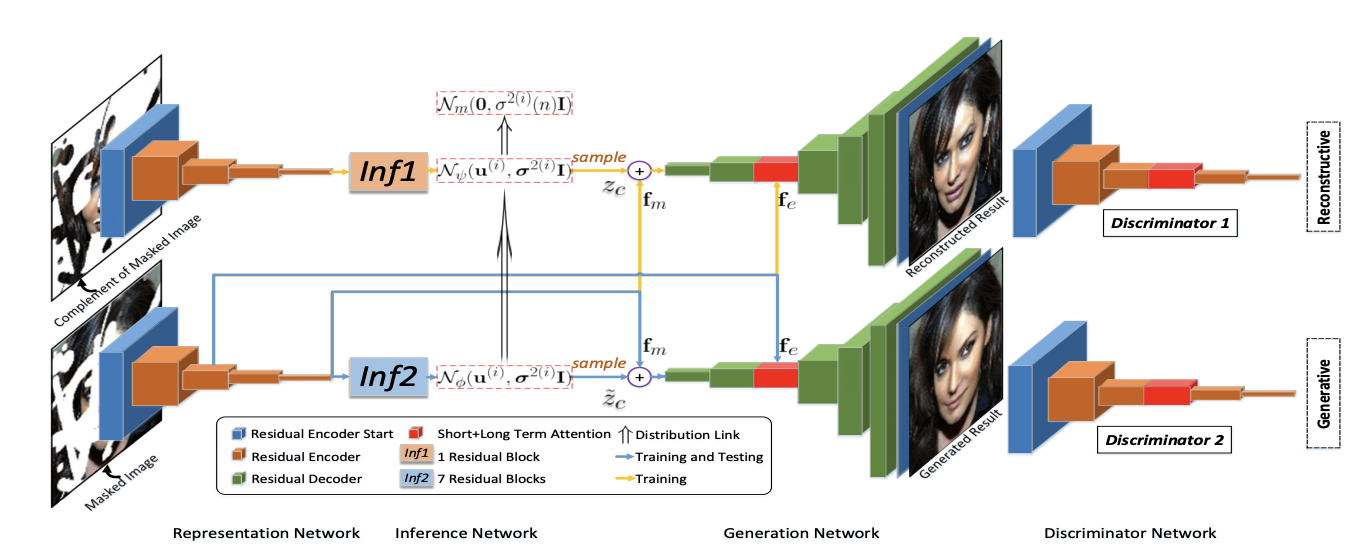}
    \caption{Overview of the PIC architecture of Pluralistic Image Completion \cite{zheng2019pluralistic}. Figure from \cite{zheng2019pluralistic}.}
    \label{fig:PIC}
\end{figure}

The work \cite{zheng2019pluralistic} aims to estimate a probability distribution $p(x|y)$ from which to sample, where $y$ represents an incomplete image and $x$ one of its possible completions. 
They propose to use the Conditional Variational Autoencoder (CVAE)  \cite{sohn2015learning}  approach described above which estimates a parametric distribution over a latent space, as in equation \eqref{eq:elbo5}, from which sampling is possible. 
However, in \cite{zheng2019pluralistic} the authors observe that if they explicitly promote the inpainted output to be similar to the ground truth image (either by any error-based loss such as, for instance, the L$^1$ distance, or as the authors show, by maximizing $\mathbb{E}_{q_{\psi}(z|x)}\left[\log p_{\theta}(x|y,z)\right]$ in \eqref{eq:elbo5} while ${\mathrm{KL}}\left(q_{\psi}(z|x,y)| | p_{\theta}(z|x)\right)$ tends to zero), it results in a lack of diverse outputs.
Alternatively, one could impose to fit the distribution of the training dataset by an adversarial approach including a discriminator as described in Section~\ref{ssec:GAN}. However, this approach is highly unstable. 
Instead, they propose a probabilistic framework with a dual pipeline composed of two paths. 
See a detailed pipeline in Figure \ref{fig:PIC}.
One is the \emph{reconstructive path} which is a VAE-based model that utilizes the ground truth to get a prior distribution of missing parts, $x|_{O}$, with the variance on the latent variables' prior depending on the hole area, and rebuild the exact same ground truth image from this distribution. 
The other is a \emph{generative path} for which the conditional prior, based only on the visible regions, is coupled to the distribution obtained in the reconstructive path to generate multiple and diverse samples. 
Both parts are framed in an adversarial approach to fit the distribution of the training dataset. 
Accordingly, the whole training loss is a combination of three types of terms. First, they use the KL divergences between the mentioned distributions. Second, the appearance/error terms based on the L$^1$ norm, where in the generative path it only has into account the visible pixels. And third and lastly, an adversarial discriminator-based term. It is based on the L$^1$ difference among the discriminator features of the ground truth and the reconstructed image for the reconstructive path, and on the discriminator value on the generated image for the generative path. Additionally, to exploit the distant relation among the encoder and decoder, they use a modified self-attention layer that captures fine-grained features in the encoder and more semantic generative features in the decoder. 

\noindent\textbf{UCTGAN: Diverse Inpainting based on Unsupervised Cross-Space Translation \cite{zhao2020uctgan}}
\bigskip

\begin{figure}
    \centering
    \includegraphics[width=\textwidth]{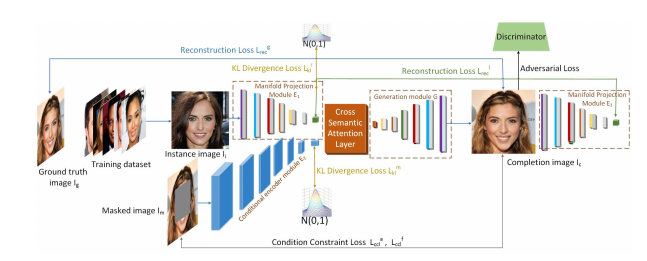}
    \caption{Overview of the architecture of UCTGAN: Diverse Inpainting based on Unsupervised Cross-Space Translation \cite{zhao2020uctgan}. Figure from \cite{zhao2020uctgan}.}
    \label{fig:UCTGAN}
\end{figure}

The authors of \cite{zhao2020uctgan} aim to produce multiple and diverse solutions conditioned by an instance image that guides the reconstruction, again aiming to maximize the conditional log-likelihood involving the variational lower bound \eqref{eq:elbo5} on the training dataset. They call their proposal UCTGAN. The pipeline is presented in Figure \ref{fig:UCTGAN}. They use a two-encoders network that transforms the instance image and the corrupted image to a low-dimensional manifold space. A cross semantic attention layer combines the information in both low-dimensional spaces. Consecutively, a generator is used to compute the conditional reconstructed image. 

The objective loss is composed of four terms. First, a constraint loss in the uncorrupted pixels is applied by minimizing the L$^{1}$ norm of the difference both at pixel and feature level. Second, The \rm{KL} divergence is used to project the low-dimensional manifold space of the instance image and masked image into a multivariate normal distribution space. Additionally, the L$^{1}$ norm of the difference in the low-dimensional manifold space of the instance image and the ground truth image is added. Finally, all the training is framed in an adversarial approach using the vanilla GAN \eqref{eq:vanillaGAN}, where the discriminator works in the image space.

\bigskip
\noindent\textbf{Generating Diverse Structure for Image Inpainting With Hierarchical VQ-VAE \cite{peng2021generating} }
\bigskip

The multiple inpainting proposal in \cite{peng2021generating} leverages three generative strategies, namely, variational autoencoders, generative adversarial methods, and autoregressive models. We first review below the main ideas of autoregressive models and then describe the proposal \cite{peng2021generating}.

\subsection{Autoregressive Models}\label{ssec:autoregressive}
In autoregressive models \cite{van2016pixel,oord2016conditional,chen2018pixelsnail}, the likelihood $p_{\theta}(x)$ is learned by choosing an order of the data variables $x=(x_1,x_2,\dots,x_n)\in{\mathcal{X}}$, frequently related to values on the $n$ pixels of an image, and exploiting the fact that the joint distribution can be decomposed as
\begin{equation}\label{eq:autore}
    p(x)=p(x_1,x_2,\dots,x_n)=p(x_1)\prod\limits_{i=2}^n p(x_i|x_1,\dots,x_{i-1}).
\end{equation}
More generally, a similar decomposition to \eqref{eq:autore} can be obtained by splitting the set of variables in smaller disjoint subsets. In this case, and considering the variable order of $x_1,x_2,\dots,x_n$ to be represented by a directed and non-cyclic graph, one has 
\begin{equation}
    p(x)=p(x_1,x_2,\dots,x_n)=p(x_1)\prod\limits_{i=2}^m p(x_i|S(x_i)),
\end{equation}
where $S(x_i)$ is the set of parent variables of variable $i$, and $m\leq n$. 

Autoregressive models have been used to learn a probability distribution or a conditional distribution, for instance in the context of VAEs (cf. Section \ref{ssec:VAE}) to model the prior or the decoder, and also to tackle several problems in imaging such as image generation (\emph{e.g.}~\cite{razavi2019generating}), super resolution (\emph{e.g.}~\cite{dahl2017pixel}), inpainting (\emph{e.g.}~\cite{peng2021generating}) or image colorization (\emph{e.g.}~\cite{zhao2020pixelated,guadarrama2017pixcolor,royer2017probabilistic}),  and also for other types of data such as audio and speech synthesis (\emph{e.g.}~\cite{oord2018parallel}) or text (\emph{e.g.}~\cite{bowman2015generating}).

\bigskip
\noindent\textbf{Generating Diverse Structure for Image Inpainting With Hierarchical VQ-VAE \cite{peng2021generating} }
\bigskip

\begin{figure}
    \centering
    \includegraphics[width=\textwidth]{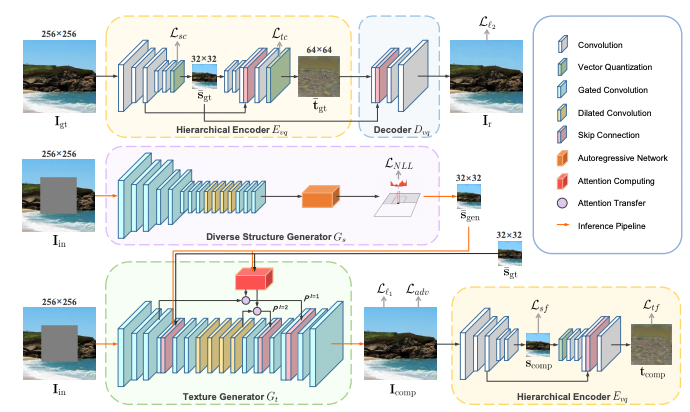}
    \caption{Overview of the architecture of Generating Diverse Structure for Image Inpainting With Hierarchical VQ-VAE (DSI-VQVAE) \cite{peng2021generating}.  Figure from \cite{peng2021generating}.}
    \label{fig:VQ-VAE}
\end{figure}

Inspired by the hierarchical vector quantized variational autoencoder (VQ-VAE) \cite{razavi2019generating}  whose hierarchical architecture disentangles structural and textural information, the authors of \cite{peng2021generating} propose a two-stage pipeline (cf. Figure \ref{fig:VQ-VAE}). 
As already pointed out by \cite{razavi2019generating}, by using a two-step approach instead of directly computing the final inpainted image, they aim to generate richer structure and texture images than previous VAE-based methods that often produce a distorted structure or blurry textures. 

The first stage of \cite{peng2021generating}, known as \emph{diverse structure generator},  generates multiple low-resolution results each of which has a different structure by sampling from a conditional autoregressive distribution. 
The second stage, known as \emph{texture generator}, uses an encoder-decoder architecture with a structural attention module that refines each low-resolution result separately by augmenting texture. The structural information module facilitates the capture of distant correlations. They further reuse the VQ-VAE to calculate two feature losses, which help improve structure coherence and texture realism, respectively.

The authors first train the hierarchical VQ-VAE and, afterward, the diverse structure generator ($G_s$ depending on parameters $\theta$) and the texture generator ($G_t$ depending on parameters $\varphi$) are trained separately. These generators are later on used for inference. 
The structure generator $G_s$ is constructed via a conditional autoregressive network
for the distribution over structural features. In inference, it will generate different structural features via sampling. Its objective loss is defined as the negative log-likelihood
\begin{equation}\label{eq:DSIVQVAEgenstr}
\mathcal{L}_{\ell}(\theta)= -{\mathbb{E}}_{x_{\mathrm{gt}}\sim\mathbb{P}_{\mathcal{X}}}[\log (p_\theta(s_{\mathrm{gt}}|y,M)]
\end{equation}
where $y$ is the input image to be inpainted on the points of $O$ where the hole mask $M$ is equal to $1$,  $\mathbb{P}_{\mathcal{X}}$ denotes the distribution of the training dataset,  $s_{\mathrm{gt}}$ denote the vector quantized structural features of the ground truth at the coarser scale given by the hierarchical VQ-VAE, and $\theta$ the parameters of $G_s$.

Besides, the objective loss for the texture generator $G_t$ is composed by: (i) the L$^{1}$ norm comparing the inpainted solution to the ground truth at pixel level, (ii) an adversarial loss using the discriminator trained with the SN-PatchGAN hinge version \cite{YuLinYangShenLuHiang2018free} 
applied to the resulting image and, moreover, (iii) a structural feature loss $\mathcal{L}_{ts}(\varphi)$ and (iv) a textural feature loss $\mathcal{L}_{tt}(\varphi)$. These last two losses are defined similarly using a multi-class cross-entropy loss. In particular, the structural feature loss is defined as
\begin{equation}
    \mathcal{L}_{ts}(\varphi)= -\sum_{k,j} \alpha_{k,j}\log\left(\mathrm{softmax}(10\, \delta_{k,j})\right),
\end{equation}
where $\delta_{k,j}$ denotes the truncated distance
similarity score between the k-th feature vector of $s_{\mathrm{comp}}$ (computed from the inpainted image using the trained encoder) and the j-th prototype vector of the structural codebook of VQ-VAE. $\alpha_{k,j}$ is an indicator of the prototype vector class. That is, $\alpha_{k,j}=1$ when the k-th feature vector of $s_{\mathrm{gt}}$,  belongs to the j-th class of the structural codebook; otherwise, $\alpha_{k,j}=0$. The authors define the textural feature loss $\mathcal{L}_{tt}(\varphi)$ in an analogous way.

As mentioned, in Section \ref{sec:results} we will experimentally analyze this method. It will be denoted there as DSI-VQVAE.

\subsection{Image Transformers}\label{sec:transformers}

Self-attention-based architectures, in particular Transformers \cite{vaswani2017attention} are well explored architectures in Natural Language Processing (NLP). Transformers use a self-attention mechanism to model long range relationships between the elements of an input sequence (for instance, in a text) that has shown to be more efficient than Recurrent Neural Networks. They have achieved state-of-the-art results in several tasks not only in the field of NLP but also more recently for computer vision problems. The vanilla transformer \cite{vaswani2017attention} and its variants have been successfully applied in computer vision to, \emph{e.g.}~inpainting \cite{wan2021high,yu2021diverse}, object detection \cite{carion2020end}, image classification \cite{dosovitskiy2020image}, colorization \cite{kumar2021colorization}, super resolution \cite{yang2020learning}.

Instead of using inductive local biases 
like CNNs, transformers in imaging aim to have a global receptive field. For this, the image is first transformed by, as in the most basic approach, flattening the spatial dimensions of the input feature map into a sequence of features of size $M\times N\times F$, where $M\times N$ represents the flattened spatial dimensions and $F$ the depth of a feature map. Then self-attention is applied over the extracted sequence. To ease the associated high computational cost, some authors substitute spatial pixels by patches.  
The attention mechanism looks at the input sequence and decides for each position which other parts of the sequence or image are important. 
More specifically, the transformers will transform the set of inputs, called \emph{tokens}, using sequential blocks of multi-headed self-attention, which relate embedded inputs to each other. 
It is worth notice, that transformers will maintain the number of tokens throughout all computations. If tokens were related to pixels, each pixel would have a one-to-one correspondence with the output, thus, maintaining the spatial resolution of the original input image. Since transformers are set-to-set functions, they do not intrinsically retain the information of the spatial position for each individual token, thus the embedding is concatenated to a learnable position embedding to add the positional information to the representation.

One advantage of using a transformer for image restoration is that it naturally supports pluralistic outputs by directly optimizing the underlying data distribution. One drawback is the computational complexity that increases quadratically with the input length, thus making it difficult to directly synthesize high-resolution images.

\bigskip

\noindent\textbf{High-Fidelity Pluralistic Image Completion with Transformers  \cite{wan2021high}}
\bigskip

The authors of \cite{wan2021high} exploit the benefits of both, transformers and CNNs. The use of transformers will enforce a global structural understanding and pluralism support in the inpainted region, at a coarse resolution. On the other hand, the use of CNNs will allow working with high-resolution images without a high computational cost due to its capacity of estimating local textures efficiently.

\begin{figure}[h!]
    \centering
    \includegraphics[width=\textwidth]{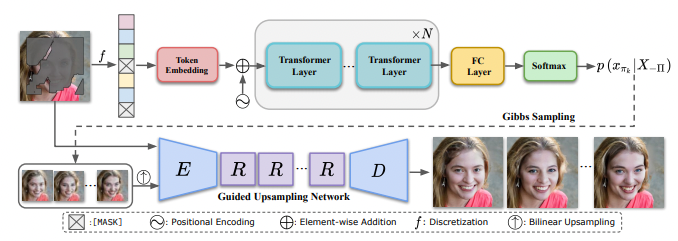}
    \caption{Overview of the architecture of High-Fidelity Pluralistic Image Completion with Transformers \cite{wan2021high}, referred to as ICT. Figure from \cite{wan2021high}.}
    \label{fig:ICT}
\end{figure}

Concretely, in this work, image completion is performed in two steps as shown in Figure \ref{fig:ICT}. In the first step, given a corrupted image, the authors use transformers to produce the probability distribution of structural appearance of complete images given the incomplete one. Low-resolution results can be obtained by sampling from this distribution with diversities that recover pluralistic coherent image structures. In the second step, guided by the computed image structures together with the available pixels of the input image, another upsampling CNN model is used to render high fidelity textures for missing regions meanwhile ensuring coherence with neighboring pixels.

If $X_{\Pi}$ denotes the set of masked tokens $x_{\pi_k}$ (where $\Pi=\{\pi_1,\dots,\pi_K\}$ denote their indexes), and $X_{-\Pi}$ denotes the set of unmasked tokens (corresponding to the visible regions), then the transformer is optimized by minimizing the negative log-likelihood of the masked tokens $x_{\pi_k}$, conditioned  the visible regions $X_{-\Pi}$, that is,
\begin{equation}\label{eq:MLM}
   \mathcal{L}_{\text{MLM}}(\theta) = \mathbb{E}_{X} \frac{1}{K}\sum_{k=1}^{K}  -\log p (x_{\pi_k} | X_{-\Pi}; \theta)
\end{equation}
where $\theta$ contains the parameters of the transformer, and the subindex MLM stands for the \emph{masked language model} which is similar to the one in BERT \cite{devlin2018bert}.
One particularity of the ICT model is that each token attends simultaneously to all positions thanks to bi-directional attention. This enables the generated distribution to
capture the full context, thus leading to a consistency between generated contents and unmasked region.

Once the transformer is trained, instead of directly sampling the entire set of masked positions which would lead to non-plausible results due to the independence property, they apply Gibbs sampling to iteratively sample tokens at different locations. To do so,
in each iteration, a grid position is sampled from
$p (x_{\pi_k}| X_{-\Pi}, X_{<\pi_k},\theta)$ with the top-${\mathcal{K}}$ predicted elements, where $X_{<\pi_k}$ denotes the previous generated tokens. 

The second step is to perform texture refinement at the original resolution using a CNN, which is optimized by minimizing the  L$^{1}$  loss between the predicted image and the ground truth, together with an adversarial loss based on the vanilla GAN (cf. \eqref{eq:vanillaGAN} in Section \ref{ssec:GAN}).

\bigskip

\noindent\textbf{Diverse image inpainting with bidirectional and autoregressive transformers \cite{yu2021diverse}}

\bigskip

\begin{figure}
    \centering
    \includegraphics[width=\textwidth]{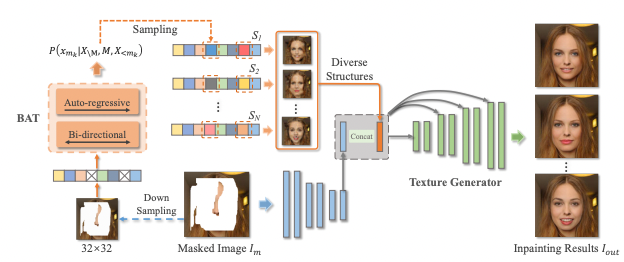}
    \caption{Overview of the BAT architecture of Diverse image inpainting with bidirectional and autoregressive transformers \cite{yu2021diverse}. Figure from \cite{yu2021diverse}.}
    \label{fig:BAT}
\end{figure}

This proposal exploits, as in \cite{wan2021high}, a two steps strategy where transformers will encode global structure understanding and high-level semantics at a first stage, followed by a CNN-based generation of additional texture.  While \cite{wan2021high} leverages bi-directional attention with the masked language model (MLM) as in BERT \cite{devlin2018bert}, the authors of \cite{yu2021diverse} propose BAT-Fill, that combines autoregressive models and bidirectional models (cf. Figure \ref{fig:BAT}). 
The first transformer-based step estimates the distribution of inpainted low-resolution structures from which to sample, from an input damaged image, a set $\{s_1,\dots,s_J\}$ of plausible complete structures.
Instead of only using a masked language model  like BERT and \cite{wan2021high} (see above), that use bidirectional contextual information but predicts each masked token separately and independently (which can result in inconsistency in the generated result),  BAT-Fill incorporates autoregressive modeling (factorizing the predicted tokens with the product rule). The input sequence of tokens is sorted by first having the visible tokens (permuted) and then the missing tokens (with the original order). In this way, the autoregressive model starts at the position of the first missing pixel. The BAT training objective is given by
\begin{equation}
   \mathcal{L}_{\text{BAT}}(\theta) = \mathbb{E}_{X} \frac{1}{K}\sum_{k=1}^{K} - \log p (x_{\pi_k}|X_{-\Pi}, M, X_{<\pi_{k}}; \theta).
\end{equation}
where we have used the same notations as in \eqref{eq:MLM}, namely, $K$ is the length of of masked tokens, $X_{-\Pi}$ are all the unmasked tokens (corresponding to the visible regions). Finally,  $ X_{<\pi_{k}}$ denote the previous predicted tokens, and $M$ the masked positions. 

Finally, they construct a  texture generator based on  CNN-based synthesis, which is optimized by minimizing the L$^1$ loss between the predicted image and the ground truth together with an adversarial loss and a perceptual loss~\cite{johnson2016perceptual}. 

In inference, each masked token is predicted bidirectionally and autoregressively. As in \cite{wan2021high}, they iteratively use top-${\mathcal{K}}$ sampling to randomly sample from the ${\mathcal{K}}$ most likely next tokens.

\section{From Single Image Evaluation Metrics to Diversity Evaluation} \label{sec:evalmetrics}

Currently, there is no consensus on automatic evaluation methods for single or diverse inpainting.
As the problem is to recover a visually plausible image, performing quantitative evaluation is not trivial as the solution is not unique and the plausibility is a subjective term. Nevertheless, several evaluation metrics have been proposed through the years. We first detail those used for evaluating inpainting methods, one image at the time, before presenting the metric used as diversity scores.

Full reference metrics compare the ground truth image with the inpainted result. Famous measures in this category include L$^1$, L$^2$ distances, PSNR, or SSIM (Wang et al., 2004). These metrics analyze the ability of the model to reconstruct the original image content.  Nevertheless, it is easy to demonstrate that they do not well characterize the realism of an image. Being close to the ground truth image does not ensure being realistic.
Other perceptual metrics have been proposed and are supposed to be more consistent with human judgment. In particular,  Learned Perceptual Image Patch Similarity (LPIPS)~\cite{Zhang_2018_CVPR} has been demonstrated to correlate well with the human perceptual similarity. It relies on the observation that hidden activations in CNNs trained for image classification are indeed a space where distance can strongly correlate with human judgment. Precisely, LPIPS computes a weighted L$^2$ norm between deep features of pair of images:
\begin{equation}
    \rm{LPIPS}(x,x_{gt}) = \sum_l\frac{1}{M_lN_l}\sum_{ij}\|w_l\odot (\Phi_r^l(i,j) - \Phi_{gt}^l(i,j))\|_2^2
\end{equation}
where $x$ is the reconstructed image, $x_{gt}$ is the ground truth, $l$ is a layer number, $(i,j)$ a pixel, $w_l$ are weights for each features, and $\Phi^l$ and $\Phi_{gt}^l \in \mathbb{R}^{M_l\times N_l \times C_l}$ are features unit-normalized in the channel dimension.
LPIPS has been used in the context of inpainting when generating one image (\textit{e.g.} \cite{zheng2021tfill}). In \cite{elpips}, it was shown that standard adversarial attack techniques can easily fool LPIPS. Therefore a slightly different metric called E-LPIPS (Ensemble LPIPS) is proposed by applying random simple image transformations and dropout. Nonetheless, up to our knowledge, it has never been used in the context of inpainting.   \\

When, apart from the set of images, there is available corresponding image categories, other metrics, that are also supposed to be following human judgment, can be used.
The inception score (IS)~\cite{salimans2016improved} was designed to measure how realistic the output from a GAN is. This score measures the variety of a set of generated images as well as the probability distribution of each image classification. This is done by comparing the class distribution of each image, which should have a low entropy, with the marginal distribution of the whole set, which should have high entropy:
\begin{equation}
\rm{IS(G)} = \exp\left(\mathbb{E}_{x\sim p_g}\rm{KL}\left(p(y|x) || p(y)\right)\right)
\end{equation}
where $p_g$ is the model distribution of the whole set given by the generative model $G$, $x$ an image sampled from $p_g$, $p(y|x)$ the conditional class distribution, $\rm{KL}$ the Kullback-Leibler  divergence and $p(y)$ the marginal class distribution. As detailed in \cite{barratt2018note},  inception score has it own limitations:  sensitivity to small changes in network weights,  misleading results when used beyond the ImageNet dataset~\cite{rosca2017variational}, adversarial examples when used for model optimization. The IS score was adapted to diverse inpainting in~\cite{zhao2020uctgan}, leading to the Modified Inception Score (MIS). When performing inpainting, there is only one kind of image and so $p(y)$ can be removed.  The MIS is then defined as
\begin{equation}\label{eq:MIS}
\rm{MIS(G)} = \exp\left(\mathbb{E}_{x\sim p_g}\sum_i \left(p(y_i|x) \log p(y_i|x)\right)\right),
\end{equation}
where $y_i$ of is the class label of the $i^{th}$ generated sample. 
Another improvement of the IS is the Fréchet Inception Distance (FID)~\cite{heusel2017gans} that compares the statistics of generated images to the ones of original images. FID uses the Inception pre-trained model to extract the feature vectors of real images and fake images, and compare their feature-wise means ($\mu_r$, $\mu_f$) and covariances ($\Sigma_r$, $\Sigma_f$):
\begin{equation}\label{eq:FID}
\rm{FID} =\lVert \mu_r-\mu_f \rVert^2 + Tr(\Sigma_r+\Sigma_f+2(\Sigma_r\Sigma_f)^{1/2}).
\end{equation}
Fréchet Inception Distance has been widely used for validating single and diverse inpainting results in recent papers (\textit{e.g.} \cite{peng2021generating, liu2021pd,yu2021diverse}).

\bigskip

\noindent
\textbf{Measuring diversity} 

\bigskip

In the context of pluralistic inpainting, following the idea proposed for image-to-image translation in \cite{zhu2017toward}, LPIPS has been used as a \textit{diversity score} to measure how perceptually different the generated images are~\cite{cai2020piigan, zhao2020uctgan, liu2021pd}. 
The higher the LPIPS, the more diversity is present in the results. For instance, in \cite{cai2020piigan}, they compute the average distance between the 10000 pairs randomly generated from the 1000 center-masked image samples. 
LPIPS is computed on the full inpainting results and mask-region inpainting results, respectively.

\begin{table}[!ht]
\caption{Generative methods for diverse inpainting: experimental conditions. Random regular and irregular masks are generated as in \cite{zheng2019pluralistic}.}
\label{tab:methods_exp}  
\centering
\begin{tabular}{p{1.7cm}p{1.6cm}p{2.6cm}p{4.5cm}P{0.55cm}}
\toprule
\textbf{Method} &  \textbf{Input Size} & \textbf{Train Datasets} & \textbf{Training Masks} & \textbf{Code}  \\
\specialrule{\heavyrulewidth}{1pt}{1pt}
PIC   & $ 256 \times 256 $ & - Celeba-HQ  \newline - ImageNet \newline - Paris \newline - Places2  & - Regular (center $128 \times 128$ + random) \newline - Irregular (random)  & \cmark \\
\noalign{\smallskip}\hline\noalign{\smallskip}
PiiGAN  & $128 \times 128$ & - CelebA \newline - Mauflex \newline - Agricultural Disease &  - center $64 \times 64$ & \cmark \\
\noalign{\smallskip}\hline\noalign{\smallskip}
UCTGAN  &$256 \times 256$ & - Celeba-HQ  \newline - ImageNet \newline - Paris \newline - Places2 &  - Regular (center $128 \times 128$  + random) \newline - Irregular (random) & \xmark \\
\noalign{\smallskip}\hline\noalign{\smallskip}
DSI-VQVAE & $ 256 \times 256 $ & - Celeba-HQ  \newline - ImageNet \newline - Places2  & - Regular (center $128 \times 128$  + random) \newline - Irregular (random) & \cmark  \\
\noalign{\smallskip}\hline\noalign{\smallskip}
ICT& $ 256 \times 256 $ & - FFHQ \newline  - ImageNet \newline - Places2  & - Irregular Pconv~\cite{liu2018image}  & \cmark \\
\noalign{\smallskip}\hline\noalign{\smallskip}
PD-GAN &  $256 \times 256 $  &  - Celeba-HQ  \newline - Paris StreetView \newline - Places2    & - Irregular Pconv~\cite{liu2018image} &  \xmark \\
\noalign{\smallskip}\hline\noalign{\smallskip}
BAT& $ 256 \times 256 $ & - CelebA-HQ \newline  - Paris StreetView \newline - Places2  & - Irregular Pconv~\cite{liu2018image}  &\cmark \\
\bottomrule
\end{tabular}
\end{table}

\section{Experimental Results}\label{sec:results}

In this section, we present a quantitative and qualitative comparison of several existing methods for multiple image inpainting. 
We include an assessment of both the quality and the diversity of the inpainted solutions.
All the results shown in this section are thanks to publicly available code together with pre-trained weights provided by the authors.
In Table \ref{tab:methods_exp}, we summarize, for all the methods previously reviewed, the conditions in which the experiments were conducted.
Note that, among these methods, only PIC (\cite{zheng2019pluralistic}), PiiGAN (\cite{cai2020piigan}), DSI-VQVAE (\cite{peng2021generating}), ICT (\cite{wan2021high}) and BAT (\cite{yu2021diverse}) provide source code and pre-trained models. In the rest of this section, we describe the experimental settings in Section \ref{subsec:experi} including datasets and used masks, quantitative results in Section \ref{subsec:QuantitativeAnalysis} including proximity to ground truth, perceptual quality, and inpainting diversity, and finally a qualitative analysis is provided in Section \ref{subsec:QualitativeAnalysis}. 

\subsection{Experimental Settings}\label{subsec:experi}



Table \ref{tab:methods_exp} lists all the explained methods together with the training dataset and corresponding training masks.
Aiming for a fair comparison, we compare and test the methods trained on the same training images \emph{i.e.} the VAE-based model PIC (\cite{zheng2019pluralistic}), the VQVAE-based model DSI-VQVAE (\cite{peng2021generating}) and the two transformer-based models ICT (\cite{wan2021high}) and BAT (\cite{yu2021diverse}). Notice that we do not analyze the performance of PiiGAN (\cite{cai2020piigan}), as the training datasets and size images are different from the other methods. 

\bigskip
\noindent\textbf{Datasets}
\bigskip

We evaluate the methods on the three datasets Celeba-HQ \cite{karras2018progressive}, Places2 \cite{zhou2017places}, and ImageNet \cite{russakovsky2015imagenet}. 
All the evaluated models take as input images of resolution $256 \times 256$. 
Due to the long inference time of DSI-VQVAE and ICT methods (see Table~\ref{tab:runningT}), quantitative experiments are made on 100 randomly selected images from each training dataset. For each kind of mask (see below) and for each image, we sample 25 different results. 

For Celeba-HQ, the $1024\times1024$ resolution images are resized to  $256\times256$.
For Places2 and ImageNet, the compared methods were trained on $256\times256$ patches either by resizing the input images (PIC), either by cropping them, randomly (DSI), or to the center patch (BAT), either by both cropping and resizing (ICT). We will both consider center-cropped and resized versions of the input images to ensure a fair comparison among the trained models.

Note that ICT is not trained on Celeba-HQ but on the FFHQ face dataset \cite{karras2019style}. FFHQ contains higher variation then Celeba-HQ in terms of age, ethnicity and image
background. It also has a good coverage of accessories. Images from both datasets are however similarly aligned and cropped. Therefore, we still give the results of the ICT method tested on the Celeba-HQ dataset but the reader should remember this difference when analysing the results. 

\bigskip
\noindent\textbf{Inpainting Masks}
\bigskip

\begin{figure}[h!]
    \centering
\fbox{\includegraphics[width=0.12\textwidth]{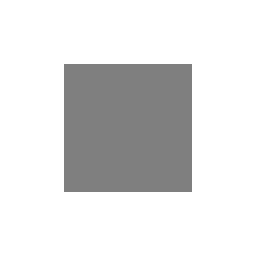}}
\fbox{\includegraphics[width=0.12\textwidth]{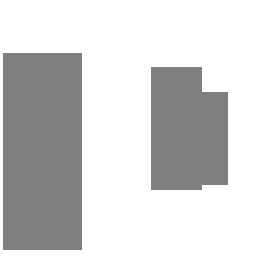}}
\fbox{\includegraphics[width=0.12\textwidth]{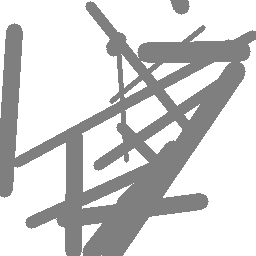}}
\fbox{\includegraphics[width=0.12\textwidth]{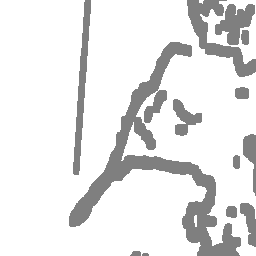}}
\fbox{\includegraphics[width=0.12\textwidth]{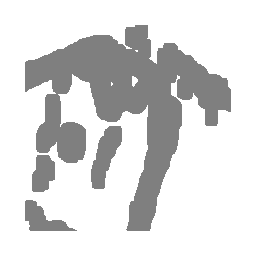}}
\fbox{\includegraphics[width=0.12\textwidth]{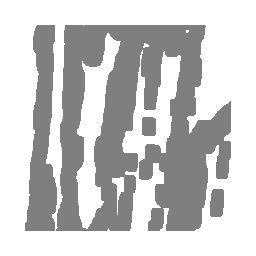}}
\caption{Example for each kind of mask considered for evaluation. In grey the hidden pixels. From left to right: center, random regular, random irregular, and irregular Pconv masks from \cite{liao2018edge} with $<20\%$, $[40\%,60\%]$ and $[40\%,60\%]$ hidden pixels.}
\label{fig:ex_masks}
\end{figure}

We use the following type of masks: center, random regular, random irregular and irregular masks from \cite{liu2018image} with several different proportions of hidden pixels. Figure \ref{fig:ex_masks} shows an example of each kind of mask.
The random masks are generated once for each test image so that all the method are evaluated on the same degradation. 

We would like to highlight that the methods PIC and DSI-VQVAE train a different model for regular and irregular holes. 
Testing on center or random regular mask is realized with the former model and testing on irregular masks with the latter. The transformer-based methods ICT and BAT only train on "irregular Pconv" holes given by \cite{liu2018image}. Testing on each type of mask will be done with this unique model. 

\subsection{Quantitative Performance} \label{subsec:QuantitativeAnalysis}

We first analyse the numerical performance of each method. Table \ref{tab:comparisonCeleb} shows quantitative results on Celeba-HQ dataset.  
Additionally, results on Places2 and ImageNet are respectively shown in Tables \ref{tab:comparisonPlaces} and \ref{tab:comparisonImageN}. 

In Tables \ref{tab:comparisonPlaces} and \ref{tab:comparisonImageN} we give our results obtained by center-cropping the images on Places2 and ImageNet, respectively. 
For fair comparison, we also give in Appendix (Tables \ref{tab:comparisonPlacesRe} and  \ref{tab:comparisonImageNRe}), the results on these two datasets for resized images. The ICT method is runned, as proposed in the original paper, with its top-${\mathcal{K}}$ parameter (cf. Section \ref{sec:transformers}) set to $50$. We investigate the influence of the top-${\mathcal{K}}$ parameter in Table \ref{tab:topk}.

To measure inpainting quality, we take into account three factors: the similarity to the ground truth, the realism of inpainting outputs, and the diversity of those outputs. Definitions and details on the metrics for each factor can be found in Section \ref{sec:evalmetrics}. 
Note that contrary to \cite{zheng2019pluralistic}, we do not use here any discriminator score to select the best samples before evaluation. 

In each table, the best and second-best results by column are in bold and underlined, respectively. 

\begin{table}[ht]
\caption{Quantitative comparison of four pluralistic image inpainting methods (PIC, DSI-VQVAE, ICT and BAT) on \textbf{Celeba-HQ} and for different kind of masks (central, random regular, random irregular and from \cite{liu2018image}). 
Best and second best results by column are in bold and underlined, respectively.}
\label{tab:comparisonCeleb}
        \centering
               \begin{tabular}{P{1.5cm}P{2.3cm}P{0.85cm}P{0.85cm}P{0.7cm}P{0.35cm}P{0.85cm}P{0.85cm} P{1.7cm}}        
        \toprule
         & & \multicolumn{3}{c}{\textbf{Similarity to GT}} & & \multicolumn{2}{c}{\textbf{Realism}}  & \textbf{Diversity}\\
        \textbf{Mask} & \textbf{Method}  & PSNR$\uparrow$ & SSIM$\uparrow$ & L$^1$ $\downarrow$  & & MIS$\uparrow$  & FID$\downarrow$ &  LPIPS$\uparrow$ \\
        \toprule
          \multirow{4}{\linewidth}{Irregular $<20\%$} & PIC & $34.63$ & $0.964$ & $\mathbf{1.17}$ &  & $0.0206$ & $16.8$ & $0.0009$  \\
           & DSI-VQVAE & $\underline{35.49}$ & $\underline{0.968}$ & $1.41$ &  & $\mathbf{0.0216}$ & $11.0$ & $\underline{0.0081}$  \\
         & ICT & $34.72$ & $\underline{0.968}$ & $2.09$  & & $0.0200$ & $\mathbf{9.84}$ & $\mathbf{0.0084}$  \\
         & BAT & $\mathbf{36.25}$ & $\mathbf{0.974}$ & $\underline{1.20}$ &  & $\underline{0.0208}$ & $\underline{9.90}$ & $0.0056$  \\
        \midrule
          \multirow{4}{\linewidth}{Irregular $20\%-40\%$} & PIC & $26.69$ & $0.879$ & $4.19$ & & $\underline{0.0216}$ & $34.2$ & $0.0091$  \\
          & DSI-VQVAE & $\mathbf{27.36}$ & $0.888$ & $\underline{4.06}$ & & $\mathbf{0.0223}$ & $28.8$ & $\underline{0.0357}$  \\
         & ICT & $26.83$ & $\underline{0.891}$ & $4.71$ & & $0.0189$ & $\underline{26.7}$ & $\mathbf{0.0383}$  \\
         & BAT & $\underline{27.28}$ & $\mathbf{0.900}$ & $\mathbf{3.85}$ & & $0.0214$ & $\mathbf{20.7}$ & $0.0269$  \\
        \midrule
         \multirow{4}{\linewidth}{Irregular $40\%-60\%$} & PIC & $21.47$ & $0.745$ & $10.36$ & & $0.0153$ & $65.4$ & $0.0527$  \\
         & DSI-VQVAE & $\mathbf{22.53}$ & $0.770$ & $\underline{9.01}$ & & $\underline{0.0156}$ & $51.9$ & $\underline{0.0916}$  \\
         & ICT & $21.92$ & $\underline{0.773}$ & $9.82$ & & $0.0153$ & $\underline{50.7}$ & $\mathbf{0.0970}$  \\
         & BAT & $\underline{22.35}$ & $\mathbf{0.787}$ & $\mathbf{8.91}$ & & $\mathbf{0.0183}$ & $\mathbf{39.7}$ & $0.0731$  \\
         \midrule
                 \multirow{4}{\linewidth}{Central $128 \times 128$} 
        & PIC & $24.46$ & $0.868$ & $5.26$ & & $\underline{0.0212}$ & $23.8$ & $0.0288$  \\
         & DSI-VQVAE & $\mathbf{25.25}$ & $\underline{0.880}$ & $\mathbf{5.08}$ & & $0.0210$ & $\underline{21.7}$ & $0.0243$  \\
         & ICT & $24.45$ & $0.872$ & $6.06$ &  & $0.0170$ & $27.3$ & $\mathbf{0.0486}$  \\
         & BAT & $\underline{25.10}$ & $\mathbf{0.882}$ & $\underline{5.21}$ & & $\mathbf{0.0218}$ & $\mathbf{21.5}$ & $\underline{0.0365}$  \\
        \midrule
          \multirow{4}{\linewidth}{Random regular} & PIC & $24.16$ & $0.840$ & $7.23$ & & $0.0188$ & $33.4$ & $0.0402$  \\
         & DSI-VQVAE & $\mathbf{24.98}$ & $0.850$ & $\mathbf{6.46}$ & & $\underline{0.0200}$ & $\underline{30.5}$ & $\underline{0.0642}$  \\
         & ICT & $24.51$ & $\underline{0.852}$ & $7.24$ & & $0.0180$ & $31.3$ & $\mathbf{0.0665}$  \\
         & BAT & $\underline{24.85}$ & $\mathbf{0.860}$ & $\underline{6.52}$ & & $\mathbf{0.0209}$ & $\mathbf{24.6}$ & $ 0.0541$  \\
       \midrule
          \multirow{4}{\linewidth}{Random irregular} &  PIC & $23.47$ & $0.759$ & $8.45$ & & $0.0161$ & $73.5$ & $0.0280$  \\
         & DSI-VQVAE & $\underline{24.27}$ & $\underline{0.785}$ & $\underline{7.56}$ & & $\underline{0.0167}$ & $58.8$ & $\underline{0.0744}$  \\
         & ICT & $23.26$ & $0.781$ & $9.26$ & & $0.0148$ & $\underline{52.2}$ & $\mathbf{0.0855}$  \\
         & BAT & $\mathbf{24.36}$ & $\mathbf{0.810}$ & $\mathbf{7.13}$ & & $\mathbf{0.0186}$ & $\mathbf{40.8}$ & $0.0495$  \\
        \specialrule{\heavyrulewidth}{1pt}{1pt}
         \multirow{4}{\linewidth}{Average} 
         & PIC & $25.65$ & $0.843$ & $6.11$ & & $0.0189$ & $41.2$ & $0.0266$ \\
         & DSI-VQVAE & $\underline{26.65}$ & $\underline{0.857}$ & $\underline{5.60}$ & & $\underline{0.0195}$ & $33.8$ & $\underline{0.0497}$ \\
         & ICT & $25.95$ & $0.855$ & $6.53$ & & $0.0173$ & $\underline{33.0}$ & $\textbf{0.0575}$ \\
         & BAT & $\textbf{26.70}$ & $\textbf{0.869}$ & $\textbf{5.47}$ & & $\textbf{0.0203}$ & $\textbf{26.2}$ & $0.0410$ \\
         \bottomrule
        \end{tabular}
\end{table}

\begin{table}[ht]
\caption{Quantitative comparison of four pluralistic image inpainting methods (PIC, DSI-VQVAE, ICT and BAT) on $256 \times 256$ \textbf{center-cropped} images from \textbf{Places2}, for different kind of masks (central, random regular, random irregular and from \cite{liu2018image}).} 
\label{tab:comparisonPlaces}
        \centering
               \begin{tabular}{P{1.5cm}P{2.3cm}P{0.85cm}P{0.85cm}P{0.7cm}P{0.35cm}P{0.85cm}P{0.85cm} P{1.7cm}}             
        \toprule
         & & \multicolumn{3}{c}{\textbf{Similarity to GT}} & & \multicolumn{2}{c}{\textbf{Realism}}  & \textbf{Diversity}\\
        \textbf{Mask} & \textbf{Method}  & PSNR$\uparrow$ & SSIM$\uparrow$ & L$^1$ $\downarrow$  & & MIS$\uparrow$  & FID$\downarrow$ &  LPIPS$\uparrow$ \\
        \toprule
        
        \multirow{4}{\linewidth}{Irregular $<20\%$} 
        &  PIC & $30.48$ & $0.937$ & $\underline{2.02}$ &  & $\textbf{0.0507}$ & $36.8$ & $0.0050$ \\
         & DSI-VQVAE & $\underline{31.58}$ & $\underline{0.952}$ & $2.11$ &  & $\underline{0.0482}$ & $\underline{19.3}$ & $\underline{0.0187}$ \\
        & ICT & $29.86$ & $0.943$ & $3.64$ &  & $0.0463$ & $22.8$ & $\textbf{0.0198}$ \\
        & BAT & $\textbf{32.20}$ & $\textbf{0.957}$ & $\textbf{1.83}$ & & $0.0463$ & $\textbf{14.2}$ & $0.0158$ \\
         \midrule
        \multirow{4}{\linewidth}{Irregular $20\%-40\%$} 
        &  PIC & $23.88$ & $0.820$ & $6.46$ &  & $0.0378$ & $97.6$ & $0.0344$ \\
         & DSI-VQVAE  & $\textbf{24.20}$ & $\underline{0.844}$ & $\textbf{6.14}$ &  & $\textbf{0.0438}$ & $\underline{63.6}$ & $\underline{0.0707}$ \\
        & ICT  & $23.08$ & $0.831$ & $8.05$ &  & $\underline{0.0428}$ & $70.0$ & $\textbf{0.0769}$ \\
        & BAT & $\underline{24.10}$ & $\textbf{0.853}$ &  $\textbf{6.14}$ & & $0.0423$ & $\textbf{53.2}$ & $0.0671$ \\
        \midrule
        \multirow{4}{\linewidth}{Irregular $40\%-60\%$}
        & PIC & $19.92$ & $0.667$ & $13.75$ & & $0.0326$ & $156.1$ & $0.1309$ \\
        & DSI-VQVAE & $\textbf{20.34}$ & $\underline{0.703}$ & $\textbf{12.52}$ &  & $\textbf{0.0398}$ & $\underline{110.2}$ & $0.1566$ \\
        & ICT & $19.49$ & $0.686$ & $14.66$ &  & $\underline{0.0371}$ & $128.7$ & $\textbf{0.1668}$ \\
        & BAT & $\underline{19.98}$ & $\textbf{0.705}$ & $\underline{13.10}$ &  & $0.0364$ & $\textbf{107.0}$ & $\underline{0.1610}$ \\
         \midrule
          \multirow{4}{\linewidth}{Central $128 \times 128$} 
        & PIC & $20.98$ & $0.812$ & $9.00$ &  & $0.0435$ & $96.8$ & $0.1080$ \\
        & DSI-VQVAE & $\textbf{21.41}$ & $\underline{0.819}$ & $\underline{8.85}$ & & $0.0416$ & $\textbf{79.8}$ & $\textbf{0.1234}$ \\
        & ICT & $20.93$ & $0.812$ & $10.22$ &  & $\textbf{0.0476}$ & $92.2$ & $\underline{0.1204}$ \\
        & BAT & $\underline{21.20}$ & $\textbf{0.822}$ & $\textbf{8.76}$ &  & $\underline{0.0442}$ & $\underline{81.8}$ & $0.1190$ \\
         \midrule
        \multirow{4}{\linewidth}{Random regular} 
        & PIC & $21.70$ & $0.783$ & $10.14$ &  & $\underline{0.0425}$ & $103.8$ & $0.1124$ \\
        & DSI-VQVAE & $\textbf{22.36}$ & $\underline{0.805}$ & $\underline{9.21}$ &  & $0.0412$ & $\textbf{75.8}$ & $0.1167$ \\
        & ICT & $21.75$ & $0.796$ & $10.77$ &  & $0.0405$ & $87.1$ & $\textbf{0.1237}$ \\
        & BAT & $\underline{22.34}$ & $\textbf{0.808}$ & $\textbf{9.15}$ &  & $\textbf{0.0436}$ & $\underline{76.6}$ & $\underline{0.1200}$ \\
         \midrule
        \multirow{4}{\linewidth}{Random irregular} 
        &  PIC & $\underline{20.86}$ & $0.658$ & $12.80$ &  & $0.0255$ & $165.4$ & $0.0979$ \\
        & DSI-VQVAE & $\textbf{21.18}$ & $\underline{0.701}$ & $\textbf{11.78}$ &  & $\underline{0.0360}$ & $\underline{114.4}$ & $0.1450$ \\
        & ICT & $20.07$ & $0.681$ & $14.14$ &  & $0.0334$ & $131.9$ & $\textbf{0.1548}$ \\
        & BAT & $20.85$ & $\textbf{0.708}$ & $\underline{12.00}$ &  & $\textbf{0.0374}$ & $\textbf{103.2}$ & $\underline{0.1454}$ \\ 
         \specialrule{\heavyrulewidth}{1pt}{1pt}
        \multirow{4}{\linewidth}{Average}
        & PIC & $22.97$ & $0.780$ & $9.02$ &  & $0.0388$ & $109.2$ & $0.0814$ \\
        & DSI-VQVAE & $\textbf{23.51}$ & $\underline{0.804}$ & $\textbf{8.44}$ & & $\textbf{0.0418}$ & $\underline{76.9}$ & $\underline{0.1052}$ \\
        & ICT & $22.53$ & $0.792$ & $10.25$ &  & $0.0413$ & $88.9$ & $\textbf{0.1107}$ \\
        & BAT & $\underline{23.44}$ & $\textbf{0.809}$ & $\underline{8.97}$ &  & $\underline{0.0417}$ & $\textbf{72.7}$ & $0.1047$ \\
        \bottomrule
        \end{tabular}
\end{table}

\begin{table}[ht]
\caption{Quantitative comparison of three pluralistic image inpainting methods (PIC, DSI-VQVAE and ICT) on $256 \times 256$ \textbf{center-cropped} images from \textbf{ImageNet}, for different kind of masks (central, random regular, random irregular and from \cite{liu2018image}).} 
\label{tab:comparisonImageN}
        \centering
               \begin{tabular}{P{1.5cm}P{2.3cm}P{0.85cm}P{0.85cm}P{0.7cm}P{0.35cm}P{0.85cm}P{0.85cm} P{1.7cm}}             
        \toprule
         & & \multicolumn{3}{c}{\textbf{Similarity to GT}} & & \multicolumn{2}{c}{\textbf{Realism}}  & \textbf{Diversity}\\
        \textbf{Mask} & \textbf{Method}  & PSNR$\uparrow$ & SSIM$\uparrow$ & L$^1$ $\downarrow$  & & MIS$\uparrow$  & FID$\downarrow$ &  LPIPS$\uparrow$ \\
        \toprule
        \multirow{3}{\linewidth}{Irregular $<20\%$} 
        & PIC & $\underline{30.33}$ & $\underline{0.941}$ & $\textbf{2.02}$ &  & $\textbf{0.2416}$ & $20.2$ & $0.0036$ \\
         & DSI-VQVAE & $\textbf{30.44}$ & $\textbf{0.946}$ & $\underline{2.38}$ &  & $\underline{0.2361}$ & $\underline{12.1}$ & $\textbf{0.0199}$ \\
        & ICT & $29.23$ & $0.940$ & $3.98$ &  & $0.2323$ & $\textbf{10.7}$ & $\underline{0.0185}$ \\
       \midrule
        \multirow{3}{\linewidth}{Irregular $20\%-40\%$} 
        & PIC & $\textbf{23.02}$ & $0.797$ & $\textbf{7.37}$ &  & $0.1709$ & $83.7$ & $0.0289$ \\
        & DSI-VQVAE & $\underline{22.98}$ & $\textbf{0.809}$ & $\underline{7.56}$ &  & $\textbf{0.2015}$ & $53.4$ & $\textbf{0.0855}$ \\
        & ICT  & $22.24$ & $\underline{0.802}$ & $9.23$ &  & $\underline{0.1970}$ & $\textbf{24.9}$ & $\underline{0.0771}$ \\
        \midrule
        \multirow{3}{\linewidth}{Irregular $40\%-60\%$} 
        & PIC & $18.33$ & $0.623$ & $\underline{16.34}$ & & $0.0792$ & $183.9$ & $0.1269$ \\
        & DSI-VQVAE & $\textbf{18.92}$ & $\textbf{0.651}$ & $\textbf{14.82}$ &  & $\underline{0.1192}$ & $\underline{126.3}$ & $\textbf{0.1907}$ \\
        & ICT & $\underline{18.52}$ & $\underline{0.646}$ & $16.41$ &  & $\textbf{0.1329}$ & $\textbf{101.7}$ & $\underline{0.1700}$ \\
        \midrule
          \multirow{3}{\linewidth}{Central $128 \times 128$} & PIC & $19.87$ & $0.794$ & $\textbf{9.77}$ &  & $0.1591$ & $95.8$ & $0.1067$ \\
        & DSI-VQVAE & $\underline{20.06}$ & $\textbf{0.795}$ & $\underline{9.99}$ &  & $\textbf{0.1754}$ & $\underline{85.6}$ & $\textbf{0.1291}$ \\
        & ICT & $\textbf{20.34}$ & $\underline{0.795}$ & $10.76$ &  & $\underline{0.1753}$ & $\textbf{73.8}$ & $\underline{0.1162}$ \\
        \midrule
        \multirow{3}{\linewidth}{Random regular} 
        & PIC & $19.81$ & $0.737$ & $13.24$ & & $0.0934$ & $129.2$ & $0.1027$ \\
        & DSI-VQVAE & $\textbf{20.54}$ & $\textbf{0.756}$ & $\textbf{11.52}$ &  & $\underline{0.1305}$ & $\underline{89.3}$ & $\textbf{0.1540}$ \\
        & ICT & $\underline{20.32}$ & $\underline{0.752}$ & $\underline{12.76}$ &  & $\textbf{0.1420}$ & $\textbf{77.6}$ & $\underline{0.1360}$ \\
        \midrule
        \multirow{3}{\linewidth}{Random irregular} 
        & PIC & $\underline{19.51}$ & $0.598$ & $\underline{14.78}$ &  & $0.0645$ & $193.0$ & $0.0982$ \\
        & DSI-VQVAE & $\textbf{19.81}$ & $\textbf{0.636}$ & $\textbf{14.04}$ &  & $\underline{0.1147}$ & $\underline{136.8}$ & $\textbf{0.1757}$ \\
        & ICT & $19.02$ & $\underline{0.628}$ & $16.08$ &  & $\textbf{0.1363}$ & $\textbf{108.5}$ & $\underline{0.1574}$ \\
       \specialrule{\heavyrulewidth}{1pt}{1pt}
        \multirow{3}{\linewidth}{Average} 
        & PIC & $\underline{21.81}$ & $0.748$ & $\underline{10.59}$ & & $0.1347$ & $117.6$  & $0.0735$ \\
        & DSI-VQVAE & $\textbf{22.13}$ & $\textbf{0.765}$ & $\textbf{10.07}$ &  & $\underline{0.1629}$ & $\underline{83.9}$ & $\textbf{0.1258}$ \\
        & ICT & $21.61$ & $\underline{0.761}$ & $11.54$ & & $\textbf{0.1693}$ & $\textbf{66.2}$ & $\underline{0.1125}$ \\
        \bottomrule
        \end{tabular}
        
\end{table}

\begin{table}[ht]
    \caption{Influence of the top-${\mathcal{K}}$ parameter on the ICT results. Results obtained on Places2 dataset, with central mask.}
    \label{tab:topk}
    \centering
               \begin{tabular}{P{2cm}P{0.85cm}P{0.85cm}P{0.7cm}P{0.35cm}P{0.85cm}P{0.85cm} P{1.7cm}}             
        \toprule
         &  \multicolumn{3}{c}{\textbf{Similarity to GT}} & & \multicolumn{2}{c}{\textbf{Realism}}  & \textbf{Diversity}\\
        \textbf{top-${\mathcal{K}}$} & PSNR$\uparrow$ & SSIM$\uparrow$ & L$^1$ $\downarrow$  & & MIS$\uparrow$  & FID$\downarrow$ &  LPIPS$\uparrow$ \\
    \toprule
        5 &  $\textbf{21.76}$ & $\textbf{0.820}$ & $\textbf{6.52}$ &  & $\textbf{0.0510}$ & $\textbf{87.6}$ & $0.0854$ \\
        25 &  $\underline{21.16}$ & $\underline{0.813}$ & $\underline{10.03}$ &  & $\underline{0.0495}$ & $\underline{90.2}$ & $\underline{0.1146}$ \\
        50 & $20.93$ & $0.812$ & $10.22$ &  & $0.0476$ & $92.2$ & $\textbf{0.1204}$ \\
    \bottomrule
    \end{tabular}
\end{table}


\subsubsection{Proximity to Ground Truth}
First, to measure the similarity between the inpainting results and the ground truth (GT), we use the following metrics : peak signal-to-noise ratio (PSNR), L$^1$ loss, and structural similarity (SSIM).  For each input image to be inpainted, those metrics are averaged on the set of inpainted results. 

Note that all the compared methods enforce somehow, in their training loss, similarity between the reconstructed image and the ground truth, either at pixel or at feature level. From the results in Tables \ref{tab:comparisonCeleb}, \ref{tab:comparisonPlaces}, \ref{tab:comparisonImageN}, \ref{tab:comparisonPlacesRe} and  \ref{tab:comparisonImageNRe} we observe that, on all datasets, ICT and PIC obtained slightly lower scores than BAT and DSI-VQVAE in terms of GT similarity. A possible explanation for this performance gap is that, these two methods, contrary to the two others, consider a reconstruction loss only at image level and not at feature level. Being similar at feature levels encourages to generate images having similar low-level (pixels, contours, etc.) and higher-level semantics to the ground-truth.




\subsubsection{Perceptual Quality}
Second, to measure realism in the outputs, we measure perceptual quality by using 
Modified Inception Score (MIS) and Fréchet Inception Distance (FID) metrics (defined by \eqref{eq:MIS} and \eqref{eq:FID}, respectively). These two metrics are computed directly on the whole sets of generated or ground truth images. 

BAT, ICT, and DSI-VQVAE are the methods that provide the best scores on average on all datasets. On the opposite, PIC gives the worst results quantitatively and, as we will see later, also qualitatively.
We argue that a possible reason for the superior performance of BAT, ICT, and DSI-VQVAE is that, with different strategies, they separate the tasks of texture and structure recovery. Each task is handled with a specific subnetwork, first reconstructing structures that then guide the texture recovery. From a more practical point of view, BAT and ICT use transformers for global structure understanding and high-level semantics at a coarse resolution, and CNNs for generating textures at the original resolution. DSI-VQVAE incorporates the multi-scale hierarchical organization of VQ-VAE where the information corresponding to the texture is disentangled from the one about structure and geometry. Accordingly, DSI-VQVAE incorporates two different generators respectively devoted to both levels (cf. Section \ref{sec:achieveDiversity}). Although DSI-VQVAE and PIC are VAE-based methods, DSI-VQVAE has the advantage that first, at low resolution, it proposes diverse completions of structure inside the hole. These different structures then guide the completion of texture at high resolution. PIC does not have this global structure completion (at least, not explicitly). All in all, splitting the estimation of coarse and fine details in two distinct steps seems like a successful approach for high-quality image inpainting.

Note also that BAT is the method that achieves the best scores in terms of realism. Indeed, as explained before, autoregressive transformers have the ability to model longer dependencies across the image than CNN-based methods, which can be crucial for image inpainting.
Note that BAT outperforms the other transformer-based method ICT, especially on irregular masks and large holes. As explained in Section~\ref{ssec:autoregressive}, one can explain this difference by the fact that BAT was trained, not only with bidirectional attention but also with autoregressive sampling. Therefore, it creates better consistency of the reconstructed structures, especially for large missing regions. The very good results of the DSI-VQVAE method also proves that autoregressive modeling is a performant strategy for realistic image inpainting.



Finally, one can observe the influence of the complexity of the training dataset on the performance. 
Notice that the underlying probability distribution of CelebA-HQ dataset is semantically less complex and diverse than the one of Places2 and ImageNet and, thus, training is more difficult in the latter cases. 
We hypothesize that this affects both inpainting quality and inpainting diversity. 
Regarding quality, for example, the average FID  score on all the studied methods trained on CelebA is equal to 33.55, while in the case of Places2 and Imagenet is equal to 86.92 and 128.43, respectively. This gives us an idea of the difference in complexity for each particular dataset.

\subsubsection{Inpainting diversity} 
To measure diversity, we rely on the LPIPS metric. The higher the LPIPS is, the more diverse are the outputs.  For each generated sample, we compute the LPIPS distance with another sample randomly selected from the other 24 results from the same corrupted image.  The reported LPIPS score corresponds to this distance averaged over the 2500 selected pairs.

First and foremost, from the range of LPIPS values on the different datasets, one can again observe the influence of the complexity of the training dataset. CelebA-HQ dataset is semantically more constraint and less complex than the one of Places2 and ImageNet, leading to lower diversity in the outputs. Indeed, the LPIPS is, in average, $\sim 2$ times smaller on CelebA-HQ than on Places2 or ImageNet. Similarly, as expected, all the methods create more diverse samples on larger holes than on smaller holes. 

These observations argue for the existence of a trade-off between inpainting quality and inpainting diversity. The harder the inpainting problem gets (on a more complex dataset or for a larger hole), the more diverse outputs will be created. This trade-off, already highlighted in \cite{yu2021diverse}, also arises when parametrizing a method itself. We study in Table \ref{tab:topk} the influence of the top-${\mathcal{K}}$ parameter on the performance of the ICT algorithm. One can observe that using a smaller ${\mathcal{K}}$ creates outputs that are, on the one hand, closer to GT and more realistic but, on the other hand, less diverse.

PIC is the method giving the less diverse results on all datasets. One reason could be the aforementioned disentanglement of structure and texture of BAT, DSI-VQVAE and ICT. In practice, these three methods first attempt to produce a multiplicity of coherent structures and then fill each of the sampled structure with a deterministic texture generator. This divide-and-conquer approach makes easier the creation of diversity as it is only performed on low-resolution structures and not on the whole reconstructed output. 

ICT slightly outperforms DSI-VQVAE and BAT in terms of LPIPS on the Celeba-HQ testing images. Remind that for this experiment ICT was trained on the more diverse FFHQ dataset. This observation highlights again the influence of the training dataset on the capacity of the model to create diverse outputs.




\subsection{Qualitative Performance} \label{subsec:QualitativeAnalysis}
\begin{figure}[h!]
\centering
\scriptsize Output on $256\times256$ images masked with $128\times128$ center hole\\
\includegraphics[width=0.105\textwidth]{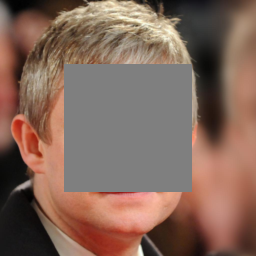}
\includegraphics[width=0.105\textwidth]{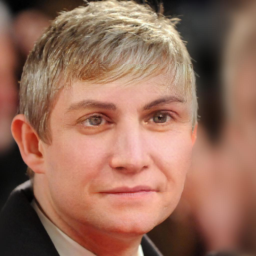}
\includegraphics[width=0.105\textwidth]{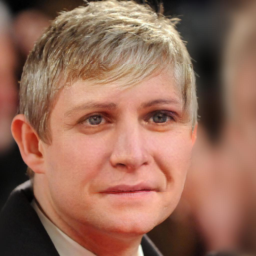}
\includegraphics[width=0.105\textwidth]{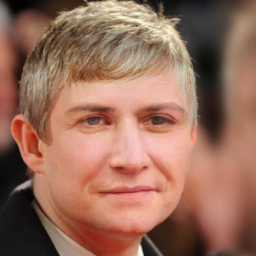}
\includegraphics[width=0.105\textwidth]{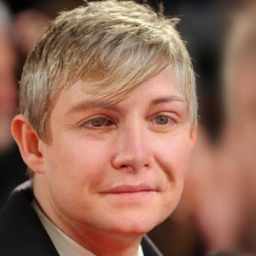}
\includegraphics[width=0.105\textwidth]{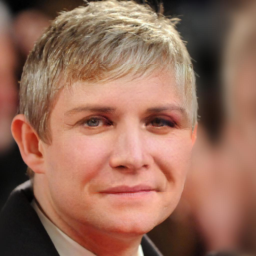}
\raisebox{0.5cm}{\rotatebox[origin=t]{90}{\scriptsize PIC}}
\\
\hspace{0.105\textwidth}
\includegraphics[width=0.105\textwidth]{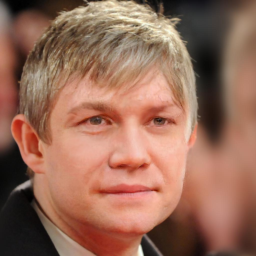}
\includegraphics[width=0.105\textwidth]{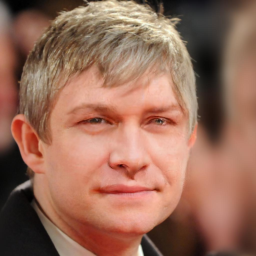}
\includegraphics[width=0.105\textwidth]{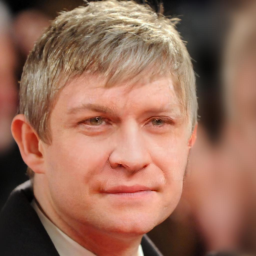}
\includegraphics[width=0.105\textwidth]{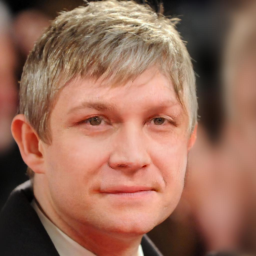}
\includegraphics[width=0.105\textwidth]{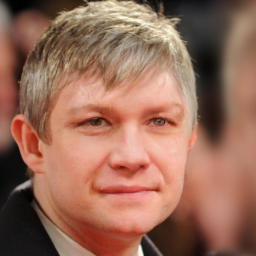}
\raisebox{0.45cm}{\rotatebox[origin=t]{90}{\scriptsize DSI-VQVAE}}\\
\hspace{0.105\textwidth}
\includegraphics[width=0.105\textwidth]{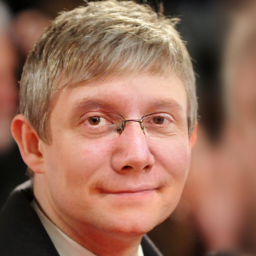}
\includegraphics[width=0.105\textwidth]{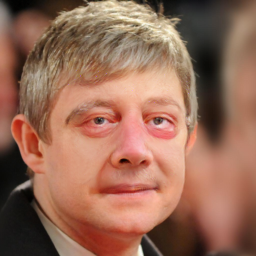}
\includegraphics[width=0.105\textwidth]{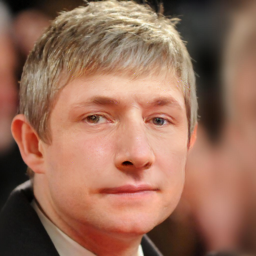}
\includegraphics[width=0.105\textwidth]{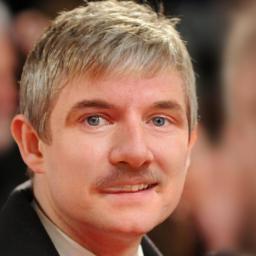}
\includegraphics[width=0.105\textwidth]{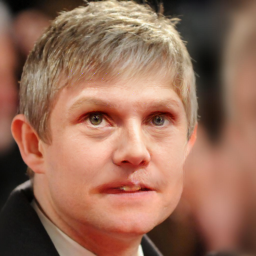}
\raisebox{0.5cm}{\rotatebox[origin=t]{90}{\scriptsize ICT}}
\\
\hspace{0.105\textwidth}
\includegraphics[width=0.105\textwidth]{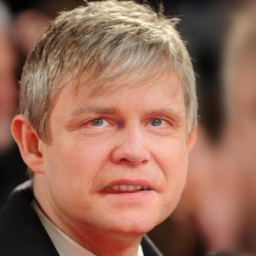}
\includegraphics[width=0.105\textwidth]{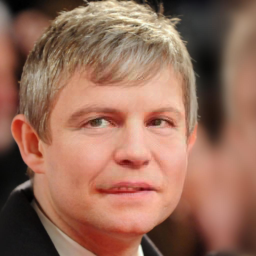}
\includegraphics[width=0.105\textwidth]{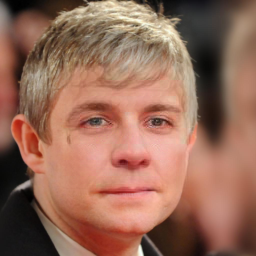}
\includegraphics[width=0.105\textwidth]{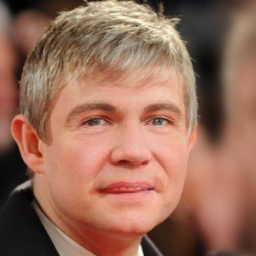}
\includegraphics[width=0.105\textwidth]{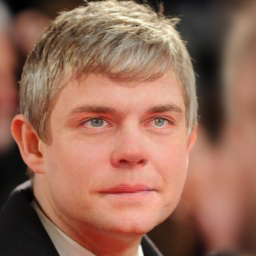}
\raisebox{0.5cm}{\rotatebox[origin=t]{90}{\scriptsize BAT}} 
\\
\scriptsize Output on $256\times256$ images masked with random regular hole
\\
\includegraphics[width=0.105\textwidth]{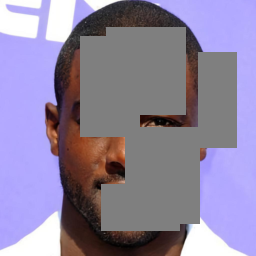}
\includegraphics[width=0.105\textwidth]{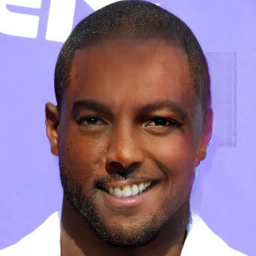}
\includegraphics[width=0.105\textwidth]{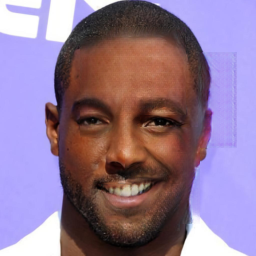}
\includegraphics[width=0.105\textwidth]{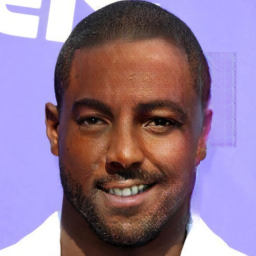}
\includegraphics[width=0.105\textwidth]{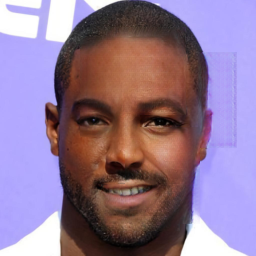}
\includegraphics[width=0.105\textwidth]{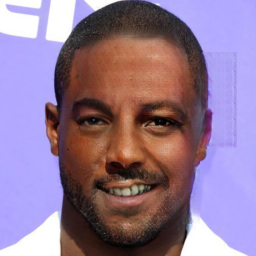}
\raisebox{0.5cm}{\rotatebox[origin=t]{90}{\scriptsize PIC}}
\\
\hspace{0.105\textwidth}
\includegraphics[width=0.105\textwidth]{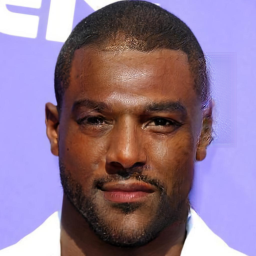}
\includegraphics[width=0.105\textwidth]{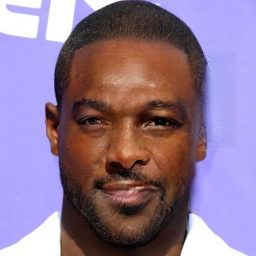}
\includegraphics[width=0.105\textwidth]{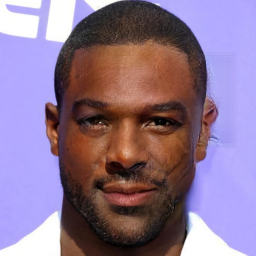}
\includegraphics[width=0.105\textwidth]{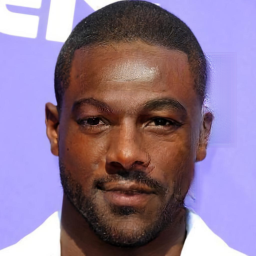}
\includegraphics[width=0.105\textwidth]{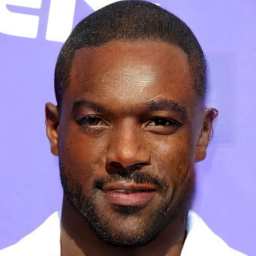}
\raisebox{0.45cm}{\rotatebox[origin=t]{90}{\scriptsize DSI-VQVAE}}
\\
\hspace{0.105\textwidth}
\includegraphics[width=0.105\textwidth]{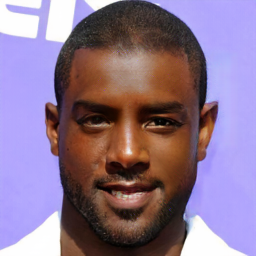}
\includegraphics[width=0.105\textwidth]{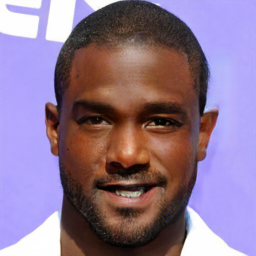}
\includegraphics[width=0.105\textwidth]{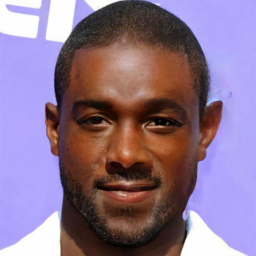}
\includegraphics[width=0.105\textwidth]{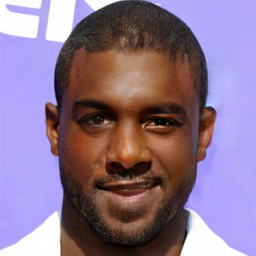}
\includegraphics[width=0.105\textwidth]{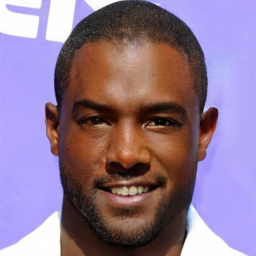}
\raisebox{0.5cm}{\rotatebox[origin=t]{90}{\scriptsize ICT}}
\\
\hspace{0.105\textwidth}
\includegraphics[width=0.105\textwidth]{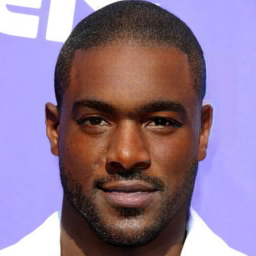}
\includegraphics[width=0.105\textwidth]{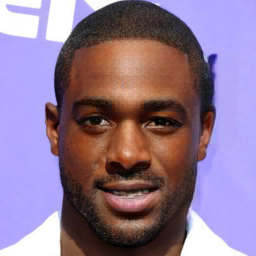}
\includegraphics[width=0.105\textwidth]{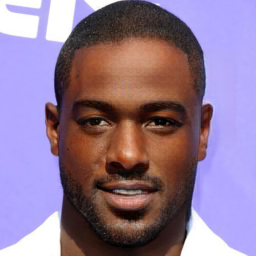}
\includegraphics[width=0.105\textwidth]{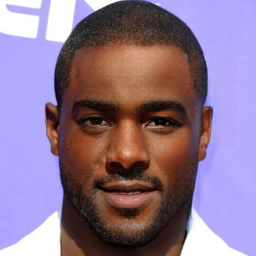}
\includegraphics[width=0.105\textwidth]{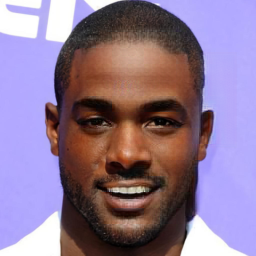}
\raisebox{0.5cm}{\rotatebox[origin=t]{90}{\scriptsize BAT}} 
\\
\scriptsize Output on $256\times256$ images masked with random irregular hole\\
\includegraphics[width=0.105\textwidth]{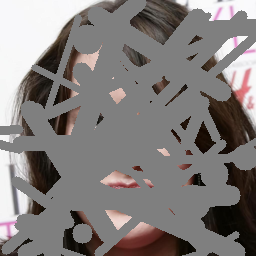}
\includegraphics[width=0.105\textwidth]{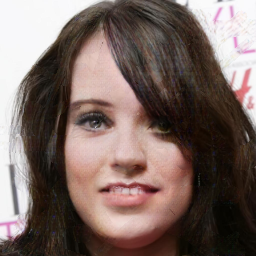}
\includegraphics[width=0.105\textwidth]{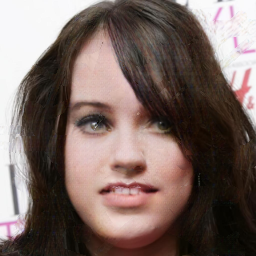}
\includegraphics[width=0.105\textwidth]{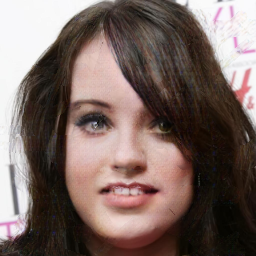}
\includegraphics[width=0.105\textwidth]{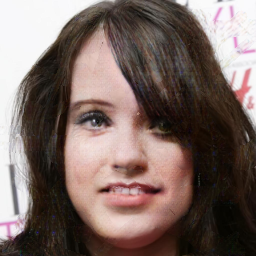}
\includegraphics[width=0.105\textwidth]{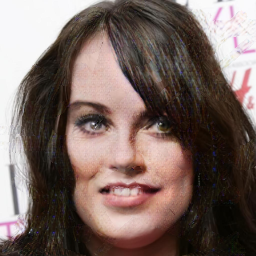}
\raisebox{0.5cm}{\rotatebox[origin=t]{90}{\scriptsize PIC}}
\\
\hspace{0.105\textwidth}
\includegraphics[width=0.105\textwidth]{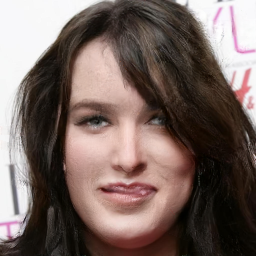}
\includegraphics[width=0.105\textwidth]{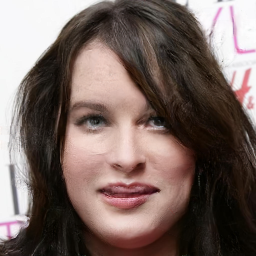}
\includegraphics[width=0.105\textwidth]{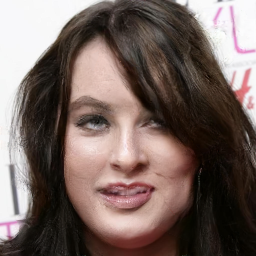}
\includegraphics[width=0.105\textwidth]{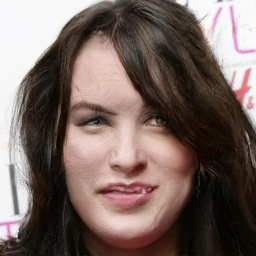}
\includegraphics[width=0.105\textwidth]{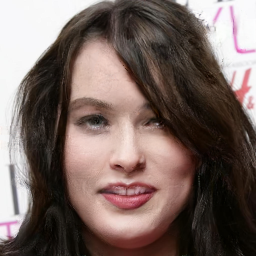}
\raisebox{0.45cm}{\rotatebox[origin=t]{90}{\scriptsize DSI-VQVAE}}
\\
\hspace{0.105\textwidth}
\includegraphics[width=0.105\textwidth]{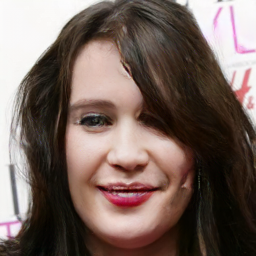}
\includegraphics[width=0.105\textwidth]{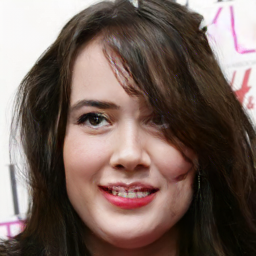}
\includegraphics[width=0.105\textwidth]{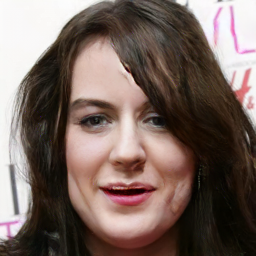}
\includegraphics[width=0.105\textwidth]{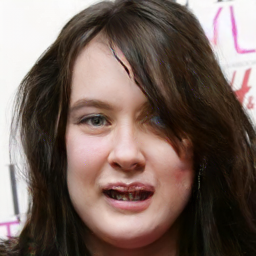}
\includegraphics[width=0.105\textwidth]{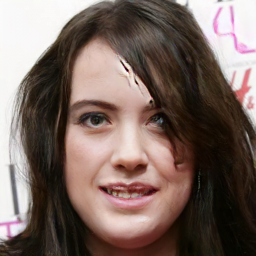}
\raisebox{0.5cm}{\rotatebox[origin=t]{90}{\scriptsize ICT}}
\\
\hspace{0.105\textwidth}
\includegraphics[width=0.105\textwidth]{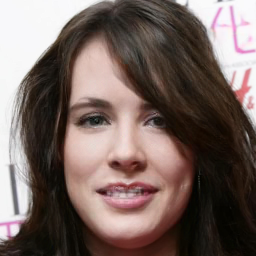}
\includegraphics[width=0.105\textwidth]{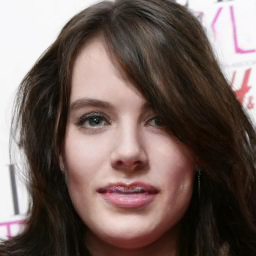}
\includegraphics[width=0.105\textwidth]{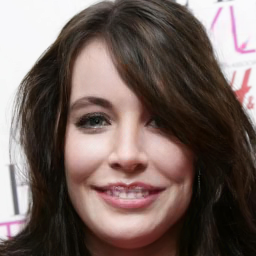}
\includegraphics[width=0.105\textwidth]{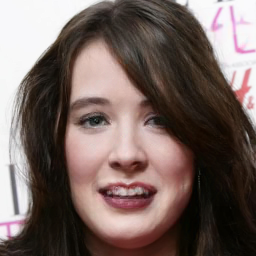}
\includegraphics[width=0.105\textwidth]{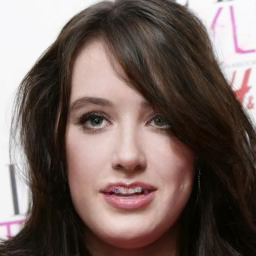}
\raisebox{0.5cm}{\rotatebox[origin=t]{90}{\scriptsize BAT}} 
\\
\caption{Diverse inpainting output on 256$\times$ 256 images from Celeba dataset with center, random regular and random irregular masks. 
For each method, out of 25 generated samples, the 5 samples with highest discriminator score are displayed. \vspace{0.5cm}}
\label{fig:Celeba}
\end{figure}

\begin{figure}[h!]
\centering
Output on $256\times256$ images masked with $128\times128$ center hole\\
\includegraphics[width=0.105\textwidth]{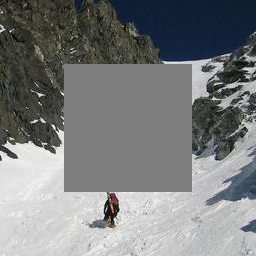}
\includegraphics[width=0.105\textwidth]{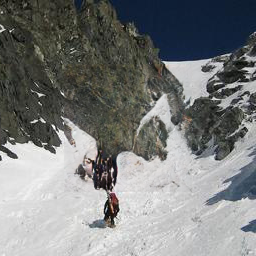}
\includegraphics[width=0.105\textwidth]{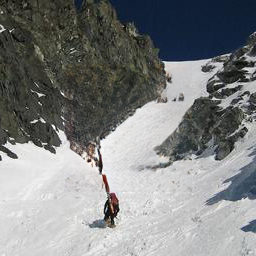}
\includegraphics[width=0.105\textwidth]{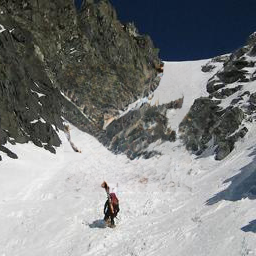}
\includegraphics[width=0.105\textwidth]{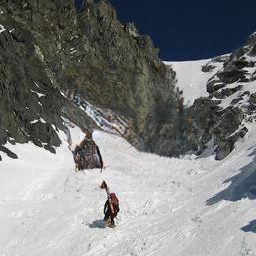}
\includegraphics[width=0.105\textwidth]{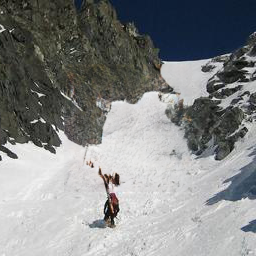}
\raisebox{0.5cm}{\rotatebox[origin=t]{90}{\scriptsize PIC}}
\\
\hspace{0.105\textwidth}
\includegraphics[width=0.105\textwidth]{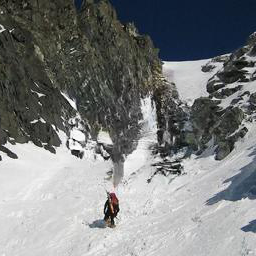}
\includegraphics[width=0.105\textwidth]{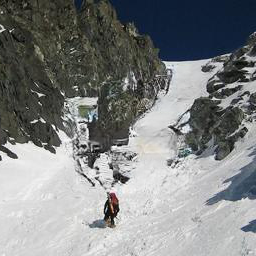}
\includegraphics[width=0.105\textwidth]{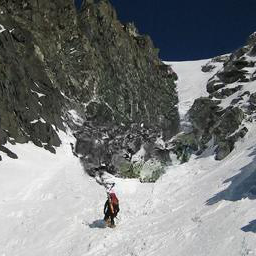}
\includegraphics[width=0.105\textwidth]{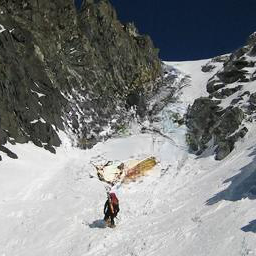}
\includegraphics[width=0.105\textwidth]{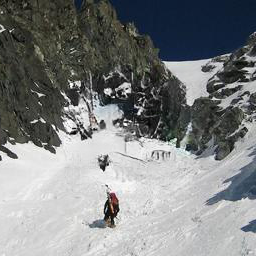}
\raisebox{0.45cm}{\rotatebox[origin=t]{90}{\scriptsize DSI-VQVAE}}\\
\hspace{0.105\textwidth}
\includegraphics[width=0.105\textwidth]{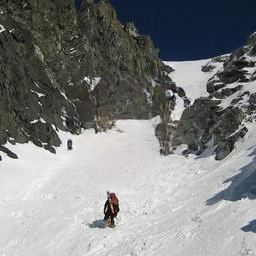}
\includegraphics[width=0.105\textwidth]{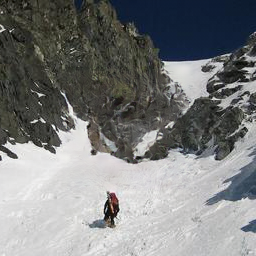}
\includegraphics[width=0.105\textwidth]{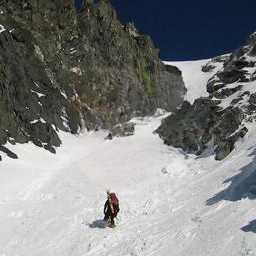}
\includegraphics[width=0.105\textwidth]{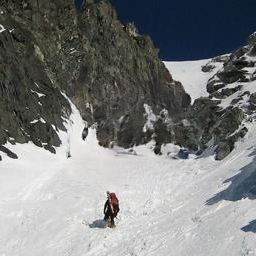}
\includegraphics[width=0.105\textwidth]{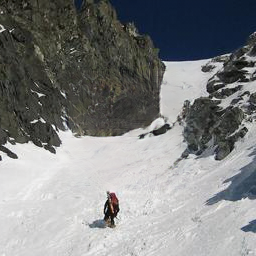}
\raisebox{0.5cm}{\rotatebox[origin=t]{90}{\scriptsize ICT}}
\\
\hspace{0.105\textwidth}
\includegraphics[width=0.105\textwidth]{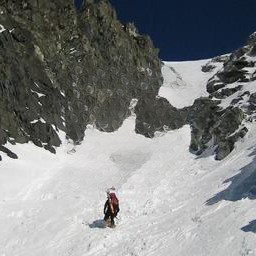}
\includegraphics[width=0.105\textwidth]{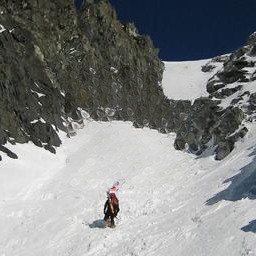}
\includegraphics[width=0.105\textwidth]{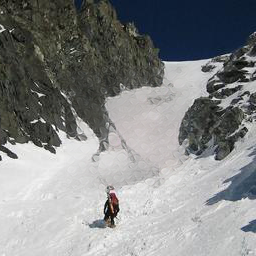}
\includegraphics[width=0.105\textwidth]{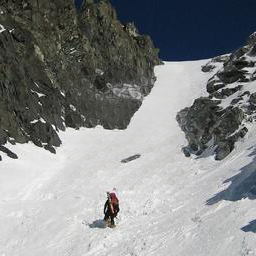}
\includegraphics[width=0.105\textwidth]{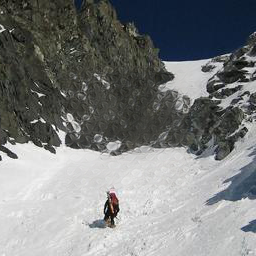}
\raisebox{0.5cm}{\rotatebox[origin=t]{90}{\scriptsize BAT}} 
\\
Output on $256\times256$ images masked with Pconv $40\%-60\%$ hole\\
\includegraphics[width=0.105\textwidth]{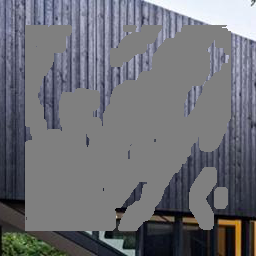}
\includegraphics[width=0.105\textwidth]{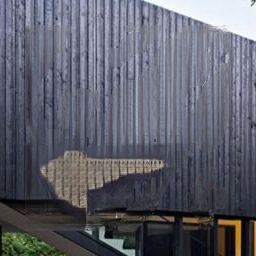}
\includegraphics[width=0.105\textwidth]{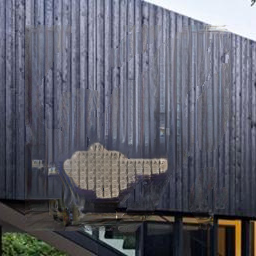}
\includegraphics[width=0.105\textwidth]{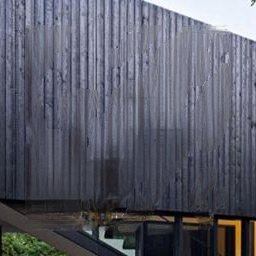}
\includegraphics[width=0.105\textwidth]{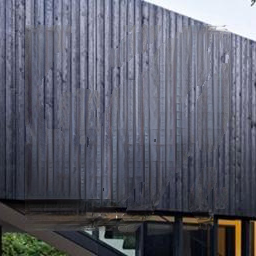}
\includegraphics[width=0.105\textwidth]{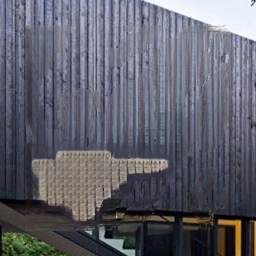}
\raisebox{0.5cm}{\rotatebox[origin=t]{90}{\scriptsize PIC}}
\\
\hspace{0.105\textwidth}
\includegraphics[width=0.105\textwidth]{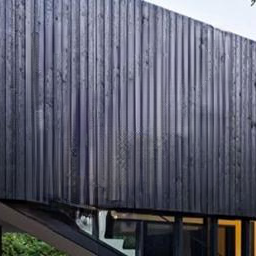}
\includegraphics[width=0.105\textwidth]{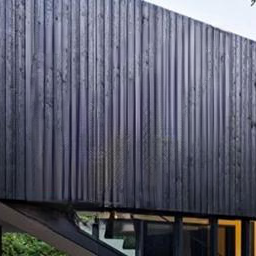}
\includegraphics[width=0.105\textwidth]{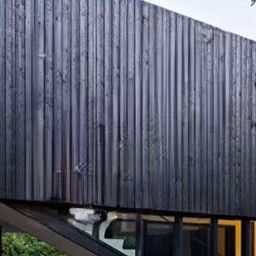}
\includegraphics[width=0.105\textwidth]{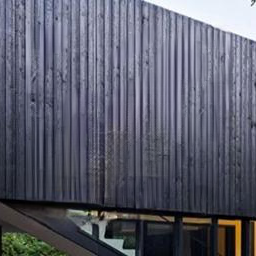}
\includegraphics[width=0.105\textwidth]{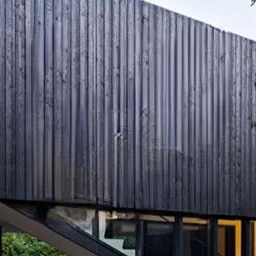}
\raisebox{0.45cm}{\rotatebox[origin=t]{90}{\scriptsize DSI-VQVAE}}\\
\hspace{0.105\textwidth}
\includegraphics[width=0.105\textwidth]{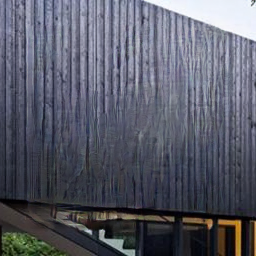}
\includegraphics[width=0.105\textwidth]{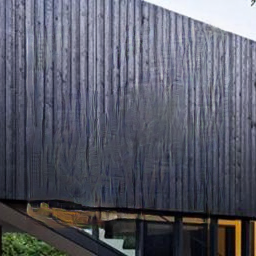}
\includegraphics[width=0.105\textwidth]{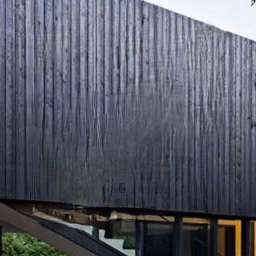}
\includegraphics[width=0.105\textwidth]{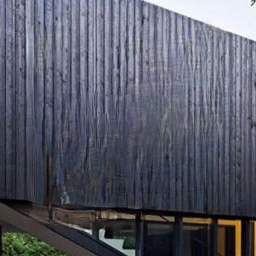}
\includegraphics[width=0.105\textwidth]{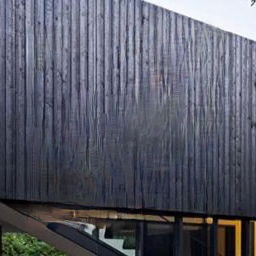}
\raisebox{0.5cm}{\rotatebox[origin=t]{90}{\scriptsize ICT}}
\\
\hspace{0.105\textwidth}
\includegraphics[width=0.105\textwidth]{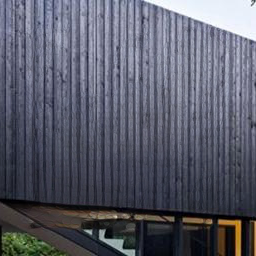}
\includegraphics[width=0.105\textwidth]{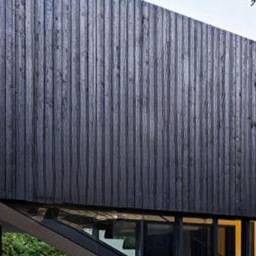}
\includegraphics[width=0.105\textwidth]{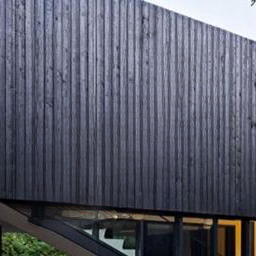}
\includegraphics[width=0.105\textwidth]{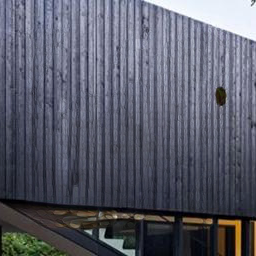}
\includegraphics[width=0.105\textwidth]{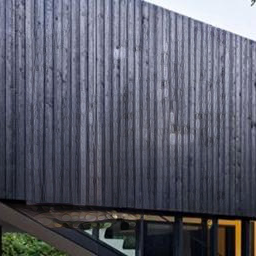}
\raisebox{0.5cm}{\rotatebox[origin=t]{90}{\scriptsize BAT}} 
\\
Output on $256\times256$ images masked with Pconv $20\%-40\%$ hole\\
\includegraphics[width=0.105\textwidth]{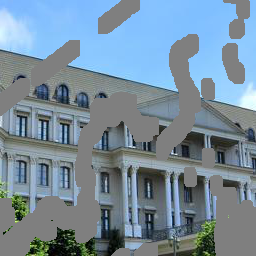}
\includegraphics[width=0.105\textwidth]{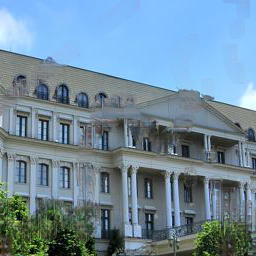}
\includegraphics[width=0.105\textwidth]{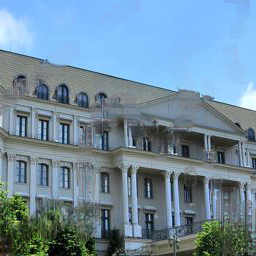}
\includegraphics[width=0.105\textwidth]{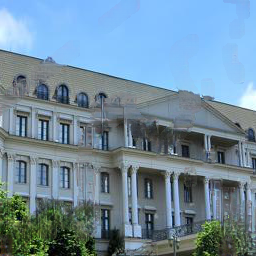}
\includegraphics[width=0.105\textwidth]{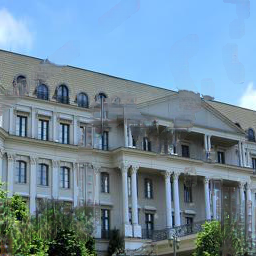}
\includegraphics[width=0.105\textwidth]{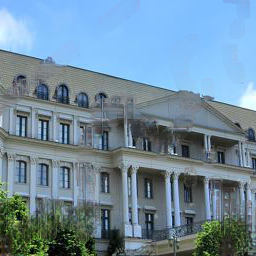}
\raisebox{0.5cm}{\rotatebox[origin=t]{90}{\scriptsize PIC}}
\\
\hspace{0.105\textwidth}
\includegraphics[width=0.105\textwidth]{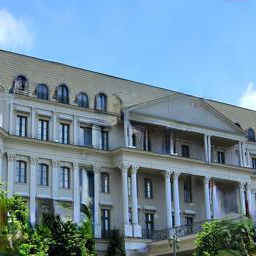}
\includegraphics[width=0.105\textwidth]{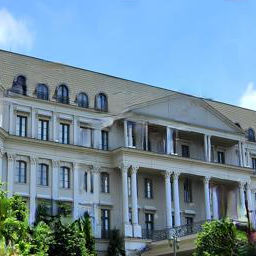}
\includegraphics[width=0.105\textwidth]{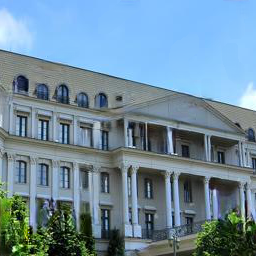}
\includegraphics[width=0.105\textwidth]{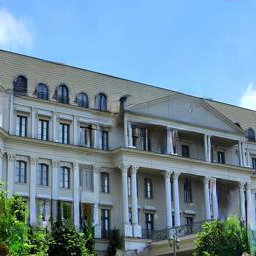}
\includegraphics[width=0.105\textwidth]{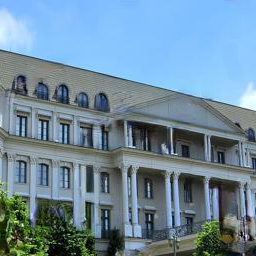}
\raisebox{0.45cm}{\rotatebox[origin=t]{90}{\scriptsize DSI-VQVAE}}
\\
\hspace{0.105\textwidth}
\includegraphics[width=0.105\textwidth]{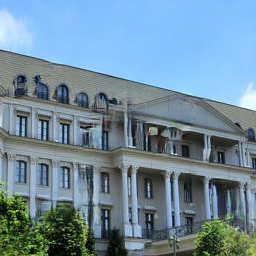}
\includegraphics[width=0.105\textwidth]{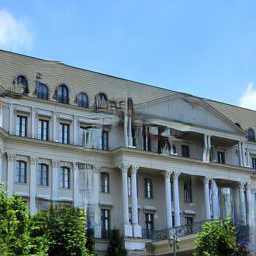}
\includegraphics[width=0.105\textwidth]{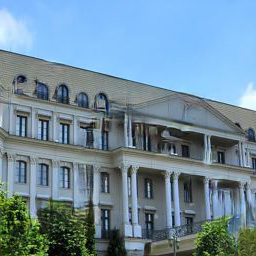}
\includegraphics[width=0.105\textwidth]{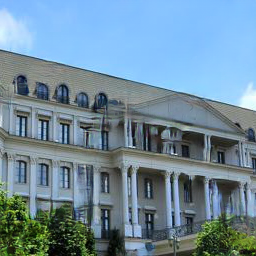}
\includegraphics[width=0.105\textwidth]{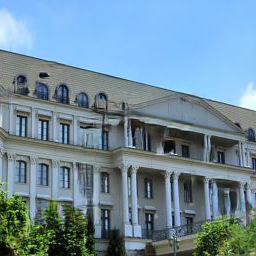}
\raisebox{0.5cm}{\rotatebox[origin=t]{90}{\scriptsize ICT}}
\\
\hspace{0.105\textwidth}
\includegraphics[width=0.105\textwidth]{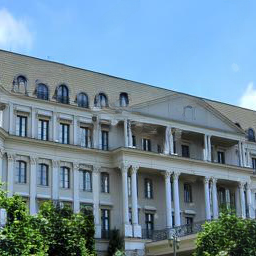}
\includegraphics[width=0.105\textwidth]{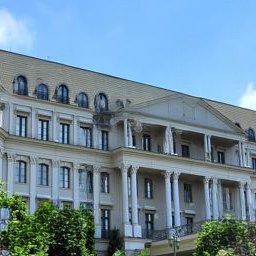}
\includegraphics[width=0.105\textwidth]{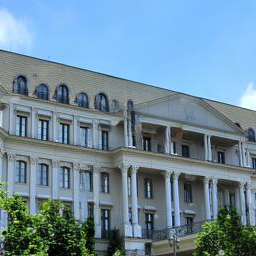}
\includegraphics[width=0.105\textwidth]{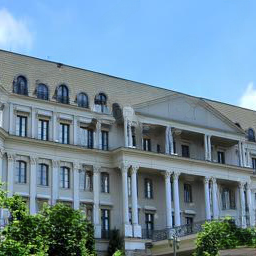}
\includegraphics[width=0.105\textwidth]{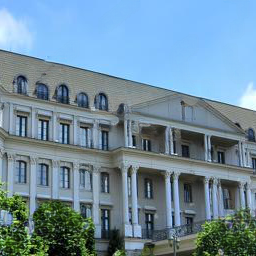}
\raisebox{0.5cm}{\rotatebox[origin=t]{90}{\scriptsize BAT}} 
\\
\caption{Diverse inpainting output on 256$\times$ 256 images from Places2 dataset with center and irregular masks with various proportion of hidden pixels. 
For each method, out of 25 generated samples, the 5 samples with highest discriminator score are displayed.}
\label{fig:Places2}
\end{figure}

\begin{figure}[h!]
\centering
Output on $256\times256$ images masked with $128\times128$ center hole\\
\includegraphics[width=0.105\textwidth]{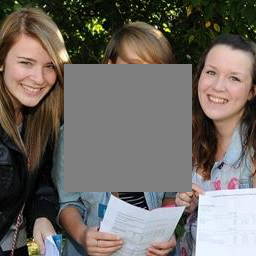}
\includegraphics[width=0.105\textwidth]{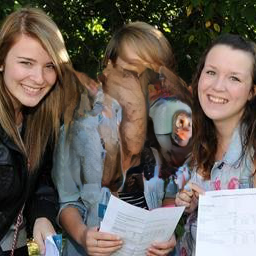}
\includegraphics[width=0.105\textwidth]{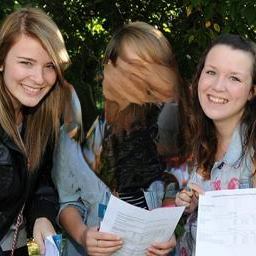}
\includegraphics[width=0.105\textwidth]{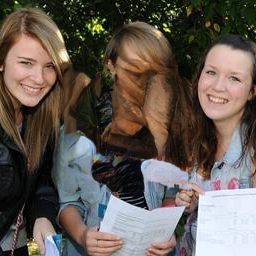}
\includegraphics[width=0.105\textwidth]{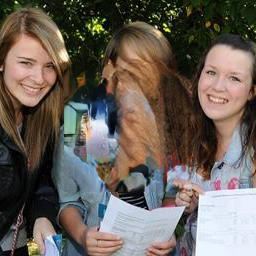}
\includegraphics[width=0.105\textwidth]{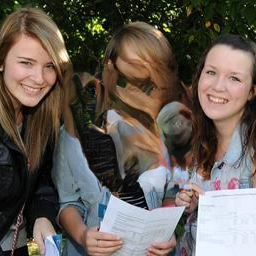}
\raisebox{0.5cm}{\rotatebox[origin=t]{90}{\scriptsize PIC}}
\\
\hspace{0.105\textwidth}
\includegraphics[width=0.105\textwidth]{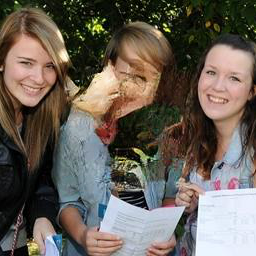}
\includegraphics[width=0.105\textwidth]{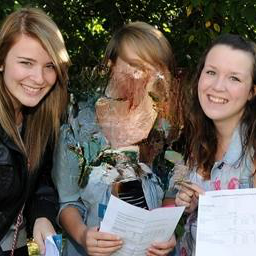}
\includegraphics[width=0.105\textwidth]{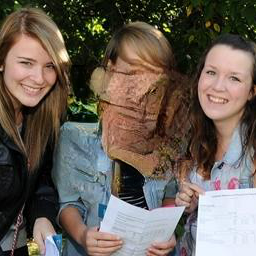}
\includegraphics[width=0.105\textwidth]{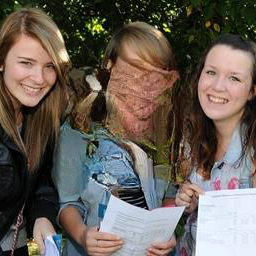}
\includegraphics[width=0.105\textwidth]{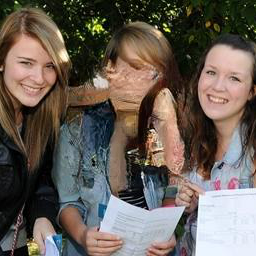}
\raisebox{0.45cm}{\rotatebox[origin=t]{90}{\scriptsize DSI-VQVAE}}\\
\hspace{0.105\textwidth}
\includegraphics[width=0.105\textwidth]{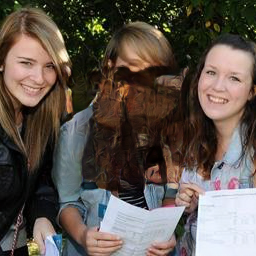}
\includegraphics[width=0.105\textwidth]{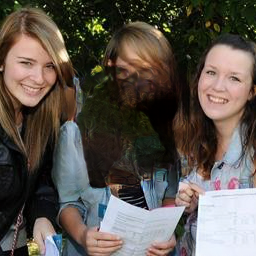}
\includegraphics[width=0.105\textwidth]{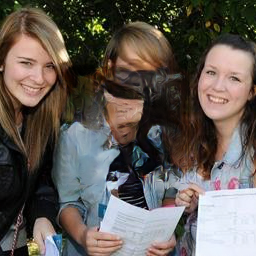}
\includegraphics[width=0.105\textwidth]{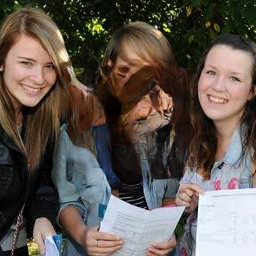}
\includegraphics[width=0.105\textwidth]{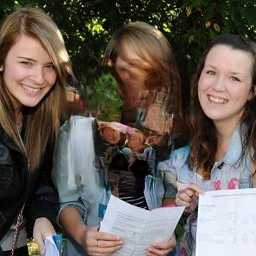}
\raisebox{0.5cm}{\rotatebox[origin=t]{90}{\scriptsize ICT}}
\\
\hspace{0.105\textwidth}
\includegraphics[width=0.105\textwidth]{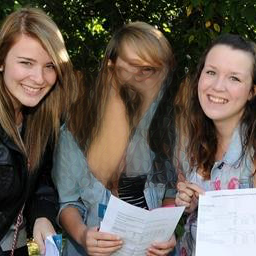}
\includegraphics[width=0.105\textwidth]{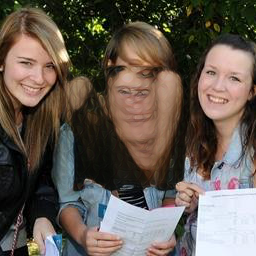}
\includegraphics[width=0.105\textwidth]{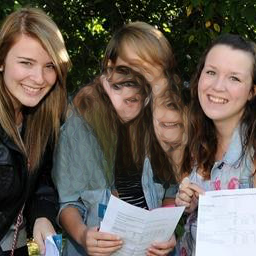}
\includegraphics[width=0.105\textwidth]{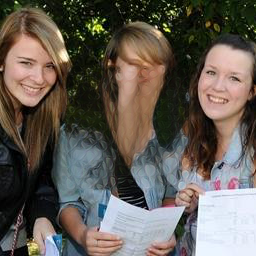}
\includegraphics[width=0.105\textwidth]{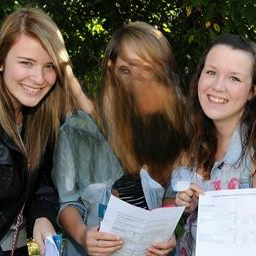}
\raisebox{0.5cm}{\rotatebox[origin=t]{90}{\scriptsize BAT}} 
\\
Output on $256\times256$ images masked with Pconv $40\%-60\%$ hole\\
\includegraphics[width=0.105\textwidth]{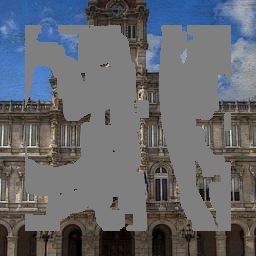}
\includegraphics[width=0.105\textwidth]{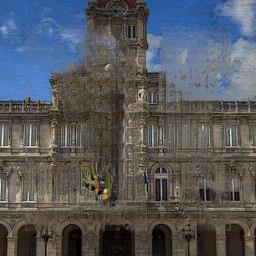}
\includegraphics[width=0.105\textwidth]{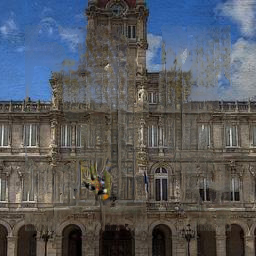}
\includegraphics[width=0.105\textwidth]{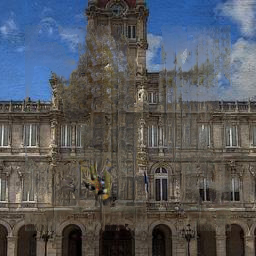}
\includegraphics[width=0.105\textwidth]{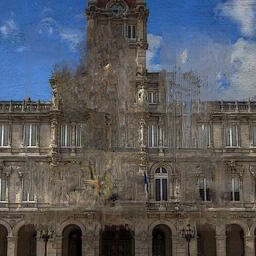}
\includegraphics[width=0.105\textwidth]{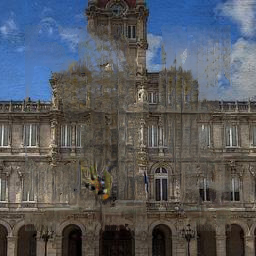}
\raisebox{0.5cm}{\rotatebox[origin=t]{90}{\scriptsize PIC}}
\\
\hspace{0.105\textwidth}
\includegraphics[width=0.105\textwidth]{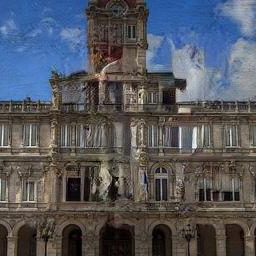}
\includegraphics[width=0.105\textwidth]{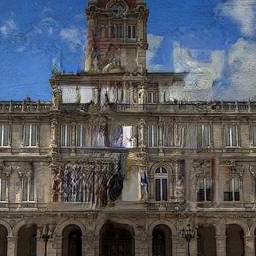}
\includegraphics[width=0.105\textwidth]{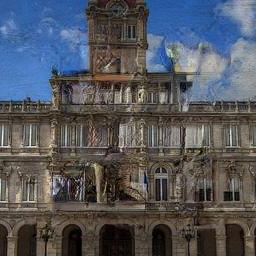}
\includegraphics[width=0.105\textwidth]{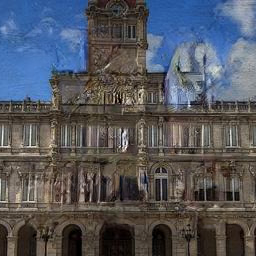}
\includegraphics[width=0.105\textwidth]{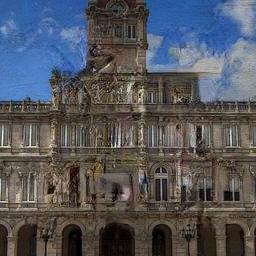}
\raisebox{0.45cm}{\rotatebox[origin=t]{90}{\scriptsize DSI-VQVAE}}\\
\hspace{0.105\textwidth}
\includegraphics[width=0.105\textwidth]{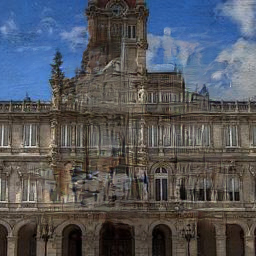}
\includegraphics[width=0.105\textwidth]{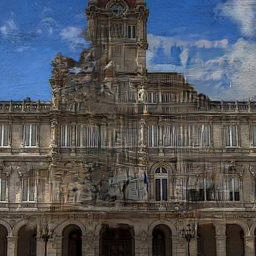}
\includegraphics[width=0.105\textwidth]{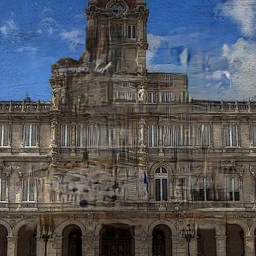}
\includegraphics[width=0.105\textwidth]{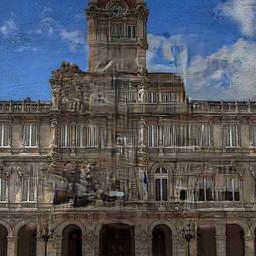}
\includegraphics[width=0.105\textwidth]{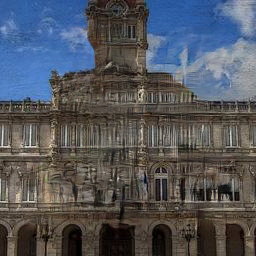}
\raisebox{0.5cm}{\rotatebox[origin=t]{90}{\scriptsize ICT}}
\\
\hspace{0.105\textwidth}
\includegraphics[width=0.105\textwidth]{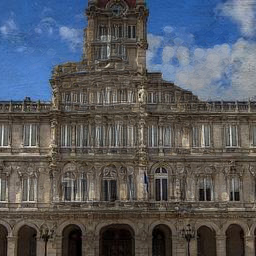}
\includegraphics[width=0.105\textwidth]{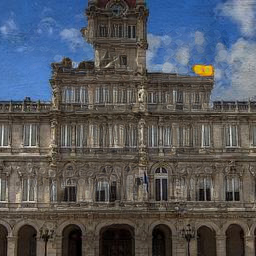}
\includegraphics[width=0.105\textwidth]{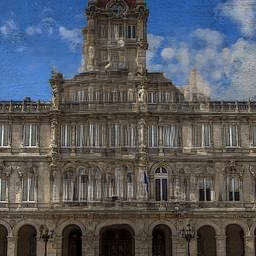}
\includegraphics[width=0.105\textwidth]{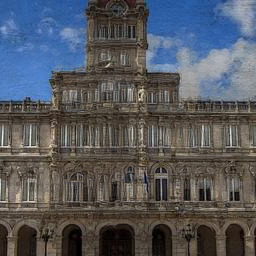}
\includegraphics[width=0.105\textwidth]{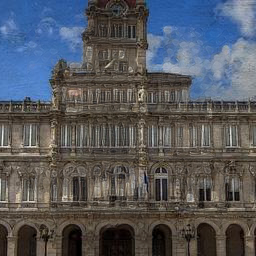}
\raisebox{0.5cm}{\rotatebox[origin=t]{90}{\scriptsize BAT}} 
\\
Output on $256\times256$ images masked with random irregulat hole\\
\includegraphics[width=0.105\textwidth]{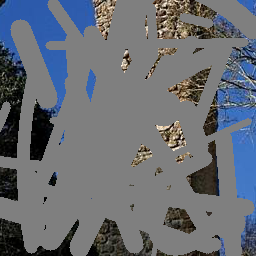}
\includegraphics[width=0.105\textwidth]{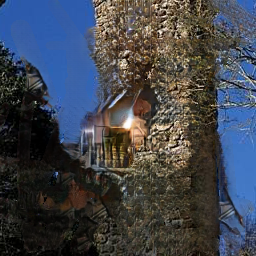}
\includegraphics[width=0.105\textwidth]{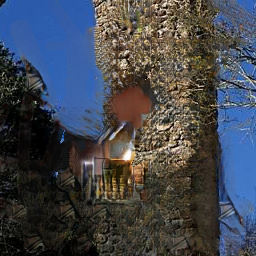}
\includegraphics[width=0.105\textwidth]{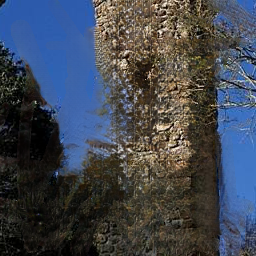}
\includegraphics[width=0.105\textwidth]{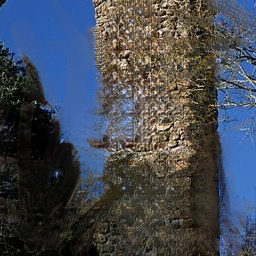}
\includegraphics[width=0.105\textwidth]{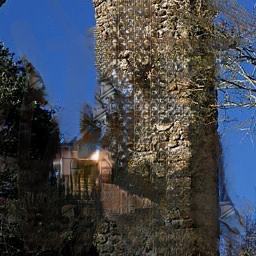}
\raisebox{0.5cm}{\rotatebox[origin=t]{90}{\scriptsize PIC}}
\\
\hspace{0.105\textwidth}
\includegraphics[width=0.105\textwidth]{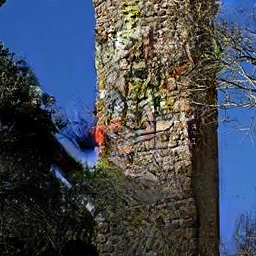}
\includegraphics[width=0.105\textwidth]{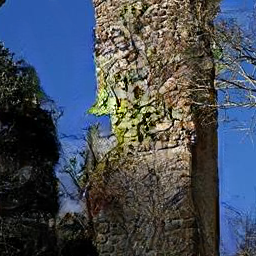}
\includegraphics[width=0.105\textwidth]{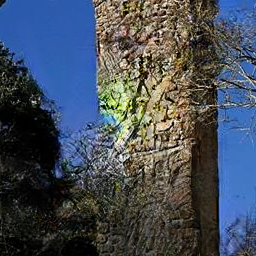}
\includegraphics[width=0.105\textwidth]{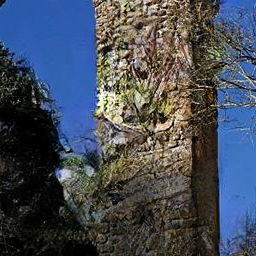}
\includegraphics[width=0.105\textwidth]{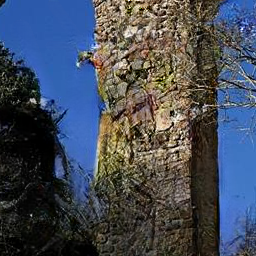}
\raisebox{0.45cm}{\rotatebox[origin=t]{90}{\scriptsize DSI-VQVAE}}
\\
\hspace{0.105\textwidth}
\includegraphics[width=0.105\textwidth]{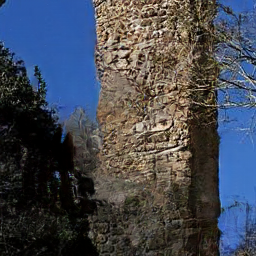}
\includegraphics[width=0.105\textwidth]{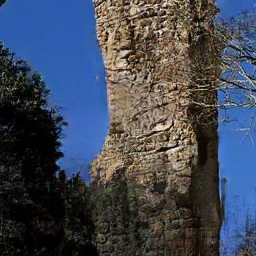}
\includegraphics[width=0.105\textwidth]{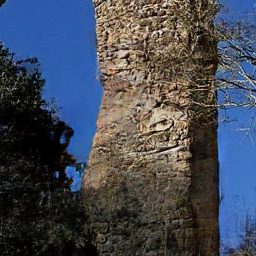}
\includegraphics[width=0.105\textwidth]{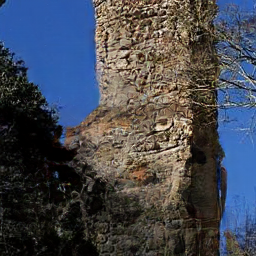}
\includegraphics[width=0.105\textwidth]{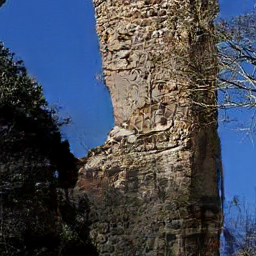}
\raisebox{0.5cm}{\rotatebox[origin=t]{90}{\scriptsize ICT}}
\\
\hspace{0.105\textwidth}
\includegraphics[width=0.105\textwidth]{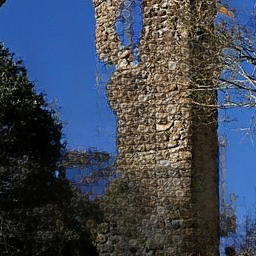}
\includegraphics[width=0.105\textwidth]{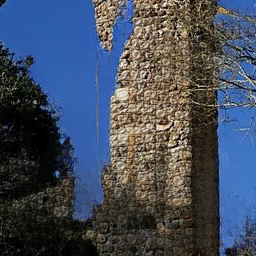}
\includegraphics[width=0.105\textwidth]{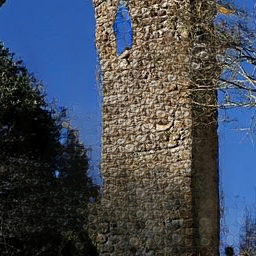}
\includegraphics[width=0.105\textwidth]{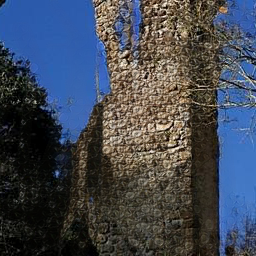}
\includegraphics[width=0.105\textwidth]{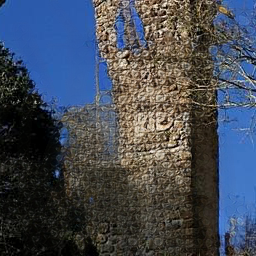}
\raisebox{0.5cm}{\rotatebox[origin=t]{90}{\scriptsize BAT}} 
\\
\caption{Diverse inpainting output on 256$\times$ 256 images from Places2 dataset with center and irregular masks with various proportion of hidden pixels. 
For each method, out of 25 generated samples, the 5 samples with highest discriminator score are displayed.}
\label{fig:Places2_2}
\end{figure}

\begin{figure}[h!]
\centering
Output on $256\times256$ images masked with random irregular hole\\
\includegraphics[width=0.105\textwidth]{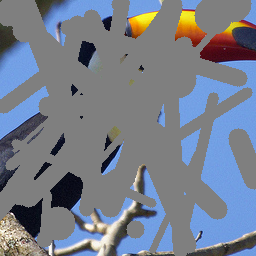}
\includegraphics[width=0.105\textwidth]{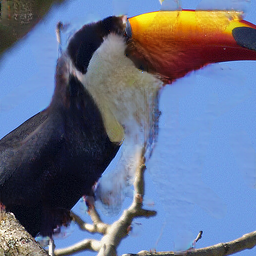}
\includegraphics[width=0.105\textwidth]{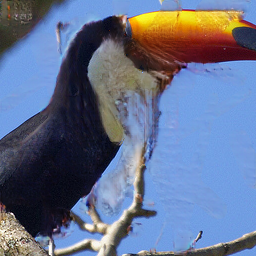}
\includegraphics[width=0.105\textwidth]{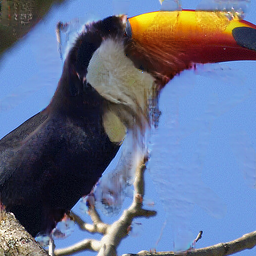}
\includegraphics[width=0.105\textwidth]{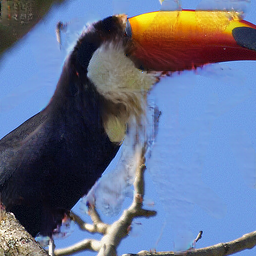}
\includegraphics[width=0.105\textwidth]{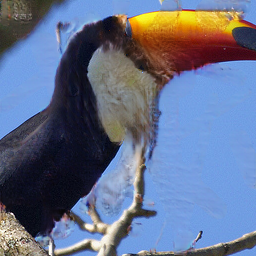}
\raisebox{0.5cm}{\rotatebox[origin=t]{90}{\scriptsize PIC}}
\\
\hspace{0.105\textwidth}
\includegraphics[width=0.105\textwidth]{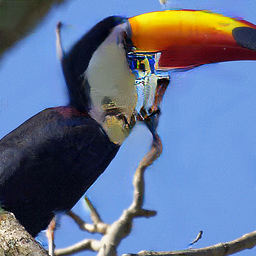}
\includegraphics[width=0.105\textwidth]{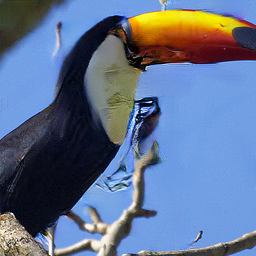}
\includegraphics[width=0.105\textwidth]{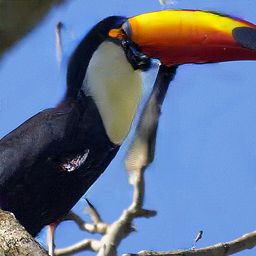}
\includegraphics[width=0.105\textwidth]{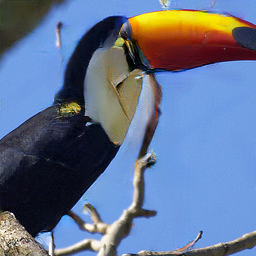}
\includegraphics[width=0.105\textwidth]{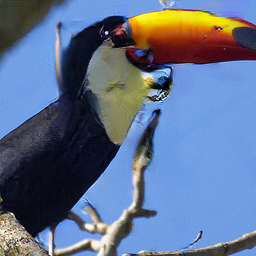}
\raisebox{0.45cm}{\rotatebox[origin=t]{90}{\scriptsize DSI-VQVAE}}\\
\hspace{0.105\textwidth}
\includegraphics[width=0.105\textwidth]{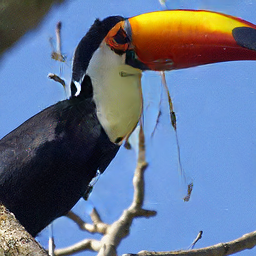}
\includegraphics[width=0.105\textwidth]{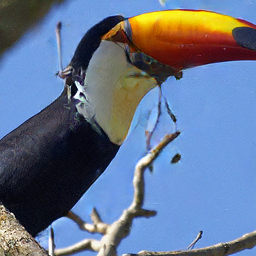}
\includegraphics[width=0.105\textwidth]{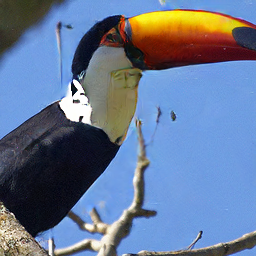}
\includegraphics[width=0.105\textwidth]{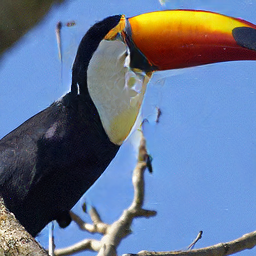}
\includegraphics[width=0.105\textwidth]{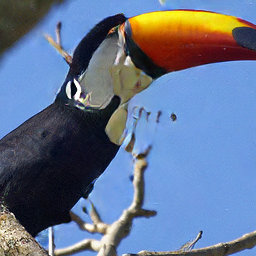}
\raisebox{0.5cm}{\rotatebox[origin=t]{90}{\scriptsize ICT}}
\\
Output on $256\times256$ images masked with random regular hole\\
\includegraphics[width=0.105\textwidth]{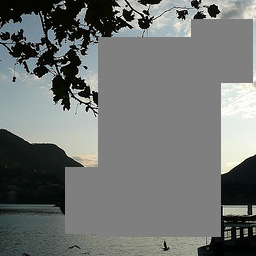}
\includegraphics[width=0.105\textwidth]{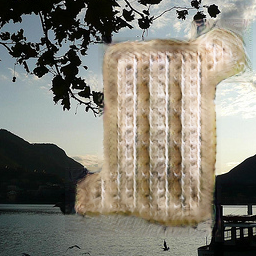}
\includegraphics[width=0.105\textwidth]{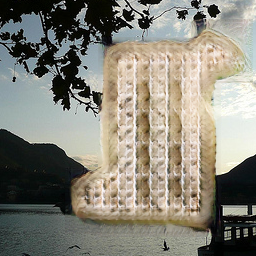}
\includegraphics[width=0.105\textwidth]{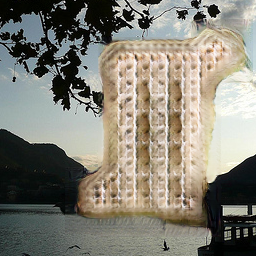}
\includegraphics[width=0.105\textwidth]{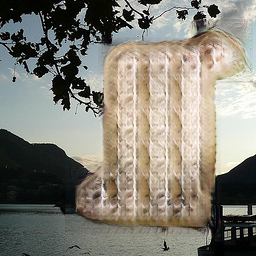}
\includegraphics[width=0.105\textwidth]{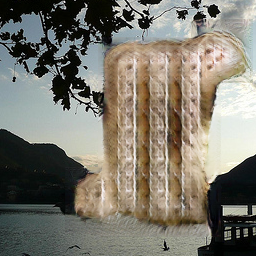}
\raisebox{0.5cm}{\rotatebox[origin=t]{90}{\scriptsize PIC}}
\\
\hspace{0.105\textwidth}
\includegraphics[width=0.105\textwidth]{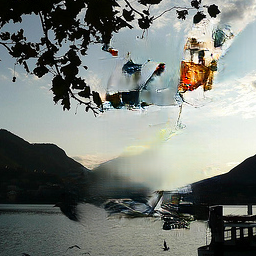}
\includegraphics[width=0.105\textwidth]{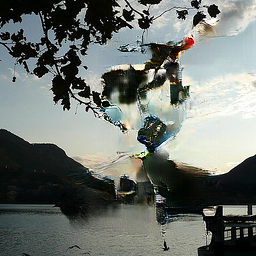}
\includegraphics[width=0.105\textwidth]{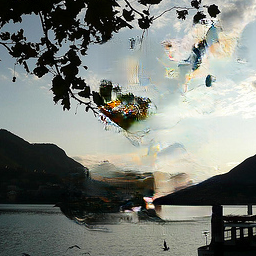}
\includegraphics[width=0.105\textwidth]{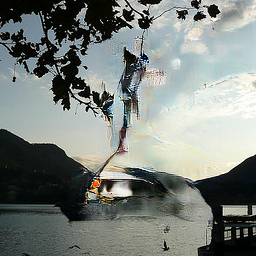}
\includegraphics[width=0.105\textwidth]{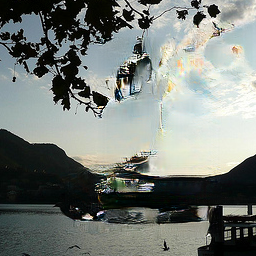}
\raisebox{0.45cm}{\rotatebox[origin=t]{90}{\scriptsize DSI-VQVAE}}\\
\hspace{0.105\textwidth}
\includegraphics[width=0.105\textwidth]{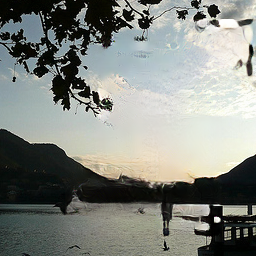}
\includegraphics[width=0.105\textwidth]{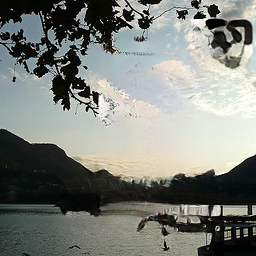}
\includegraphics[width=0.105\textwidth]{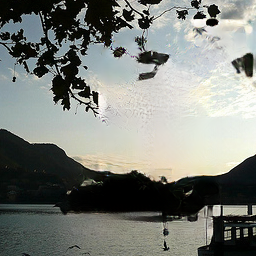}
\includegraphics[width=0.105\textwidth]{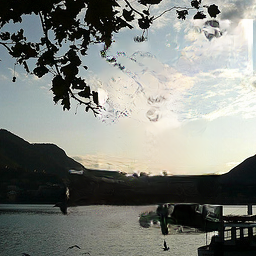}
\includegraphics[width=0.105\textwidth]{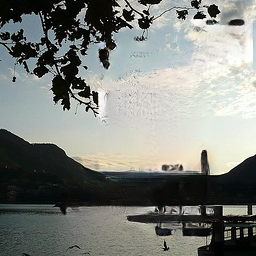}
\raisebox{0.5cm}{\rotatebox[origin=t]{90}{\scriptsize ICT}}
\\
Output on $256\times256$ images masked with Pconv $20\%-40\%$ hole\\
\includegraphics[width=0.105\textwidth]{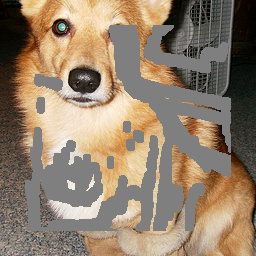}
\includegraphics[width=0.105\textwidth]{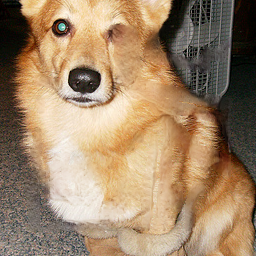}
\includegraphics[width=0.105\textwidth]{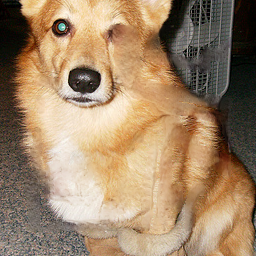}
\includegraphics[width=0.105\textwidth]{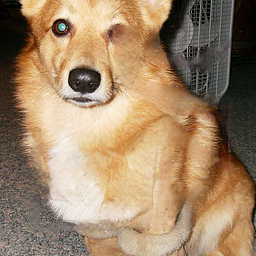}
\includegraphics[width=0.105\textwidth]{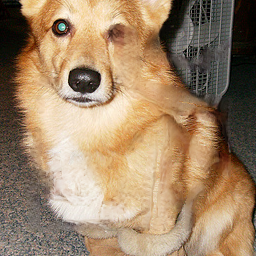}
\includegraphics[width=0.105\textwidth]{images/imagenet/crop/pconvb/PIC/18_out_23_0.54917.png}
\raisebox{0.5cm}{\rotatebox[origin=t]{90}{\scriptsize PIC}}
\\
\hspace{0.105\textwidth}
\includegraphics[width=0.105\textwidth]{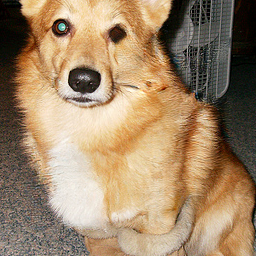}
\includegraphics[width=0.105\textwidth]{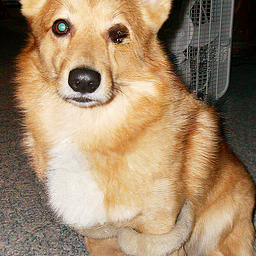}
\includegraphics[width=0.105\textwidth]{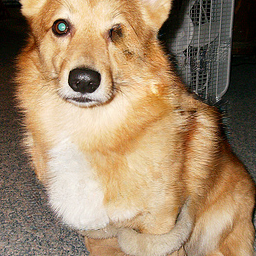}
\includegraphics[width=0.105\textwidth]{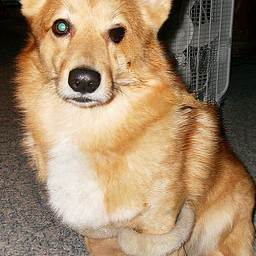}
\includegraphics[width=0.105\textwidth]{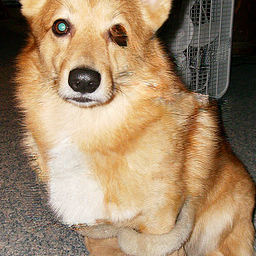}
\raisebox{0.45cm}{\rotatebox[origin=t]{90}{\scriptsize DSI-VQVAE}}
\\
\hspace{0.105\textwidth}
\includegraphics[width=0.105\textwidth]{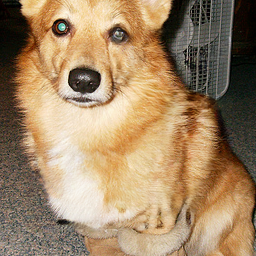}
\includegraphics[width=0.105\textwidth]{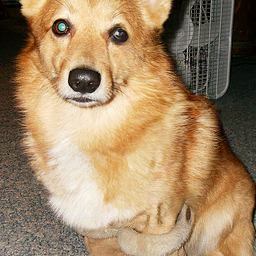}
\includegraphics[width=0.105\textwidth]{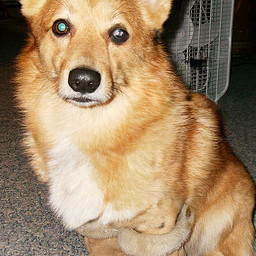}
\includegraphics[width=0.105\textwidth]{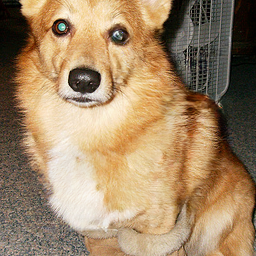}
\includegraphics[width=0.105\textwidth]{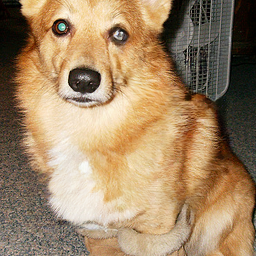}
\raisebox{0.5cm}{\rotatebox[origin=t]{90}{\scriptsize ICT}}
\\
\caption{Diverse inpainting output on 256$\times$ 256 images from ImageNet dataset with center and irregular masks with various proportion of hidden pixels. 
For each method, out of 25 generated samples, the 5 samples with highest discriminator score are displayed.}
\label{fig:ImageNetbis}
\end{figure}

For qualitative comparison, we select for each method the 5 samples with the highest discriminator score out of the 25 generated samples. We use pretrained discriminators given by each of the models, \emph{i.e.}, for PIC, the discriminator of the generative pipeline, for DSI-VQVAE, the discriminator of the texture generation module, and for ICT and BAT, the discriminator of the upsampling module. We perform this comparison on a representative selection of testing images and masks. Figure~\ref{fig:Celeba}, \ref{fig:Places2}, \ref{fig:Places2_2} and \ref{fig:ImageNetbis} show some results on CelebA-HQ, Places2 and ImageNet datasets for the methods PIC, DSI-VQVAE, ICT and BAT. BAT does not provide weights for ImageNet. Remind that ICT was not trained on Celeba-HQ but on FFHQ. Additional visual results are also given in the Appendix. 


\smallbreak

At first glance, we observe that DSI-VQVAE, ICT and BAT provide more plausibly visual results than PIC. PIC struggles to recover information on less constrained datasets like Places2 and Imagenet, and creates strong artifacts when applied to large missing regions (see second examples in Figures~\ref{fig:Places2} and~\ref{fig:ImageNetbis}). 
Among these methods, BAT and ICT propose the most realistic outputs. For instance, in Figure~\ref{fig:Celeba}, 
PIC generates results that do not maintain the proportions and harmony of a face (see the second example). DSI-VQVAE neither has a full understanding of the image: for example, in the second example in Figure~\ref{fig:Celeba} and the third example in Figure~\ref{fig:ImageNetbis}, one eye is visible in the input image but the other is not. On the opposite, transformer-based methods are able to reconstruct a left eye similar to the right. This can be explained by the capability of transformers to have a global structure understanding and high-level semantics. More examples strengthening this observation are the first example of Figure~\ref{fig:Places2} where the inpainting of the snow is sometimes no realistic, or all the ImageNet results Figure~\ref{fig:ImageNetbis}.

When images contain strong structures, like Figure~\ref{fig:Places2} and \ref{fig:Places2_2}, transformer-based methods again estimate more realistic reconstructions. This can be explained by the fact that they include previously predicted tokens in the training objective and thus, global consistency is imposed over the results. This consistency shall avoid problems in the center of big holes. In some situations, such as the middle example in Figure~\ref{fig:Places2}, the structure and texture disentanglement of DSI-VQVAE also provides good reconstructions.   

Notice that, qualitatively speaking, although ICT and BAT were only trained on irregular masks, we do not observe a drop in performance while performing inpainting on regular masks. This shows the capacity of those methods to generalize to unseen type of mission regions.

\smallbreak
In terms of diversity, transformer-based methods are visually more diverse. For example, in Figure~\ref{fig:Celeba}, each transformer-based inpainted face corresponds to a different expression or different person, while in the case of DSI-VQVAE all generated faces are very similar. Also, in Figure~\ref{fig:Places2_2}, even if one could imagine the result quite deterministic, ICT and BAT aims to propose multiple possibilities. Note the multiplicity of structures obtained by ICT compared to DSI-VQVAE and PIC in the chest of the dog or the skyline in Figure~\ref{fig:ImageNetbis}.

\smallbreak

Regarding the difference across datasets, while methods trained on CelebA-HQ all obtain satisfactory results (Figure~\ref{fig:Celeba}), results on Places2 (Figure~\ref{fig:Places2_2}) and ImageNet (Figure~\ref{fig:ImageNetbis}) are often not visually  satisfactory. Also, as already noticed numerically, diversity is less visible on these two datasets. 
This is probably because the models have difficulties learning the underlying multimodal distribution of these complex and diverse datasets. 
This demonstrates the need for further research on the topic to be able to deal with real inpainting scenarios.

\bigskip
\noindent\textbf{Computational Time. }
Despite image quality, an important aspect that should be considered when choosing an inpainting method is its inference time. In Table \ref{tab:runningT}, for the four analyzed methods, we give the average runtime to sample one inpainting result from a central hole on a $256\times256$ input image. We run the experiments on a single P100 GPU. Despite showing lower inpainting quality or diversity (see before), PIC is tremendously faster to run than all the other methods ($\sim$100 times faster than DSI-VQVAE and ICT and $\sim$ 50 times faster than BAT). While providing good results, inference time on autoregressive or transformer-based methods can be prohibitive for time-restricted applications. 

\begin{table}[ht]
    \caption{Average runtime to sample one inpainting result on a single P100 GPU for the four compared methods. Experiments conducted for \textbf{central} masks}
    \label{tab:runningT}
    \centering
    \begin{tabular}{cc}
    \toprule
        Method & Time (s) \\
    \toprule
        PIC & 0.4 \\
        DSI-VQVAE & 55 \\
        ICT & 43  \\
        BAT & 21 \\
        \bottomrule
    \end{tabular}
\end{table}

\section{Conclusions}\label{sec:conclusions}
In this chapter, we have tackled the question of whether generative methods are a suitable strategy to obtain multiple solutions to problems that do not have a unique solution. 
By focusing on the inpainting problem, we have reviewed the main generative models and recent learning-based image completion methods for multiple and diverse inpainting.
We have compared the methods with available code and model weights on three public datasets. 
We have shown that the transformer-based method BAT (or BAT-Fill) and the VQ-VAE-based method DSI-VQVAE provide the best results in both inpainting quality and multiple inpainting diversity. 
This is true both quantitatively and qualitatively. Our analysis highlights that their advantageous results are due to their strategy that consists in, first, sampling multiple structures inside the missing regions, and second, generating textures at higher resolution in a deterministic way. The PIC method is however computationally way faster than the concurrence. 
Moreover, our analysis shows that the multiple inpainting problem is not solved yet.
The difficulty of learning the probability distribution depending on the training dataset is also evident from our study. 
Therefore, we argue that most efforts should be made on improving and exploring new generative strategies to enhance both the quality and diversity of the solutions of this ill-posed inverse problem with multiple solutions. Finally, the computational burden of some of the transformers-based or autoregressive methods is prohibitive for sampling a high number of solutions in reasonable time. We think that this limitation has been overviewed for the purpose of image quality but should be now primarily addressed.  

\begin{acknowledgement}
 PV, CB and AB acknowledge  the EU Horizon 2020 research and innovation programme NoMADS (Marie Skłodowska-Curie grant agreement No 777826). SP acknowledges the Leverhulme Trust Research Project Grant “Unveiling the invisible: Mathematics for Conservation in Arts and Humanities”. CB and PV also acknowledge partial support by MICINN/FEDER UE project, ref. PGC2018-098625-B-I00, and  RED2018-102511-T. AB also acknowledges the French Research Agency through the PostProdLEAP project (ANR-19-CE23-0027-01). SH acknowledges the French ministry of research through a CDSN grant of ENS Paris-Saclay.
\end{acknowledgement}

\bibliographystyle{apalike}
\bibliography{references}

\section*{Appendix}

\subsection*{Additional Quantitative Results}
We provide in this section additional quantitative results on Places2 and ImageNet. Results from Tables~\ref{tab:comparisonPlacesRe} and \ref{tab:comparisonImageNRe}
were conducted in the same conditions as Tables~\ref{tab:comparisonPlaces} and \ref{tab:comparisonImageN} but with $256 \times 256$ \textbf{resized} images instead of center-cropped images. Note that, in average, on both Places2 and ImageNet, the difference between methods is very similar when computed on resized or cropped images. The modification in aspect ratio due to the resize operation does not impede the results, even for models that were trained on "real" aspect ratios. The main reason for this is that the aspect ratio is not drastically changed when resizing Places2 and ImageNet images. Another explanation is that the training datasets are large enough and the models have enough capacity for being robust to such a transformation.

\begin{table}[ht]
\caption{Quantitative comparison of three pluralistic image inpainting methods (PIC, DSI-VQVAE, ICT) on $256 \times 256$ \textbf{resized} images from \textbf{Places2}}
\label{tab:comparisonPlacesRe}
        \centering
               \begin{tabular}{P{1.5cm}P{2.3cm}P{0.85cm}P{0.85cm}P{0.7cm}P{0.35cm}P{0.85cm}P{0.85cm} P{1.7cm}}  
        \toprule
         & & \multicolumn{3}{c}{\textbf{Similarity to GT}} & & \multicolumn{2}{c}{\textbf{Realism}}  & \textbf{Diversity}\\
        \textbf{Mask} & \textbf{Method}  & PSNR$\uparrow$ & SSIM$\uparrow$ & L$^1$ $\downarrow$  & & MIS$\uparrow$  & FID$\downarrow$ &  LPIPS$\uparrow$ \\
        \toprule
        \multirow{3}{\linewidth}{Irregular $<20\%$} 
        & PIC & $\underline{29.86}$ & $0.934$ & $\textbf{2.14}$ & & $\underline{0.0489}$ & $32.3$ & $0.0055$ \\
         & DSI-VQVAE & $\textbf{30.64}$ & $\textbf{0.948}$ & $\underline{2.30}$ &  & $\textbf{0.0533}$ & $\textbf{20.0}$ & $\underline{0.0214}$ \\
        & ICT & $29.05$ & $\underline{0.939}$ & $3.83$ & & $0.0450$ & $\underline{23.0}$ & $\textbf{0.0224}$ \\
        \midrule
        \multirow{3}{\linewidth}{Irregular $[20\%,40\%]$} 
        & PIC & $\underline{22.98}$ & $0.808$ & $\underline{7.06}$ &  & $0.0394$ & $91.0$ & $0.0375$ \\
        & DSI-VQVAE  & $\textbf{23.04}$ & $\textbf{0.832}$ & $\textbf{6.92}$ &  & $\textbf{0.0443}$ & $\textbf{64.4}$ & $\underline{0.0789}$ \\
        & ICT & $22.11$ & $\underline{0.818}$ & $8.81$ &  & $\underline{0.0423}$ & $\underline{72.8}$ & $\textbf{0.0831}$ \\
       \midrule
         \multirow{3}{\linewidth}{Irregular $[40\%,60\%]$} 
        & PIC & $\underline{19.01}$ & $0.649$ & $\underline{14.71}$ &  & $0.0273$ & $144.2$ & $0.1357$ \\
        & DSI-VQVAE & $\textbf{19.15}$ & $\underline{0.684}$ & $\textbf{13.90}$ &  & $\underline{0.0287}$ & $\textbf{115.0}$ & $\underline{0.1700}$ \\
        & ICT & $18.50$ & $\textbf{0.669}$ & $15.78$ &  & $\textbf{0.0330}$ & $\underline{127.4}$ & $\textbf{0.1755}$ \\
        \midrule
                 \multirow{3}{\linewidth}{Central $128 \times 128$} 
        & PIC & $\textbf{19.50}$ & $\textbf{0.797}$ & $\textbf{10.27}$ &  & $0.0335$ & $104.5$ & $0.1129$ \\
        & DSI-VQVAE & $\underline{19.46}$ & $\textbf{0.797}$ & $\underline{10.60}$ &  & $\textbf{0.0387}$ & $\textbf{94.6}$ & $\textbf{0.1364}$ \\
        & ICT & $19.42$ & $0.796$ & $11.72$ &  & $\underline{0.0352}$ & $\underline{101.0}$ & $\underline{0.1284}$ \\
        \midrule
        \multirow{3}{\linewidth}{Random regular} 
        & PIC & $20.80$ & $0.773$ & $\underline{10.95}$ &  & $0.0359$ & $93.8$ & $0.1152$ \\
        & DSI-VQVAE  & $\textbf{21.15}$ & $\textbf{0.791}$ & $\textbf{10.48}$ &  & $\textbf{0.0426}$ & $\textbf{79.0}$ & $\underline{0.1233}$ \\
        & ICT & $\underline{21.03}$ & $\underline{0.787}$ & $11.51$ &  & $\underline{0.0382}$ & $\underline{84.3}$ & $\textbf{0.1239}$ \\
        \midrule
        \multirow{3}{\linewidth}{Random irregular} 
        & PIC & $\underline{19.91}$ & $0.640$ & $\underline{13.85}$ & & $0.0246$ & $157.7$ & $0.1023$ \\
        & DSI-VQVAE & $\textbf{20.05}$ & $\textbf{0.682}$ & $\textbf{12.98}$ &  & $\textbf{0.0329}$ & $\textbf{116.5}$ & $\underline{0.1539}$ \\
        & ICT & $19.10$ & $\underline{0.662}$ & $15.41$ &  & $\underline{0.0285}$ & $\underline{131.4}$ & $\textbf{0.1607}$ \\
               \specialrule{\heavyrulewidth}{1pt}{1pt}
        \multirow{3}{\linewidth}{Average} 
        & PIC & $\underline{22.01}$ & $0.767$ & $\underline{9.83}$ & & $0.0349$  & $103.9$ & $0.0848$ \\
        & DSI-VQVAE & $\textbf{22.25}$ & $\textbf{0.789}$ & $\textbf{9.53}$ &  & $\textbf{0.0401}$  & $\textbf{81.6}$ & $\underline{0.1140}$ \\
        & ICT &  $21.54$ & $\underline{0.779}$ & $11.18$  & & $\underline{0.0370}$ &   $\underline{90.0}$ & $\textbf{0.1157}$ \\
        \bottomrule
        \end{tabular}\vspace{0cm}
\end{table}

\begin{table}[h]
\caption{Quantitative comparison of three pluralistic image inpainting methods (PIC, DSI-VQVAE, ICT) on $256 \times 256$ \textbf{resized} images from \textbf{ImageNet}. }
\label{tab:comparisonImageNRe}
        \centering
               \begin{tabular}{P{1.5cm}P{2.3cm}P{0.85cm}P{0.85cm}P{0.7cm}P{0.35cm}P{0.85cm}P{0.85cm} P{1.7cm}} 
        \toprule
         & & \multicolumn{3}{c}{\textbf{Similarity to GT}} & & \multicolumn{2}{c}{\textbf{Realism}}  & \textbf{Diversity}\\
        \textbf{Mask} & \textbf{Method}  & PSNR$\uparrow$ & SSIM$\uparrow$ & L$^1$ $\downarrow$  & & MIS$\uparrow$  & FID$\downarrow$ &  LPIPS$\uparrow$ \\
        \toprule
        \multirow{3}{\linewidth}{Irregular $<20\%$} 
        &  PIC & $\underline{31.37}$ & $0.944$ & $\textbf{1.82}$ &  & $0.1885$ & $21.5$ & $0.0028$ \\
         & DSI-VQVAE & $\textbf{31.83}$ & $\textbf{0.952}$ & $\underline{2.08}$ &  & $\underline{0.1913}$ & $\underline{12.8}$ & $\underline{0.0175}$ \\
        & ICT & $30.21$ & $\underline{0.946}$ & $3.41$ & & $\textbf{0.2002}$ & $\textbf{12.2}$ & $\textbf{0.0203}$ \\
        \hline
        \multirow{3}{\linewidth}{Irregular $[20\%,40\%]$} 
        &  PIC & $\underline{23.13}$ & $0.807$ & $\underline{6.91}$ &  & $0.1401$ & $93.7$ & $0.0323$ \\
         & DSI-VQVAE & $\textbf{23.45}$ & $\textbf{0.825}$ & $\textbf{6.72}$ &  & $\underline{0.1617}$ & $\underline{61.8}$ & $\underline{0.0790}$ \\
        & ICT  & $22.36$ & $\underline{0.817}$ & $8.34$ & & $\textbf{0.1739}$ & $\textbf{52.1}$ & $\textbf{0.0810}$ \\
        \hline
        \multirow{3}{\linewidth}{Irregular $[40\%,60\%]$} 
        &  PIC & $\underline{18.39}$ & $\underline{0.636}$ & $15.84$ & & $0.0497$ & $198.0$ & $0.1314$ \\
        & DSI-VQVAE & $\textbf{18.95}$ & $\textbf{0.672}$ & $\textbf{14.14}$ &  & $\underline{0.0737}$ & $\underline{147.9}$ & $\textbf{0.1901}$ \\
        & ICT & $18.34$ & $\underline{0.663}$ & $15.85$ &  & $\textbf{0.0822}$ & $\textbf{120.4}$ & $\underline{0.1764}$ \\
        \hline
          \multirow{3}{\linewidth}{Central $128 \times 128$} 
        & PIC & $19.31$ & $0.795$ & $\underline{10.35}$ &  & $0.0583$ & $\underline{153.9}$ & $0.1091$ \\
       & DSI-VQVAE & $\underline{19.47}$ & $\textbf{0.800}$ & $\textbf{10.25}$ &  & $\underline{0.0700}$ & $172.1$ & $\textbf{0.1293}$ \\
        & ICT & $\textbf{19.91}$ & $\underline{0.796}$ & $11.27$ &  & $\textbf{0.0790}$ & $\textbf{120.3}$ & $\underline{0.1247}$ \\
        \hline
        \multirow{3}{\linewidth}{Random regular} 
        & PIC & $19.63$ & $0.745$ & $13.13$ &  & $0.0690$ & $150.5$ & $0.1071$ \\
        & DSI-VQVAE & $\textbf{20.13}$ & $\textbf{0.769}$ & $\textbf{11.59}$ &  & $\textbf{0.1048}$ & $\underline{113.8}$ & $\textbf{0.1457}$ \\
        & ICT & $\textbf{20.13}$ & $\underline{0.766}$ & $\underline{12.56}$ &  & $\underline{0.1028}$ & $\textbf{101.7}$ & $\underline{0.1376}$ \\
        \hline
        \multirow{3}{\linewidth}{Random irregular} 
        &  PIC & $\underline{19.70}$ & $0.618$ & $\underline{14.04}$ &  & $0.0457$ & $194.6$ & $0.1021$ \\
        & DSI-VQVAE & $\textbf{20.11}$ & $\textbf{0.665}$ & $\textbf{12.85}$ &  & $\underline{0.0642}$ & $\underline{155.9}$ & $\underline{0.1648}$ \\
        & ICT & $18.94$ & $\underline{0.649}$ & $15.33$ & & $\textbf{0.0859}$ & $\textbf{131.2}$ & $\textbf{0.1652}$ \\
                       \specialrule{\heavyrulewidth}{1pt}{1pt}
        \multirow{3}{\linewidth}{Average} 
        & PIC & $\underline{21.92}$ &  $0.758$& $\underline{10.35}$  & & $0.0919$ & $135.4$ & $0.0949$ \\
    & DSI-VQVAE  & $\textbf{22.32}$ & $\textbf{0.781}$ & $\textbf{9.61}$ & & $\underline{0.1110}$ & $\underline{107.7}$ & $\textbf{0.1211}$  \\
        & ICT  & $21.64$ & $\underline{0.773}$ & $11.13$ & & $\textbf{0.1207}$ & $\textbf{89.7}$ & $\underline{0.1175}$ \\
        \bottomrule
        \end{tabular}
\end{table}

\subsection*{Additional qualitative results}

\begin{figure}[ht]
\centering
\scriptsize Output on $256\times256$ images masked with $128\times128$ center hole\\
\includegraphics[width=0.105\textwidth]{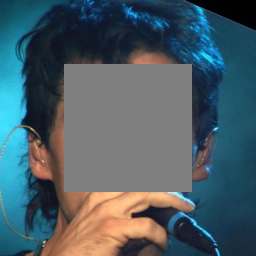}
\includegraphics[width=0.105\textwidth]{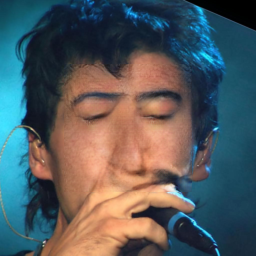}
\includegraphics[width=0.105\textwidth]{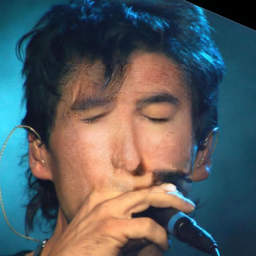}
\includegraphics[width=0.105\textwidth]{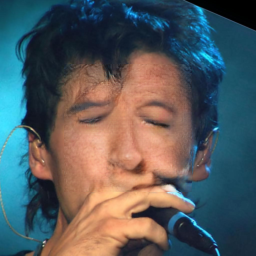}
\includegraphics[width=0.105\textwidth]{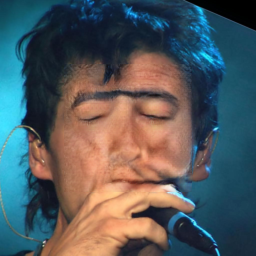}
\includegraphics[width=0.105\textwidth]{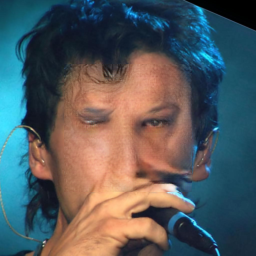}
\raisebox{0.5cm}{\rotatebox[origin=t]{90}{\scriptsize PIC}}
\\
\hspace{0.105\textwidth}
\includegraphics[width=0.105\textwidth]{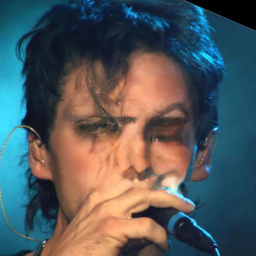}
\includegraphics[width=0.105\textwidth]{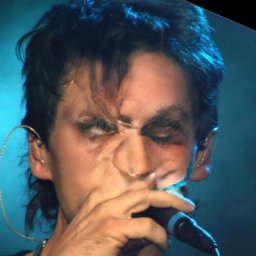}
\includegraphics[width=0.105\textwidth]{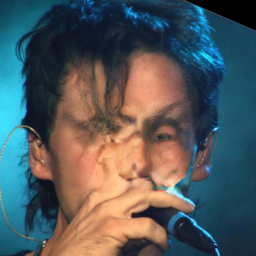}
\includegraphics[width=0.105\textwidth]{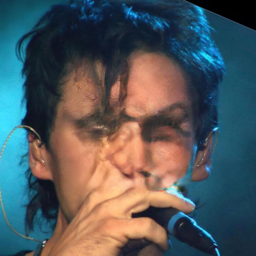}
\includegraphics[width=0.105\textwidth]{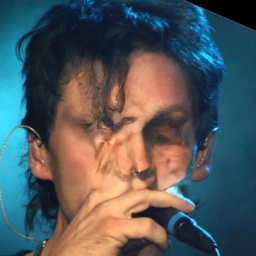}
\raisebox{0.45cm}{\rotatebox[origin=t]{90}{\scriptsize DSI-VQVAE}}\\
\hspace{0.105\textwidth}
\includegraphics[width=0.105\textwidth]{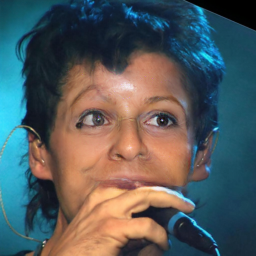}
\includegraphics[width=0.105\textwidth]{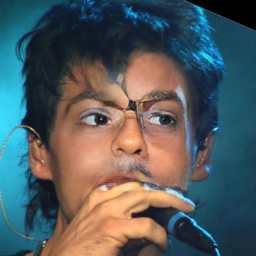}
\includegraphics[width=0.105\textwidth]{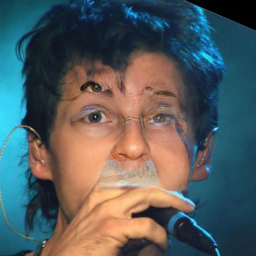}
\includegraphics[width=0.105\textwidth]{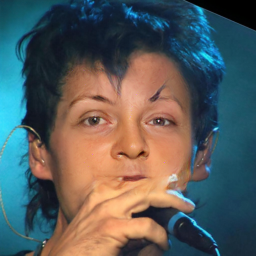}
\includegraphics[width=0.105\textwidth]{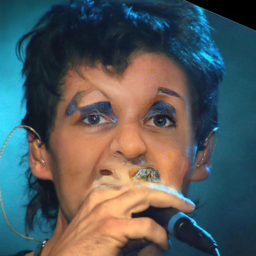}
\raisebox{0.5cm}{\rotatebox[origin=t]{90}{\scriptsize ICT}}
\\
\hspace{0.105\textwidth}
\includegraphics[width=0.105\textwidth]{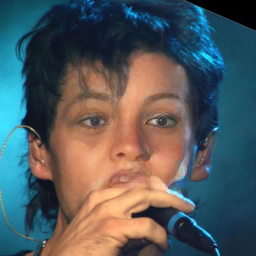}
\includegraphics[width=0.105\textwidth]{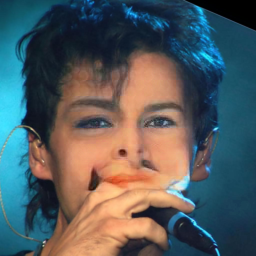}
\includegraphics[width=0.105\textwidth]{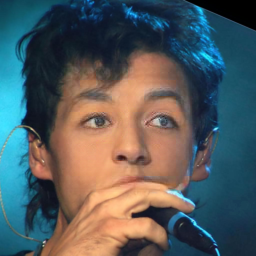}
\includegraphics[width=0.105\textwidth]{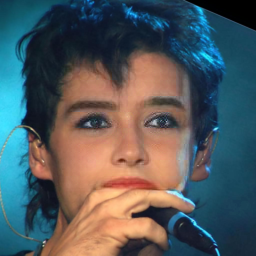}
\includegraphics[width=0.105\textwidth]{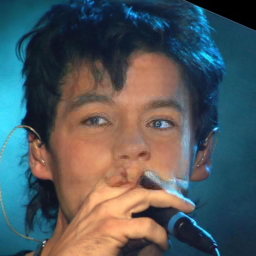}
\raisebox{0.5cm}{\rotatebox[origin=t]{90}{\scriptsize BAT}} 
\\
\scriptsize Output on $256\times256$ images masked with Pconv 20\% - 40\%
\\
\includegraphics[width=0.105\textwidth]{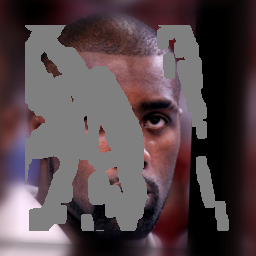}
\includegraphics[width=0.105\textwidth]{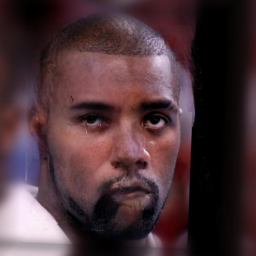}
\includegraphics[width=0.105\textwidth]{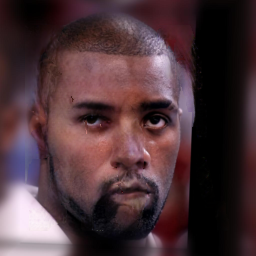}
\includegraphics[width=0.105\textwidth]{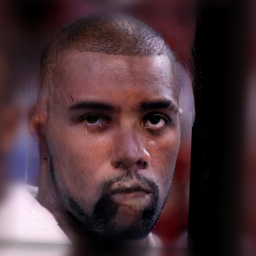}
\includegraphics[width=0.105\textwidth]{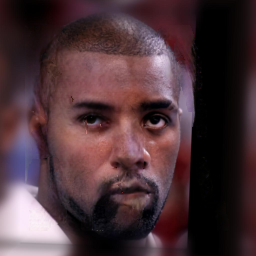}
\includegraphics[width=0.105\textwidth]{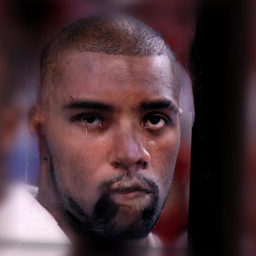}
\raisebox{0.5cm}{\rotatebox[origin=t]{90}{\scriptsize PIC}}
\\
\hspace{0.105\textwidth}
\includegraphics[width=0.105\textwidth]{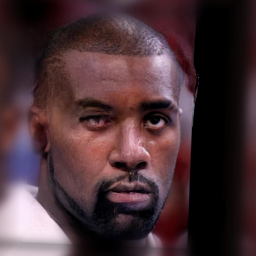}
\includegraphics[width=0.105\textwidth]{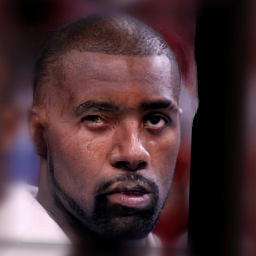}
\includegraphics[width=0.105\textwidth]{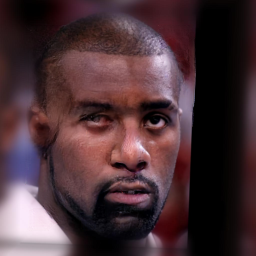}
\includegraphics[width=0.105\textwidth]{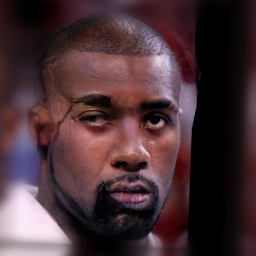}
\includegraphics[width=0.105\textwidth]{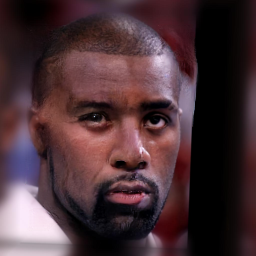}
\raisebox{0.45cm}{\rotatebox[origin=t]{90}{\scriptsize DSI-VQVAE}}
\\
\hspace{0.105\textwidth}
\includegraphics[width=0.105\textwidth]{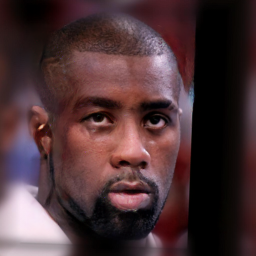}
\includegraphics[width=0.105\textwidth]{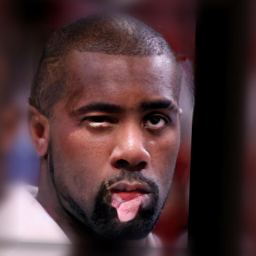}
\includegraphics[width=0.105\textwidth]{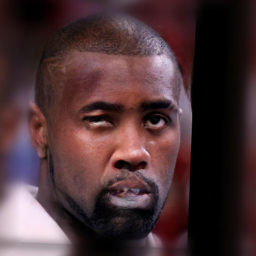}
\includegraphics[width=0.105\textwidth]{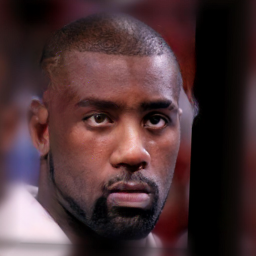}
\includegraphics[width=0.105\textwidth]{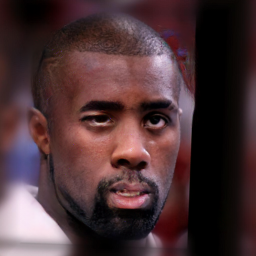}
\raisebox{0.5cm}{\rotatebox[origin=t]{90}{\scriptsize ICT}}
\\
\hspace{0.105\textwidth}
\includegraphics[width=0.105\textwidth]{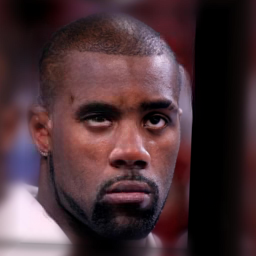}
\includegraphics[width=0.105\textwidth]{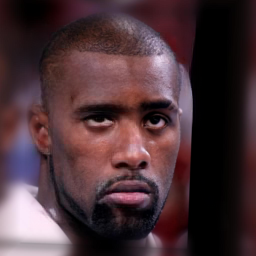}
\includegraphics[width=0.105\textwidth]{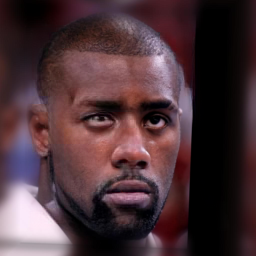}
\includegraphics[width=0.105\textwidth]{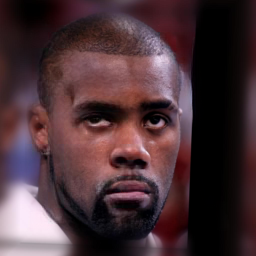}
\includegraphics[width=0.105\textwidth]{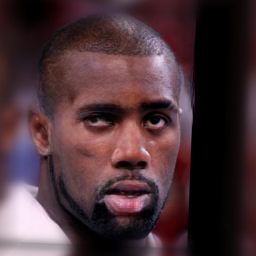}
\raisebox{0.5cm}{\rotatebox[origin=t]{90}{\scriptsize BAT}} 
\\
\scriptsize Output on $256\times256$ images masked with Pconv 40\% - 60\% \\
\includegraphics[width=0.105\textwidth]{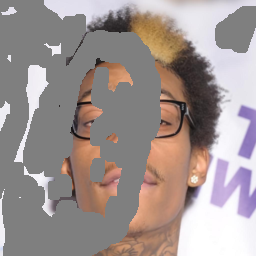}
\includegraphics[width=0.105\textwidth]{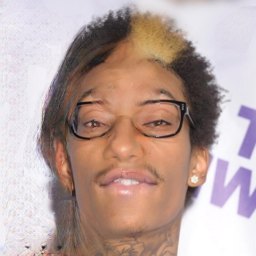}
\includegraphics[width=0.105\textwidth]{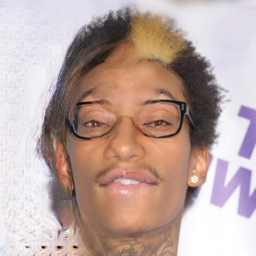}
\includegraphics[width=0.105\textwidth]{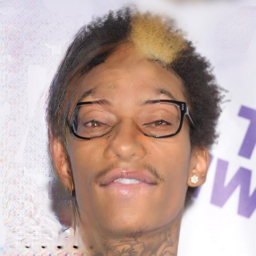}
\includegraphics[width=0.105\textwidth]{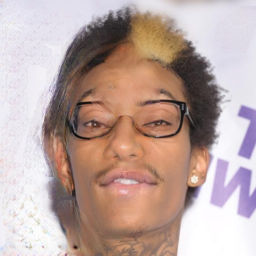}
\includegraphics[width=0.105\textwidth]{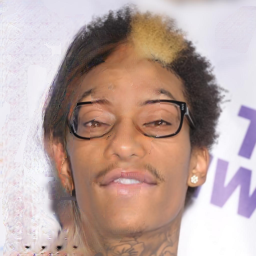}
\raisebox{0.5cm}{\rotatebox[origin=t]{90}{\scriptsize PIC}}
\\
\hspace{0.105\textwidth}
\includegraphics[width=0.105\textwidth]{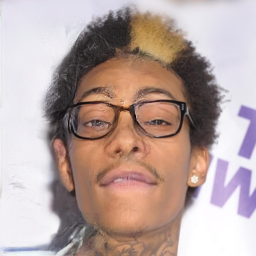}
\includegraphics[width=0.105\textwidth]{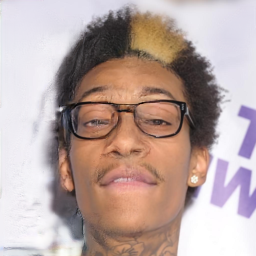}
\includegraphics[width=0.105\textwidth]{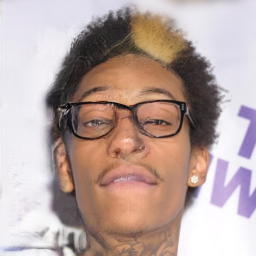}
\includegraphics[width=0.105\textwidth]{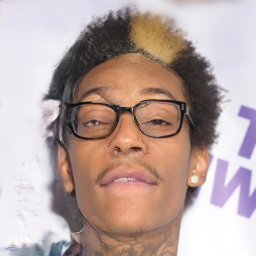}
\includegraphics[width=0.105\textwidth]{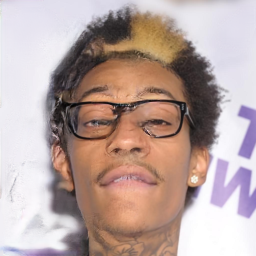}
\raisebox{0.45cm}{\rotatebox[origin=t]{90}{\scriptsize DSI-VQVAE}}
\\
\hspace{0.105\textwidth}
\includegraphics[width=0.105\textwidth]{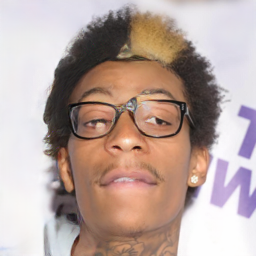}
\includegraphics[width=0.105\textwidth]{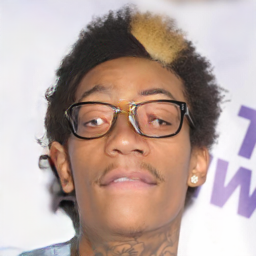}
\includegraphics[width=0.105\textwidth]{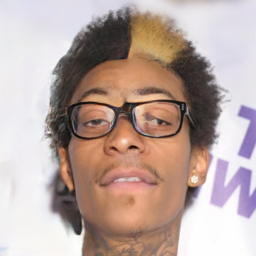}
\includegraphics[width=0.105\textwidth]{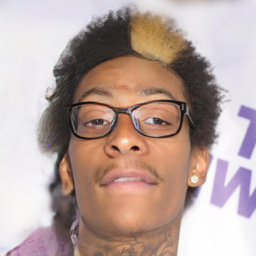}
\includegraphics[width=0.105\textwidth]{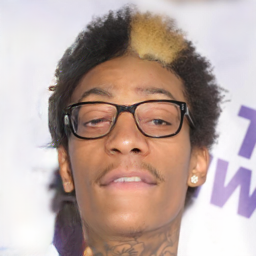}
\raisebox{0.5cm}{\rotatebox[origin=t]{90}{\scriptsize ICT}}
\\
\hspace{0.105\textwidth}
\includegraphics[width=0.105\textwidth]{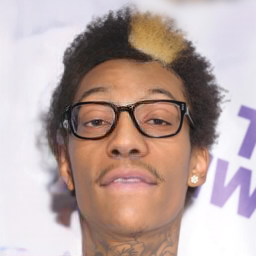}
\includegraphics[width=0.105\textwidth]{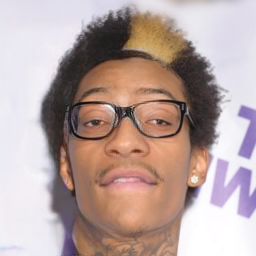}
\includegraphics[width=0.105\textwidth]{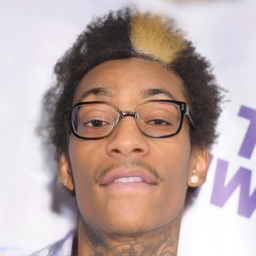}
\includegraphics[width=0.105\textwidth]{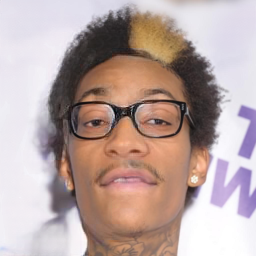}
\includegraphics[width=0.105\textwidth]{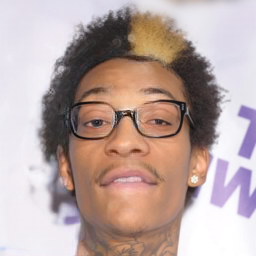}
\raisebox{0.5cm}{\rotatebox[origin=t]{90}{\scriptsize BAT}} 
\\
\caption{Diverse inpainting output on 256$\times$ 256 images from Celeba dataset with center, and irregular masks.  
For each method, out of 25 generated samples, the 5 samples with highest discriminator score are displayed.}
\label{fig:Celeba2}
\end{figure}

\begin{figure}[ht]
\centering
\scriptsize Output on $256\times256$ images masked with $128\times128$ center hole\\
\includegraphics[width=0.105\textwidth]{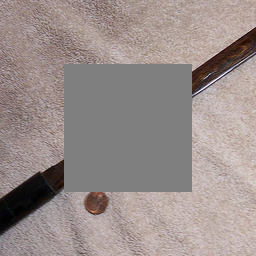}
\includegraphics[width=0.105\textwidth]{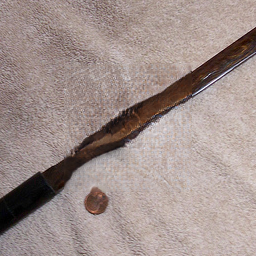}
\includegraphics[width=0.105\textwidth]{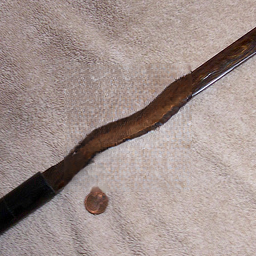}
\includegraphics[width=0.105\textwidth]{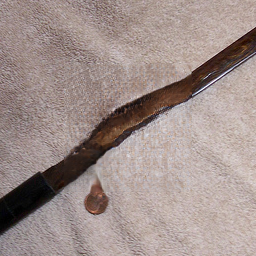}
\includegraphics[width=0.105\textwidth]{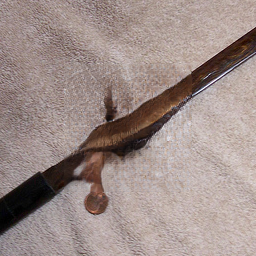}
\includegraphics[width=0.105\textwidth]{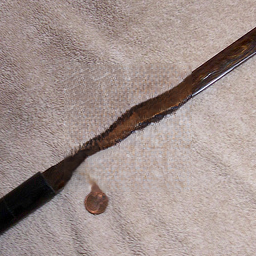}
\raisebox{0.5cm}{\rotatebox[origin=t]{90}{\scriptsize PIC}}
\\
\hspace{0.105\textwidth}
\includegraphics[width=0.105\textwidth]{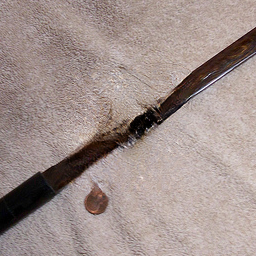}
\includegraphics[width=0.105\textwidth]{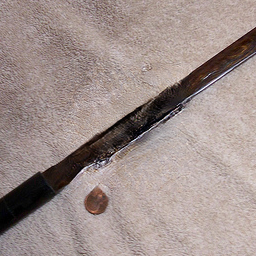}
\includegraphics[width=0.105\textwidth]{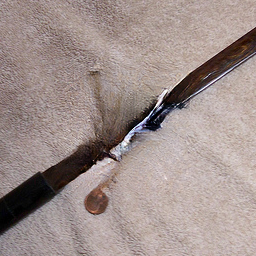}
\includegraphics[width=0.105\textwidth]{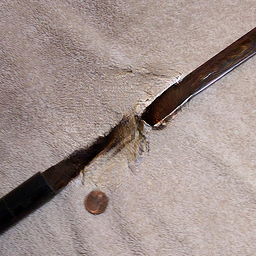}
\includegraphics[width=0.105\textwidth]{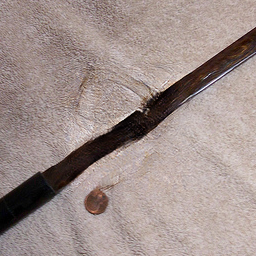}
\raisebox{0.45cm}{\rotatebox[origin=t]{90}{\scriptsize DSI-VQVAE}}\\
\hspace{0.105\textwidth}
\includegraphics[width=0.105\textwidth]{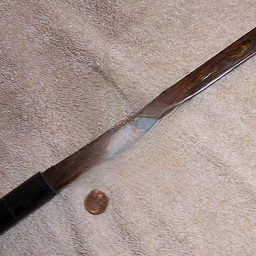}
\includegraphics[width=0.105\textwidth]{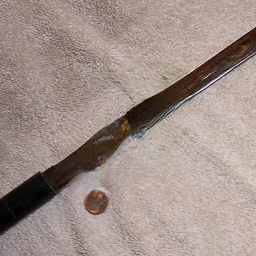}
\includegraphics[width=0.105\textwidth]{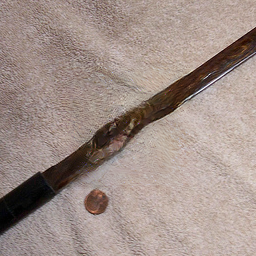}
\includegraphics[width=0.105\textwidth]{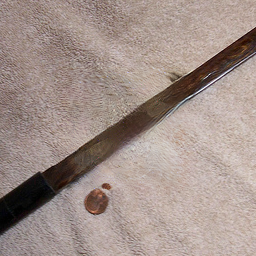}
\includegraphics[width=0.105\textwidth]{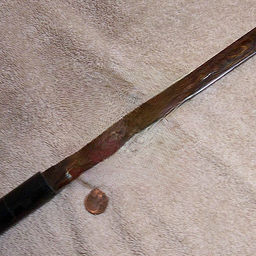}
\raisebox{0.5cm}{\rotatebox[origin=t]{90}{\scriptsize ICT}}
\\
\scriptsize Output on $256\times256$ images masked with Pconv < 20\%
\\
\includegraphics[width=0.105\textwidth]{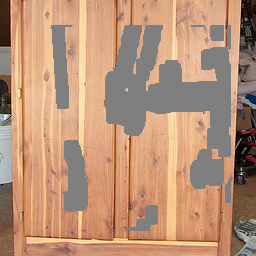}
\includegraphics[width=0.105\textwidth]{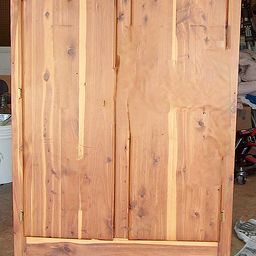}
\includegraphics[width=0.105\textwidth]{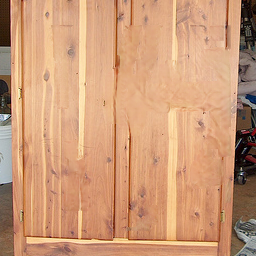}
\includegraphics[width=0.105\textwidth]{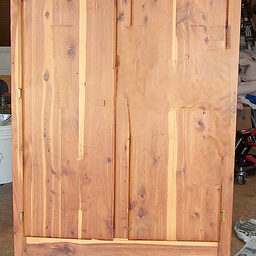}
\includegraphics[width=0.105\textwidth]{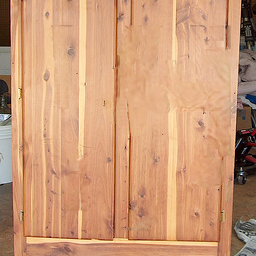}
\includegraphics[width=0.105\textwidth]{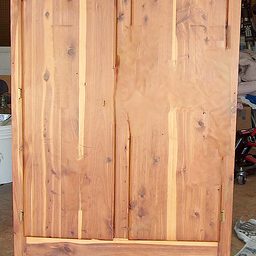}
\raisebox{0.5cm}{\rotatebox[origin=t]{90}{\scriptsize PIC}}
\\
\hspace{0.105\textwidth}
\includegraphics[width=0.105\textwidth]{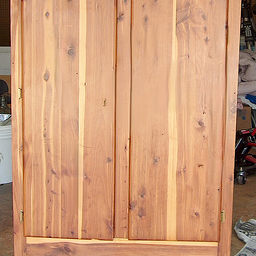}
\includegraphics[width=0.105\textwidth]{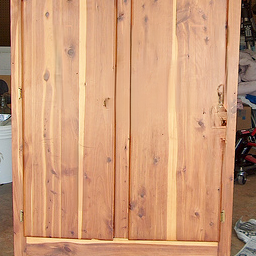}
\includegraphics[width=0.105\textwidth]{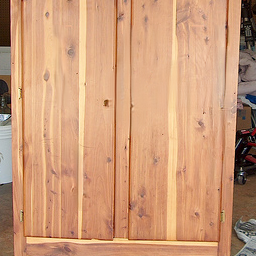}
\includegraphics[width=0.105\textwidth]{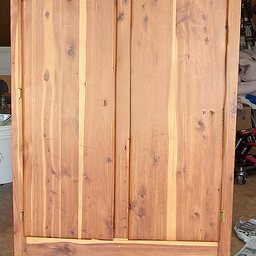}
\includegraphics[width=0.105\textwidth]{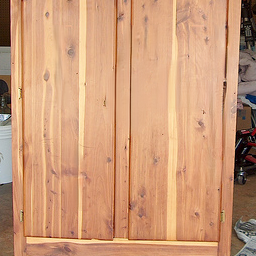}
\raisebox{0.45cm}{\rotatebox[origin=t]{90}{\scriptsize DSI-VQVAE}}\\
\hspace{0.105\textwidth}
\includegraphics[width=0.105\textwidth]{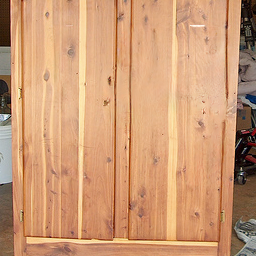}
\includegraphics[width=0.105\textwidth]{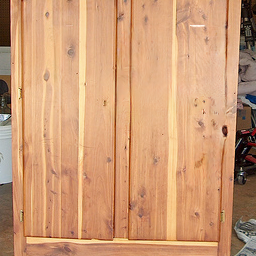}
\includegraphics[width=0.105\textwidth]{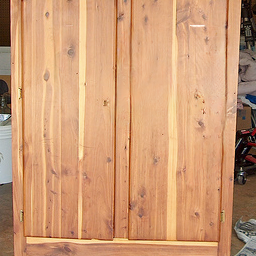}
\includegraphics[width=0.105\textwidth]{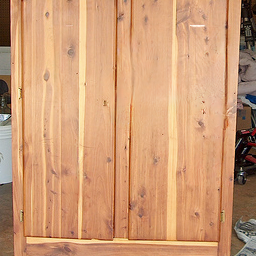}
\includegraphics[width=0.105\textwidth]{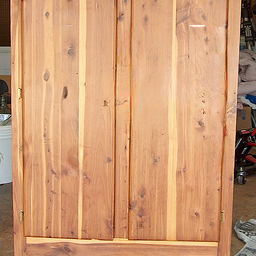}
\raisebox{0.5cm}{\rotatebox[origin=t]{90}{\scriptsize ICT}}
\\
\scriptsize Output on $256\times256$ images masked with Pconv 40\% - 60\%
\\
\includegraphics[width=0.105\textwidth]{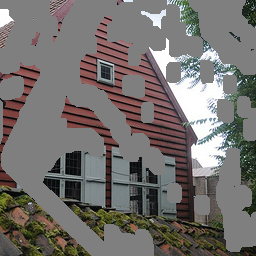}
\includegraphics[width=0.105\textwidth]{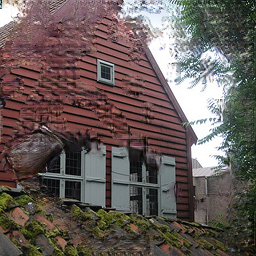}
\includegraphics[width=0.105\textwidth]{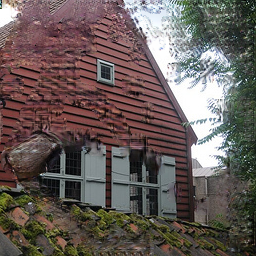}
\includegraphics[width=0.105\textwidth]{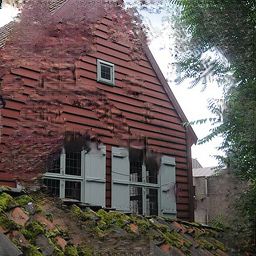}
\includegraphics[width=0.105\textwidth]{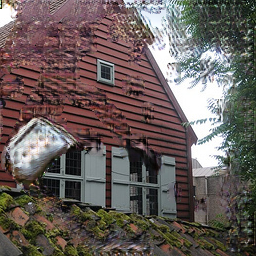}
\includegraphics[width=0.105\textwidth]{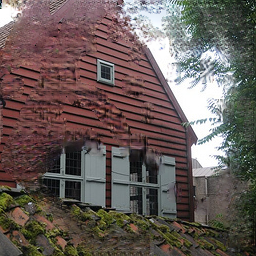}
\raisebox{0.5cm}{\rotatebox[origin=t]{90}{\scriptsize PIC}}
\\
\hspace{0.105\textwidth}
\includegraphics[width=0.105\textwidth]{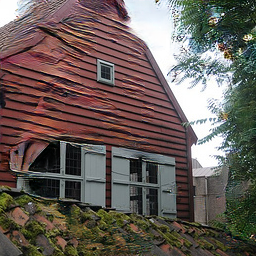}
\includegraphics[width=0.105\textwidth]{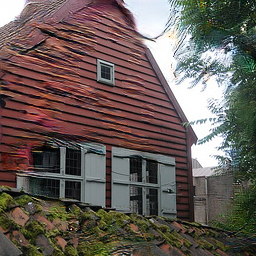}
\includegraphics[width=0.105\textwidth]{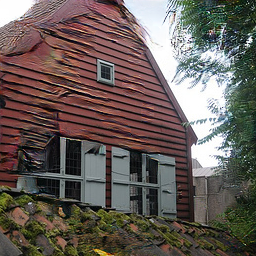}
\includegraphics[width=0.105\textwidth]{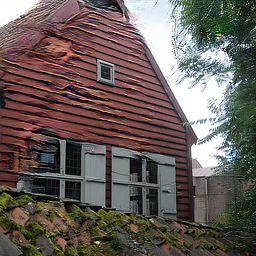}
\includegraphics[width=0.105\textwidth]{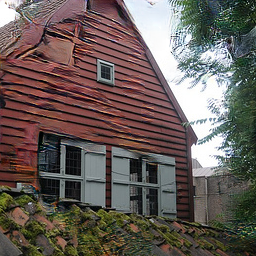}
\raisebox{0.45cm}{\rotatebox[origin=t]{90}{\scriptsize DSI-VQVAE}}\\
\hspace{0.105\textwidth}
\includegraphics[width=0.105\textwidth]{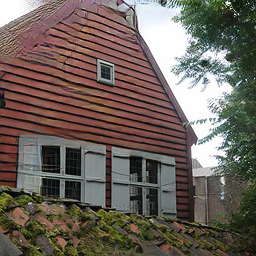}
\includegraphics[width=0.105\textwidth]{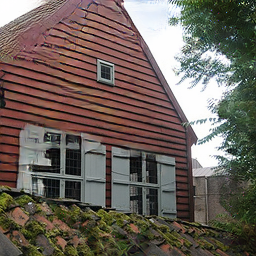}
\includegraphics[width=0.105\textwidth]{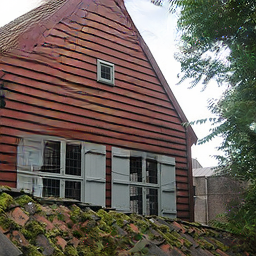}
\includegraphics[width=0.105\textwidth]{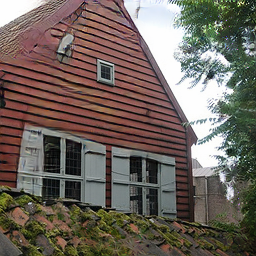}
\includegraphics[width=0.105\textwidth]{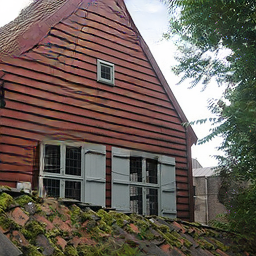}
\raisebox{0.5cm}{\rotatebox[origin=t]{90}{\scriptsize ICT}}
\\
\caption{Diverse inpainting output on 256$\times$ 256 images from ImageNet dataset with center and irregular masks with different hidded proportions.  
For each method, out of 25 generated samples, the 5 samples with highest discriminator score are displayed.}
\label{fig:imagenet2}
\end{figure}

In Figures~\ref{fig:Celeba2} and \ref{fig:imagenet2} we show additional inpainting visual results on Celeba-HQ and ImageNet datasets. 

\end{document}